\documentclass[acmsmall]{acmart}

\usepackage{hyperref}
\usepackage{adjustbox}
\usepackage{array,graphicx}
\usepackage{booktabs}
\usepackage{multirow}
\usepackage{pifont}
\usepackage{tabularray}
\usepackage{color}
\newcommand*\rot{\rotatebox{90}}
\newcommand*\OK{\ding{51}}

\AtBeginDocument{%
 \providecommand\BibTeX{{%
  \normalfont B\kern-0.5em{\scshape i\kern-0.25em b}\kern-0.8em\TeX}}}

\setcopyright{acmcopyright}
\copyrightyear{2024}
\acmYear{xxxx}
\acmDOI{XXXXXXX.XXXXXXX}

\acmJournal{JACM}
\acmVolume{1}
\acmNumber{1}
\acmArticle{1}
\acmMonth{1}
\acmSubmissionID{TBA}
\usepackage{comment}

\usepackage{xcolor}

\newcommand{\redtext}[1]{\textcolor{black}{#1}}
\newcommand{\bluetext}[1]{\textcolor{black}{#1}}

\newcommand{\orangetext}[1]{\textcolor{black}{#1}}
\newcommand{\greentext}[1]{\textcolor{black}{#1}}

\begin{document}

\title{Natural Language Processing for Dialects of a Language: A Survey}

\author{Aditya Joshi}
\email{aditya.joshi@unsw.edu.au}
\orcid{0000-0003-2200-9703}
\affiliation{%
 \institution{University of New South Wales}
 \city{Sydney}
 \country{Australia}
 \postcode{2052}
}

\author{Raj Dabre}
\email{raj.dabre@nict.go.jp}
\orcid{0000}
\affiliation{%
 \institution{National Institute of Information and Communications Technology}
 \city{Kyoto}
 \country{Japan}
}

 \author{Diptesh Kanojia}
\email{d.kanojia@surrey.ac.uk}
\orcid{0000-0001-8814-0080}
\affiliation{%
 \institution{Institute for People-Centred AI, University of Surrey}
 \city{Guildford}
 \country{United Kingdom}
}

\author{Zhuang Li}
\email{zhuang.li1@monash.edu}
\orcid{0000-0002-9808-9992}
\affiliation{%
 \institution{Monash University}
 \city{Melbourne}
 \country{Australia}
}

\author{Haolan Zhan}
\email{haolan.zhan@monash.edu}
\orcid{0000-0001-9628-5992}
\affiliation{%
 \institution{Monash University}
 \city{Melbourne}
 \country{Australia}
}

\author{Gholamreza Haffari}
\email{Gholamreza.Haffari@monash.edu}
\orcid{0000-0001-7326-8380}
\affiliation{%
 \institution{Monash University}
 \city{Melbourne}
 \country{Australia}
}

\author{Doris Dippold}
\email{d.dippold@surrey.ac.uk}
\orcid{0000-0001-6193-4710}
\affiliation{%
 \institution{University of Surrey}
 \city{Guildford}
 \country{United Kingdom}
}
\renewcommand{\shortauthors}{Joshi, et al.}

\begin{abstract}
State-of-the-art natural language processing (NLP) models are trained on massive training corpora, and report a superlative performance on evaluation datasets. This survey delves into an important attribute of these datasets: the dialect of a language. Motivated by the performance degradation of NLP models for \redtext{dialectal} datasets and its implications for the equity of language technologies, we survey past research in NLP for dialects in terms of datasets, and approaches. We describe a wide range of NLP tasks in terms of two categories: natural language understanding (NLU) (for tasks such as dialect classification, sentiment analysis, parsing, and NLU benchmarks) and natural language generation (NLG) (for summarisation, machine translation, and dialogue systems). The survey is also broad in its coverage of languages which include English, Arabic, German\orangetext{,} among others. We observe that past work in NLP concerning dialects goes deeper than mere dialect classification, and \redtext{extends to several NLU and NLG tasks}.  \orangetext{For these tasks, we describe classical machine learning using statistical models, along with} the recent \orangetext{deep learning-based approaches based on pre-trained language models}. We expect that this survey will be useful to NLP researchers interested in building equitable language technologies by rethinking LLM benchmarks and model architectures.


\end{abstract}



\received{\redtext{10 Jan 2024}}
\received[revised]{\redtext{19 Nov 2024}}
\received[accepted]{TBA}

\maketitle

\section{Introduction}

Natural language processing (NLP) is an area of artificial intelligence that deals with \redtext{processing of human language in its textual form}. NLP tasks are broadly viewed as two categories: natural language understanding (NLU) and natural language generation (NLG). The former broadly covers language understanding tasks \orangetext{such as dialect identification or sentiment classification}, as \redtext{well as} tasks such as morphosyntactic analysis. The latter includes tasks where both the input and the output are textual sequences \redtext{(for example, summarisation)}. The state-of-the-art NLP, for both NLU and NLG, is based on Transformer-based models\orangetext{~\cite{naveed2023comprehensive,zhao2023survey}}. Large language models (LLMs) that use decoders in the Transformer architecture have significantly increased attention toward NLP \orangetext{leading to LLM-based applications in several domains such as medicine, business or law}. LLMs released by commercial organisations report an increasingly higher number of parameters and, \redtext{as a result}, \orangetext{improved performances on several NLP tasks.} NLP approaches using LLMs are largely viewed as black-box models trained on massive corpora whose composition is not accurately known. This survey dissects one of many attributes in which variations may exist in the training and test corpora: dialects of a language.

\redtext{Traditionally, }a dialect is defined as the regionally or locally based \redtext{variety} of a language~\cite{haugen1966dialect}. Wikipedia defines a dialect as ``a variety of a language that is a characteristic of a particular group of the language's speakers.'' \orangetext{\citet{zampieri2021similar} state that dialects are language varieties characterised by
systematic patterns of variation}. \redtext{The current notion of dialect has extended to language varieties arising due to factors such as political reasons, country of origin, migration histories, historical factors, register shifts and so on.} \orangetext{ In} fact, there is an association between perceived social hierarchies and dialects of a language, \redtext{leading to a term `sociolect'}~\cite{kroch1986toward}. \redtext{For the sake of brevity, we use `dialects' as an umbrella term to refer to} `\redtext{dialects/national varieties}/cultural variants\redtext{/sociolects}' of a language while acknowledging that \redtext{the distinction between dialects and language is nuanced~\cite{sandel2015dialects}}. \redtext{An example of a dialect} is \redtext{the national variety}, \orangetext{Australian English, whose phonemes are predominantly derived from} Southern British English \redtext{and other Englishes}~\cite{cox2007australian,cox2006acoustic}, but has also developed its own unique vocabulary~\cite{moore1999vocabulary}. \orangetext{Overlapping with dialects are Creole languages that develop from the process of different languages simplifying and mixing into a new form (often, a pidgin), and then that form expanding and elaborating into a full-fledged language with native speakers, all within a fairly brief period. \citet{lent2023creoleval} highlight the social and scholarly stigmatisation of Creole languages that has resulted in limited advances in NLP for these languages.}

In general, our survey is catalysed by the recent efforts in extending LLMs on NLP tasks for dialects of different languages. As researchers continue to look `under the hood' of LLMs, \redtext{dialectal} differences in training and testing datasets \redtext{are being} increasingly scrutinised, and \redtext{adaptation techniques to improve their performance on different} dialects \redtext{are being devised}. As a result, we hope that this survey will help readers and researchers understand past work in NLP techniques for dialects of a language, and contribute to ideas about fair and equitable NLP in the future.

\redtext{There have been related surveys in the past}.~\citet{zampieri2020natural} describes the available corpora, and past approaches to fundamental NLP problems such as POS tagging and parsing, along with applications to NLP. Our survey builds upon theirs in three ways. Firstly, we cover a wider range of downstream tasks \orangetext{such as summarisation and sentiment analysis}. Also, this survey contains recent papers, which highlight increasingly growing attention towards NLP for dialects. Finally, the exposition of our survey \orangetext{adopts} a deep learning-centric view\orangetext{, by covering deep learning-based approaches, in particular, the recent LLM-based approaches}. Another survey by~\citet{blodgett2020language} describes biases of different kinds in an analysis of language technologies, including \redtext{dialectal} bias. We derive from their survey to formulate the motivation and trends in NLP for dialects. \redtext{Similarly,~\citet{jauhiainen2019automatic} present a survey of automatic language identification\orangetext{,} which does not differentiate between dialect or language identification, and mention that dialect identification may be a more challenging task.} Finally, extensive surveys focusing on languages from the Middle East have been reported~\cite{darwish2021panoramic, shoufan2015natural}. These are surveys of NLP for standard and \redtext{dialectal} Arabic, primarily focusing on dialect identification and synthesis in the form of machine translation. Our survey unifies the efforts in dialects of languages belonging to multiple language families. The contribution of our survey is:
\begin{itemize}
\item We present past work in terms of NLU and NLG tasks\orangetext{, and include both pre-deep learning and deep learning techniques.}
\item We highlight trends and future directions, and provide summary tables that will help researchers interested in \redtext{dialectal} NLP research.
\item The survey covers a broad range of languages from around the \orangetext{world.}
\end{itemize}

The rest of the paper is organised as follows. We motivate the need for a discussion on dialects in Section~\ref{sec:motivation}. We define the scope of the paper and highlight key trends in Section~\ref{sec:scopetrends}. We then cover dialect-specific resources in Section~\ref{sec:resources}. Following that, Section~\ref{sec:nlu} covers several NLU tasks: dialect identification, sentiment analysis, parsing, and NLU benchmarks. Section~\ref{sec:nlg} presents relevant approaches in NLG for machine translation, summarisation and so on. Finally, we conclude the survey and discuss future work in the context of NLP research as well as social/ethical implications in Section~\ref{sec:concl}. The survey contains several summary tables that will be useful for future research.

\section{Motivation}
\label{sec:motivation}
\subsection{Linguistic Challenges Posed by Dialects}

\begin{figure}
    \centering
    \includegraphics[width=0.6\linewidth]{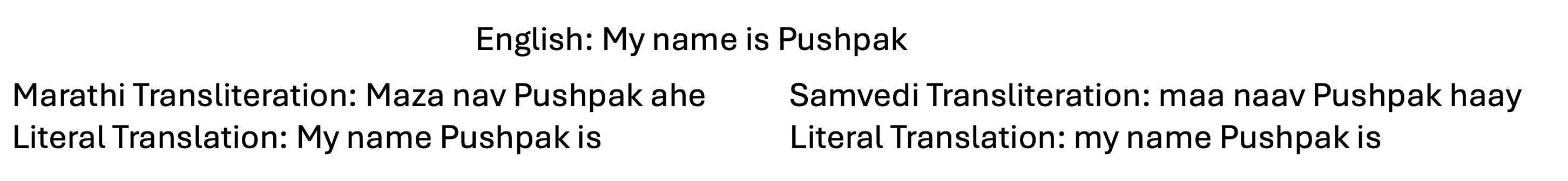}
    \caption{An example sentence highlighting the differences between Marathi and its Samvedi dialect.}
    \label{fig:marathi-samvedi}
\end{figure}

\redtext{Dialectal} differences primarily occur in terms of \orangetext{orthography,} syntax and vocabulary. Some examples of \redtext{dialectal} differences in English are: `I might could help you with that' observed in \orangetext{Southern US,} Australian and New Zealand English~\cite{morin2023double, coats-2022-corpus} as well as British and Irish English~\cite{coats2023double},  `Inside tent can not see leh !' in \orangetext{Singaporean English~\cite{wang-etal-2017-universal} or} the \orangetext{uncommon} placement of adverbs in native speakers of Asian languages as in `Already, I have done it.'~\cite{nagata-2014-language}. Also, consider the case of the Samvedi dialect of Marathi, one of $42$, \redtext{where we give an example of the Samvedi and Marathi sentences in Table~\ref{fig:marathi-samvedi}.} Samvedi does not exhibit word order differences compared to standard Marathi, but it involves heavy pronunciation relaxation \redtext{(\textit{ahe} -> \textit{hay}, and \textit{maza} -> \textit{maa}) and the usage of older words.} Another challenge in handling dialects is that two dialects of the same language can be mutually unintelligible. A classic example of this \orangetext{is the} case of the \href{https://en.wikipedia.org/wiki/Tsugaru_dialect}{Aomori} and \href{https://en.wikipedia.org/wiki/Okinawan_Japanese}{Okinawan} dialects of Japanese\orangetext{,} which has a total of \href{https://en.wikipedia.org/wiki/Japanese_dialects}{47} known dialects. Therefore, it is not enough to collect data for one dialect and assume that it will help in NLP for another dialect\orangetext{,} which indicates that special attention will need to be paid to each dialect to ensure that it will be well-represented. Dialects assume further importance when people from different cultural backgrounds interact with \orangetext{one another}\orangetext{. ~\citet{wang-etal-2022-perceptual}} show that monophthongal vowels spoken by Australian English speakers may be difficult to be understood by Mandarin English listeners.

\redtext{Dialects are also associated with} pragmatics, with influences derived from macro-social factors such as region, social class, ethnicity, gender, age~\cite{haugh2012politeness}. For example, \citet{schneider2012appropriate} observes differences in small talk across inner circle varieties \orangetext{of English, i.e., varieties from countries where it is the primary language~\cite{kachru1992other}. They also observed differences between speakers of different ages and genders. This suggests that the notion of `dialect' can be linked to factors beyond geographical distribution.}~\citet{merrison2012getting} showed that, in student requests to university staff, there were differences in the way obligation was expressed, and that these differences were linked to different ways of claiming social standing. \orangetext{~\citet{meyer2014culture} compares interactions of Australians with people from other cultures in terms of (a) building trust with colleagues, (b) leading teams of a culturally dissimilar background \textit{etc.} An example in the book states that an Australian may invest in shorter small talk than a Mexican with a colleague.} Noting the differences in the pragmatic strategies of different dialect speakers provide an important social perspective on dialectal variation. However, these are currently not sufficiently accounted for in NLP. 

\subsection{Rethinking LLM benchmarks}
There are more English language speakers in countries such as India than the United States, Australia and England~\cite{dunn-2019-modeling}. In addition, an even larger number of speakers have acquired English in a classroom context (\textit{e.g.,} in countries such as China, Germany or Russia) and use it mainly as a contact language for specific transactional purposes, \textit{e.g.,} business or education. This latter perspective has been described through the notion of English as a lingua franca as ``the common language of choice […] among speakers who come from different lingua-cultural backgrounds''~\cite{jenkins2009english}. Despite that, the corpora used to train language models and more importantly, the datasets used to evaluate them do not necessarily reflect dialectal variations within a language. \bluetext{\citet{inoue2021interplay} examine the performance of BERT-based models for varieties/dialects of Arabic, and show that dialect proximity of pre-training and fine-tuning data bears impact on the performance of the downstream task}. In the case of GPT-4, the evaluation dataset consists of questions from the MMLU benchmark written in Standard American English. Standard benchmarks used to claim performance of a language model for English primarily contain Standard American English. It has been found that the performance does not extend to NLU tasks for dialects of English~\cite{ziems-etal-2022-value}. \greentext{Further, a recent work by \citet{fleisig-etal-2024-linguistic} analyses the output of ChatGPT for varieties of English, and shows that the generated output may be of poorer quality and be prone to stereotyping for non-standard dialects of English.}
These findings holds for most foundation models that are trained on large amounts of data. The distribution of languages in the training corpora is either not known or difficult to determine. 

\subsection{Fair and equitable technologies}
NLP systems that are deployed to serve multicultural communities must be mindful of the variations between different dialects. Evaluation and mitigation of disparity between dialects become an overgrowing need in times when language models claim excellent language performance using datasets from a specific dialect alone. \orangetext{Some examples showing the impact of dialects on the performance of NLP tasks are presented in Table~\ref{tab:examplesimpact}. We note that these papers are from the past few years, which have otherwise witnessed a great development in the reported performance of NLP models.}

\orangetext{Some implications of dialects in terms of} sociological factors \orangetext{are}:
\begin{enumerate} 
\item \textbf{Performance of NLP models and per-capita GDP}: A recent work by~\citet{kantharuban-etal-2023-quantifying} show the \redtext{dialectal} gap in performance of LLM-based solutions for machine translation and automatic speech recognition for several dialects\orangetext{, similar to ~\citet{ahia2023all} who show the same for topologically diverse languages}. They show a positive correlation between gross domestic product per capita and the efficacy of \redtext{dialectal} machine translation.
\item \textbf{Healthcare monitoring}:~\citet{jurgens-etal-2017-incorporating} show that there exists a disparity between popular dialect speakers and others in the case of healthcare monitoring\footnote{They also propose a method to mitigate the disparity.}.
\item \textbf{Racial biases in hate speech detection}:~\citet{okpala2022aaebert} show that hate speech classifiers may lean towards predicting a text as true if it uses African-American English.
\item \bluetext{\textbf{Prejudice in the prediction of employability and criminality}: \citet{hofmann2024dialect} show that dialects may introduce bias in the output of language models. As a result, a person's output with respect to their employability or criminality may be affected based on the dialects they use.}
\end{enumerate}

NLP may not perform as well for dialects of a language, particularly spoken by historically marginalized communities such as the African-American community. This has been shown for language identification where dialects are not predicted as the language since they differ from the standard version of the language~\cite{blodgett-etal-2016-demographic}. 

An idea closely related to the survey is the `Bender rule' in NLP research. The Bender rule states that the language of datasets used for evaluation must be stated explicitly without assuming English to be the implicit default~\cite{ducel2022we}. We similarly believe that languages are not monoliths and \redtext{dialectal} differences must be clearly stated. Similarly, ~\citet{hovy2021importance} \orangetext{show} that incorporating \redtext{dialectal} aspects is closely related to social factors of language. As a result, incorporating an understanding of dialects of a dataset is pivoting for fairer NLP tools. 

\begin{table}
\centering
\begin{tblr}{
  width = \linewidth,
  colspec = {Q[225]Q[257]Q[617]},
  hline{1,8} = {-}{0.08em},
  hline{2,5} = {-}{0.05em},
}
\textbf{NLP Task}                  & \textbf{Paper} & \textbf{Impact}                                                                                 \\
{Language \\classification}        &~\cite{blodgett-etal-2016-demographic} & {Language detection shows lower performance\\for African-American English.}                     \\
{Sentiment \\classification}       &~\cite{okpala2022aaebert} & {Text in African-American English may be\\predicted \redtext{more commonly} as hate speech.}                    \\
{Natural Language \\Understanding} &~\cite{ziems-etal-2022-value} & {Popular models perform worse on GLUE tasks\\for African-American English text.}               \\
Summarisation                      &~\cite{keswani2021dialect} & {Generated multi-document summaries\\may be biased towards majority dialect.}    \\
Machine translation                &~\cite{kantharuban-etal-2023-quantifying}  & {Significant drop in MT from and to dialects of\\~Portuguese/Bengali/etc. to and from English.} \\
Parsing                            &~\cite{scannell-2020-universal} & {Lower performance of parsers on \redtext{Manx Gaelic} \\as compared to Irish/Scottish \redtext{Gaelic}.}       
\end{tblr}
\caption{Examples of adverse impact on NLP task performance due to \redtext{dialectal} variations.}
\label{tab:examplesimpact}
\end{table}


\begin{figure*}
  \centering
  \includegraphics[width=0.38\textwidth]{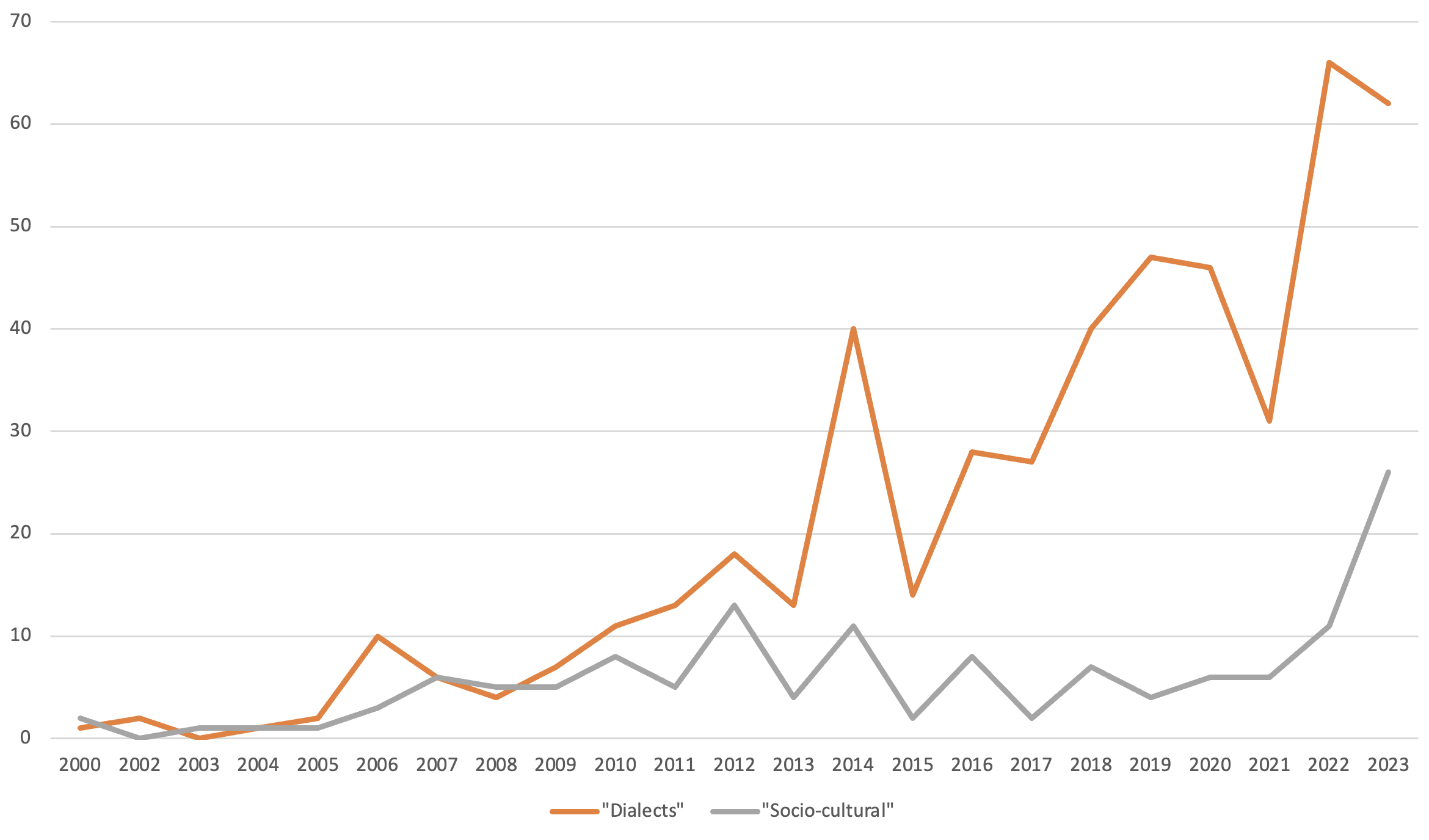}
  \caption{Number of relevant `papers-per-year' for keywords `dialects' and `socio-cultural' in the ACL Anthology.}
  \label{fig:timeline}
\end{figure*}
\subsection{Recent work}
One observes a renewed interest in using dialects to inform NLP tasks, as shown in Figure~\ref{fig:timeline}. The figure was generated using the ACL anthology\orangetext{\footnote{\url{https://aclanthology.org/info/development/}\orangetext{; Accessed on 9th January, 2024.}}}. For ``dialects'', we use `dialect', `national variety' (subword for inflections of `variety'), `national variation', and `Creole'. For ``socio-cultural'', we use the words ``cultural", and ``socio-cultural". We restrict to the year range 2000-2023. 

Dialect awareness has been shown to improve the performance of NLP tasks such as machine translation~\cite{sun-etal-2023-dialect}, speech recognition~\cite{pluss-etal-2023-stt4sg}. Recent works have also focused on dialect-aware NLP tasks as in the case of machine translation of dialect to standard language translation as in the case of Chinese~\cite{lu-etal-2022-exploring} (for \redtext{Hokkien}, a dialect of Chinese).

\section{Scope \& Trends}
\label{sec:scopetrends}
The focus of this survey \orangetext{is on NLP} approaches that are aware of dialects: either in the form of the choice of the dataset, incorporation in the model or evaluation along dimensions involving dialect. The survey provides a broad introduction to past NLP research on dialects spoken in different parts of the world. In the forthcoming subsections, we clarify the scope of this paper (Section~\ref{sec:scope}) and highlight key trends (Section~\ref{sec:trends}) that are described in detail in the following sections.

\subsection{Scope}
\label{sec:scope}
We select papers that mention the dialect as an attribute of interest. The focus on dialects is either based on the evaluation datasets or the model innovations to improve performance on dialect-specific datasets.

We keep the following out of scope, primarily to effectively manage the scope of the paper:
\begin{enumerate}
  \item \textbf{Code-mixing}: Code-mixing involves the use of words from two or more languages, often to reduce cognitive \orangetext{load. This} survey does not focus on code-mixing.
  \item \textbf{Implicit selection biases}: We also acknowledge that selection biases in datasets may introduce \redtext{dialectal} variations. For example, a dataset of tweets downloaded from a specific country is likely to have predominant dialects spoken in the country. However, we cannot locate these papers in particular, or, for social implications, claim that they are based on \redtext{dialectal} variations of a language without the authors mentioning so.
  \item \textbf{Accent variations}: Finally, we focus on `text'-based research while acknowledging that the speech processing community has a rich history of using acoustic data centered around accent. \orangetext{To this end, we briefly touch upon speech since dialects and speech are intertwined to a certain degree. However, our primary focus is on text since the text aspect has received a lot more attention than the speech aspect. This also sets up a situation where a future survey can expand on the speech aspect of dialectal processing. The focus on the textual form is the typical purview of NLP}.
  \item \textbf{Systematic review}: This survey is not a systematic review \orangetext{in the sense that we do not exhaustively cover \textbf{all} works on dialects due to limited time and paper space. Instead}, we select key representative papers based on our interpretation of the innovation, \orangetext{which, according to us, cover key progress and innovation in NLP for dialects}. We acknowledge that we may have missed out on \orangetext{some} important papers in the field. We will incorporate these papers as communicated by readers/reviewers. However, we cover a broad range of approaches in the survey.
  \item \textbf{Linguistic studies}: While we acknowledge similar rich linguistic work in terms of understanding dialects, we focus on NLP tasks\footnote{We cover dialect classification in the section on natural language understanding.}. For example, dialectometry is a research area that studies variations in dialects of a language~\cite{goebl1993dialectometry} but is not included in the survey.
\end{enumerate}
\begin{table*}
\resizebox{!}{0.4\paperheight}{%
\begin{tabular}{|l|l|p{0.3cm}|p{0.3cm}|p{0.3cm}|p{0.3cm}|p{0.3cm}|p{0.3cm}|p{0.3cm}|p{0.3cm}|p{0.3cm}|p{0.3cm}|p{0.3cm}|p{0.3cm}|p{0.3cm}|p{0.3cm}|p{0.3cm}|p{0.3cm}|}
\toprule
& \multicolumn{6}{|l|}{Languages} & \multicolumn{4}{|l|}{Innovation} & \multicolumn{7}{|l|}{Problem/Area}                \\ \midrule
 & \rot{English}        & \rot{Chinese} & \rot{Arabic} & \rot{German}       & \rot{Indic Languages} & \rot{Other} & \rot{Dataset}      & \rot{Method/Model}      &
 \rot{Evaluation/Metric} & \rot{Benchmark} & \rot{Dialect Classification} & \rot{Sentiment Analysis} & \rot{Machine Translation} & \rot{Morphology/Parsing} & \rot{Conversational AI} & \rot{Summarisation} & \rot{Speech/Visual}\\ \midrule

\cite{nerbonne1997measuring}& \OK           &              &                    &     &              &                        &            &              &\OK                    &                 &\OK                        &            &              &                    &       &          &       \\
\cite{nerbonne2001computational}& \OK           &              &                    &   &                &                        &            &              &\OK                    &                 &\OK                        &            &              &                    &       &          &       \\
\cite{chiang2006parsing}&            &\OK              &                    &   &                &                        &            &\OK              &                    &                 &                        &            &              &\OK                    &       &          &       \\
\cite{habash2006magead}&            & \OK             &                    &    &               &                        &            &\OK              &                    &                 &                        &            &\OK              &                    &       &          &       \\
\cite{chitturi-hansen-2008-dialect}& \OK           &              &     &                 &                 &                        &            &\OK              &                    &                 &\OK                        &            &              &                    &       &          &       \\
\cite{paul2011dialect}& \OK          &   &              & \OK                   & \OK                & \OK                       &            &\OK              &                    &                 &                        &            &\OK              &                    &       &          &       \\
\cite{lui2013classifying}& \OK           &  &              &                    &                 &                        &            & \OK             &                    &                 &\OK                        &            &              &                    &       &          &       \\
\cite{abdul2014sana}&          &    & \OK             &                    &                 &                        &\OK            &              &                    &                 &                        &\OK            &              &                    &       &          &       \\
\cite{cotterell2014multi}& &             & \OK             &                    &                 &                        &\OK            &              &                    &                 &\OK                        &            &              &                    &       &          &       \\
\cite{darwish-etal-2014-verifiably}&  &            & \OK             &                    &                 &                        &            &\OK              &                    &                 &\OK                        &            &              &                    &       &          &       \\
\cite{dogruoz-nakov-2014-predicting}&  &            &              &                    &                 &\OK                        &            &\OK              &                    &                 &\OK                        &            &              &                    &       &          &       \\
\cite{estival-etal-2014-austalk}& \OK       &      &              &                    &                 &                        &\OK            &              &                    &                 &                        &            &              &                    &       &          &\OK       \\
\cite{jeblee-etal-2014-domain}& \OK    &         &\OK              &                    &                 &                        &            &\OK              &                    &                 &                        &            &\OK              &                    &       &          &       \\
\cite{zampieri2014report}& \OK     &        &              &                    &                 &\OK                        &\OK            &              &                    &\OK                 &\OK                        &            &              &                    &       &          &       \\
\cite{jorgensen-etal-2015-challenges}& \OK  &           &              &                    &                 &                        &            &              & \OK                   &                 &                        &            &              &\OK                    &       &          &       \\
\cite{xu-etal-2015-building}&       &    \OK   &              &                    &                 &\OK                        &\OK            &              &                    &                 &                        &            &              &\OK                    &       &          &       \\
\cite{zampieri2015overview}& \OK   &          &              &                    &                 &                        &\OK            &              &                    &\OK                 &\OK                        &            &              &                    &       &          &       \\
\cite{ali2016botta}&       &       &\OK              &                    &                 &                        &            &\OK              &                    &                 &                        &            &              &                    &\OK       &          &       \\
\cite{blodgett-etal-2016-demographic}& \OK &            &              &                    &                 &                        &            &\OK              &                    &                 &\OK                        &            &              &                    &       &          &       \\
\cite{burghardt-etal-2016-creating}&     &         &              & \OK                   &                 &                        &\OK            &              &                    &                 &\OK                        &            &              &                    &       &          &       \\
\redtext{\cite{eskander2016creating}} & & & \OK & & & & & \OK & & & & & \OK & & & & \\
\cite{goutte2016discriminating}& \OK     &        &              &                    &                 &\OK                        &            &              &\OK                    &                 &\OK                        &            &              &                    &       &          &       \\
\cite{malmasi2016discriminating}& \OK     &        &              &                    &                 &\OK                        &            &              &\OK                    &                 &\OK                        &            &              &                    &       &          &       \\
\cite{azouaou2017alg}&    &          &              &                    &                 &\OK                        &\OK            &              &                    &                 &\OK                        &            &              &                    &       &          &       \\
\cite{bowers-etal-2017-morphological}&   &           &              &                    &                 &\OK                        &            &\OK              &                    &                 &                        &            &              & \OK                   &       &          &       \\
\cite{criscuolo2017discriminating}& \OK   &          &              &                    &                 &\OK                        &            &              &\OK                    &                 &\OK                        &            &              &                    &       &          &       \\
\cite{hassan-etal-2017-synthetic}& \OK   &          &              &                    &                 &\OK                        &\OK            &\OK              &                    &                 &                        &            &\OK              &                    &       &          &       \\
\cite{jurgens-etal-2017-incorporating}& \OK &            &              &                    &                 &\OK                        &\OK            &\OK              &                    &                 &\OK                        &            &              &                    &       &          &       \\
\cite{mdhaffar2017sentiment}&        &      &\OK              &                    &                 &                        &            &\OK              &                    &                 &                        &\OK            &              &                    &       &          &       \\
\cite{simaki-etal-2017-identifying}& \OK  &           &              &                    &                 &                        &            &\OK              &                    &                 &\OK                        &            &              &                    &       &          &       \\
\cite{abdul2018you}&      &        &\OK              &                    &                 &                        &\OK            &              &                    &                 &\OK                        &            &              &                    &       &          &       \\
\cite{assiri2018towards}&      &        &\OK              &                    &                 &                        &            &\OK              &                    &                 &                        &\OK            &              &                    &       &          &       \\
\cite{blodgett-etal-2018-twitter}& \OK   &          &              &                    &                 &                        &            &\OK              &                    &                 &                        &            &              &\OK                    &       &          &       \\
\cite{darwish2018multi}&        &      &\OK              &                    &                 &                        &            &\OK              &                    &                 &                        &            &\OK              &                    &       &          &       \\
\redtext{\cite{erdmann2018addressing}} & & & \OK & & & & & \OK & \OK & & & & & & & & \\
\cite{elmadany2018arsas}&    &          &\OK              &                    &                 &                        &\OK            &              &                    &                 &                        &\OK            &              &                    &       &          &\OK       \\
\cite{elmadany2018improving}&    &          &\OK              &                    &                 &                        &            &\OK              &                    &                 &                        &            &              &                    &\OK       &          &       \\
\redtext{\cite{salameh2018fine}} & & & \OK & & & & \OK & & & & & \OK & & & & &  \\
\cite{baly2019arsentd}&   &           &\OK              &                    &                 &                        &\OK            &              &                    &                 &                        &\OK            &              &                    &       &          &       \\
\cite{fadhil2019ollobot}&     &         &\OK              &                    &                 &                        &            &\OK              &\OK                    &                 &                        &            &              &                    &\OK       &          &       \\
\cite{joukhadar2019arabic}&       &       &\OK              &                    &                 &                        &            &\OK              &                    &                 &                        &            &              &                    &\OK       &          &       \\
\cite{mulki2019syntax}&         &     &\OK              &                    &                 &                        &            &\OK              &                    &                 &                        &\OK            &              &                    &       &          &       \\

\cite{sap2019risk}& \OK       &      &              &                    &                 &                        &            &              &\OK                    &                 &                        &\OK            &              &                    &       &          &       \\
\cite{zampieri-etal-2019-report}& \OK &            &              &\OK                    &                 &\OK                        &\OK            &              &                    &\OK                 &\OK                        &            &              &\OK                    &       &          &       \\
\cite{ahmed2020design}&       &       &              &                    &                 &\OK                        &            &\OK              &                    &                 &                        &            &              &                    &\OK       &          &     \\
\cite{al2020nabiha}&       &       &\OK              &                    &                 &                        &            &\OK              &                    &                 &                        &            &              &                    &\OK       &          &       \\
\cite{alshareef2020seq2seq}&    &          &\OK              &                    &                 &                        &            &\OK              &                    &                 &                        &            &              &                    &\OK       &          &       \\
\cite{demszky2020learning}&  &            &              &              &\OK                 &                        &            &\OK              &   \OK                 &                 & \OK                   &           &              &                    &       &          &       \\
\cite{dunn-adams-2020-geographically}& \OK         &    &              &                    &                 &\OK                        &\OK            &              &                    &\OK                 &\OK                        &            &              &                    &       &          &       \\
\cite{hanani2020spoken}&    &          &\OK              &                    &                 &                        &            &\OK              &                    &                 &\OK                        &            &              &                    &       &          &       \\
\cite{hou2020classification}&     &    \OK     &              &                    &                 &\OK                        &            &\OK              &                    &                 &\OK                        &            &              &                    &       &          &       \\
\cite{mozafari2020hate}& \OK     &        &              &                    &                 &                        &            &\OK              &                    &                 &                        &\OK            &              &                    &       &          &       \\
\redtext{\cite{tan2020mind}} & \OK   &          &              &                    &                 &                        &            &\OK              &                    &                 &                        &            & \OK           \OK  &                    &       &          &       \\ 
\cite{zhao-etal-2020-semantic}& \OK   &          &              &                    &                 &                        &            &\OK              &                    &                 &                        &            &              &\OK                    &       &          &       \\
\cite{ball2021differential}& \OK      &       &              &                    &                 &                        &            &\OK              &                    &                 &                        &\OK            &              &                    &       &          &       \\
\cite{ben2021multilingual}&      &        &              &                    &                 &\OK                        &            &\OK              &                    &                 &                        &            &              &                    &\OK       &          &       \\
\cite{boujou2021open}&      &        &\OK              &                    &                 &                        &\OK            &              &                    &                 &\OK                        &\OK            &              &                    &       &          &       \\
\cite{el-mekki-etal-2021-domain}&     &         &\OK              &                    &                 &                        &            &\OK              &                    &                 &                        &\OK            &              &                    &       &          &       \\
\cite{guellil-etal-2021-one}&      &        &\OK              &                    &                 &                        &\OK            &\OK              &                    &                 &                        &\OK            &              &                    &       &          &       \\
\cite{keswani2021dialect}& \OK     &        &              &                    &                 &                        &            &\OK              &                    &                 &                        &            &              &                    &       &\OK          &       \\
\cite{kumar-etal-2021-machine}& \OK   &          &\OK              &                    &                 &\OK                        &            &\OK              &                    &                 &                        &            &\OK              &                    &       &          &       \\
\cite{zhang2021disentangling}& \OK   &          &              &                    &                 &                        &            &\OK              &                    &                 &                        &\OK            &              &                    &       &          &       \\
\cite{chow2022singlish}& \OK     &        &              &                    &                 &                        &            &\OK              &                    &                 &                        &            &              & \OK                   &       &          &       \\
\cite{coats-2022-corpus}& \OK    &         &              &                    &                 &                        &\OK            &              &                    &                 &                        &            &              &                    &       &          & \OK      \\
\cite{eggleston-oconnor-2022-cross}& \OK  &           &              &                    &                 &                        &            &\OK              &                    &                 &                        &            &              &\OK                    &       &          &       \\
\cite{fuad2022araconv}&      &        &\OK              &                    &                 &                        &            &\OK              &                    &                 &                        &            &              &                    &\OK       &          &       \\
\cite{harris2022exploring}& \OK    &         &              &                    &                 &                        &            &\OK              &                    &                 &                        &\OK            &              &                    &       &          &       \\
\cite{husain2022weak}&      &        &              &                    &                 &\OK                        &            &\OK              &                    &                 &                        &\OK            &              &                    &       &          &       \\
\cite{inoue-etal-2022-morphosyntactic}&   &           &\OK              &                    &                 &                        &            &\OK              &                    &                 &                        &            &              &\OK                    &       &          &       \\
\cite{kanjirangat2022early}&     &         &\OK              &\OK                    &\OK                 &                        &            &              &\OK                    &                 &                        &\OK            &              &                    &       &          &       \\
\cite{kaseb2022saids}&      &        &\OK              &                    &                 &                        &            &\OK              &                    &                 &                        &\OK            &              &                    &       &          &       \\
\cite{kasen-etal-2022-norwegian}&    &          &              &                    &                 &\OK                        &            &\OK              &                    &                 &                        &            &              &\OK                    &       &          &       \\
\cite{liu-etal-2022-singlish}& \OK    &         &              &                    &                 &\OK                        &            &\OK              &                    &                 &                        &            &\OK              &                    &       &          &       \\
\cite{lu-etal-2022-exploring}&      &   \OK     &              &                    &                 &\OK                        &\OK            &\OK              &                    &                 &\OK                        &            &              &                    &       &          &       \\
\cite{okpala2022aaebert}& \OK     &        &              &                    &                 &                        &            &\OK              &                    &                 &                        &\OK            &              &                    &       &          &       \\
\cite{olabisi2022analyzing}& \OK    &         &              &                    &                 &                        &\OK            &              &                    &                 &                        &            &              &                    &       &\OK          &       \\
\cite{rajai2022dealing}& \OK      &       & \OK             &                    &                 &                        &            &              &\OK                    &                 &                        &            &\OK              &                    &       &          &       \\
\cite{saadany-etal-2022-semi}&    &          &\OK              &                    &                 &\OK                        &            &              &\OK                    &                 &                        &            &\OK              &                    &       &          &       \\
\cite{artemova-plank-2023-low}&    &          &              &\OK                    &                 &                        &\OK            &              &\OK                    &                 &                        &            &\OK              &                    &       &          &       \\
\cite{held-etal-2023-tada}& \OK    &         &              &                    &                 &                        &            & \OK             &\OK                    &                 &                        &            &              &                    &       &          &       \\
\cite{kantharuban-etal-2023-quantifying}&          &    &              &\OK                    &                 &\OK                        &            &              &\OK                    &                 &                        &            &              &                    &       &          &       \\
\cite{kuparinen-etal-2023-dialect}&      &        &              &\OK                    &                 &\OK                        &            &              &\OK                    &                 &                        &            &              &                    &       &          &       \\
\cite{lameli2023measure}&    &          &              &\OK                    &                 &                        &            &\OK              &                    &                 &\OK                        &            &              &                    &       &          &       \\
\cite{le-luu-2023-parallel}&     &         &              &                    &                 &\OK                        &\OK            &              &                    &                 &                        &            &              &                    &       &          &       \\
\cite{lent2023creoleval}&      &        &              &                    &                 &\OK                        &\OK            &              &\OK                    &\OK                 &\OK                        &\OK            &              &                    &       &          &       \\
\cite{maurya2023utilizing}&    &          &              &                    &\OK                 &                        &            &\OK              &                    &                 &                        &            &\OK              &                    &       &          &       \\
\cite{pluss-etal-2023-stt4sg}&      &        &              &\OK                    &                 &\OK                        &\OK            &              &                    &                 &                        &            &              &                    &       &          &\OK       \\
\cite{ramponi2023diatopit}&     &        &              &                    &                 &\OK                        &\OK            &              &                  &                 &\OK                        &            &              &                    &       &          &       \\
\cite{riley-etal-2023-frmt}& \OK  &     \OK      &              &                    &                 &\OK                        &            &\OK              &\OK                    &                 &                        &\OK            &              &                    &       &          &       \\
\cite{zhan2023socialdial}&   & \OK          &              &                    &                 &                        &\OK            &              &   \OK                 &\OK                 &                        &            &              &                    &\OK       &          &       \\
\redtext{\cite{zhan2024renovi}} &   & \OK           &              &                    &                 &                        & \OK            &              &  \OK                  &\OK                 &                        &            &              &                    &\OK       &          &       \\
\redtext{\cite{artemova2024exploring}} &   &          &              & \OK                   &                 &                        & \OK            &              &  \OK                  &\OK                 &                        &            &              &                    &\OK       &          &       \\
\cite{ziems-etal-2023-multi}& \OK &            &              &                    &                 &                        &\OK            &              &                    &\OK                 &                        &            &              &                    &       &          &       \\
\greentext{\cite{ahia-etal-2024-voices}}&  &            &              &                    &                 & \OK                       & \OK           &              &                    &  \OK               &                        &            & \OK             &                    &       &          & \OK      \\
\greentext{\cite{talafha-etal-2024-casablanca}}&  &            &\OK              &                    &                 &                        & \OK           &              &                    &  \OK               & \OK                     &            &            &                    &       &          & \OK      \\
\greentext{\cite{dinh-etal-2024-multi}}&  &            &              &                    &                 & \OK                       & \OK           &              &                    &                &  \OK                   &            &            &                    &       &          & \OK      \\
\greentext{\cite{vidal-gorene-etal-2024-cross}}&    &            &              &                    &                 &  \OK               &     \OK       &              &                    &  \OK           &                  &            &            &  \OK      &       &          &     \\
\greentext{\cite{dabre-etal-2024-machine}}&    &            &              &                    & \OK          &                &     \OK       &              &                    &  \OK           &                  &            & \OK            &       &       &          &     \\
\bottomrule
\end{tabular}}
\caption{\label{tab:bigtable}State of NLP research on Dialects.}
\end{table*}

\subsection{Trends}
\label{sec:trends}
\redtext{Table~\ref{tab:bigtable}} summarises the papers covered in this survey. We identify three trends in the past work:
\begin{enumerate}
  \item \textbf{Tasks in focus}: Older research dealt with \redtext{dialectal} datasets primarily for dialect classification. Past work shows performance degradation when the text contains dialects of a language as compared to the predominant (\textit{i.e.}, standard) form.
  \item \textbf{Languages in focus}: The papers reporting work on dialects of Arabic are significantly \redtext{more} than those for dialects of other languages. This has also been accelerated by research forums focusing on Arabic NLP. While the work in English is predominantly for \redtext{the} African-American dialect of English, recent papers examine other dialects such as Indian English, Singaporean English and so on.
  \item \textbf{Mitigation is more than perturbation}: Modifying a sentence or its representation to or from its \redtext{dialectal} variations has been achieved by perturbation techniques of varying complexity. However, recent papers show that dialect mitigation can be integrated into the model architecture itself using adversarial networks~\cite{ball2021differential}, hypernetworks~\cite{xiao2023task}, etc.
\end{enumerate}

It may seem that NLP for dialects of a language only pertains to datasets, \textit{i.e.}, it does not need any specialised handling beyond the introduction of a new dataset. However, we observe that the adaptation of NLP techniques for dialects operates at several points in a typical NLP pipeline:
\begin{enumerate}
\item \textbf{Training resources}: Labeled datasets \orangetext{(including treebanks)} and lexicons in dialects of a language have been reported in the past. This includes datasets with dialect labels along with additional task-specific labels, where the task is an NLP research problem.
\item \textbf{Models}: Models have been enhanced with several techniques, as may be typical of the time of the research. The fact that dialect-aware NLP can benefit from model adaptations and not dataset replacement alone is a key point of the survey.
\item \textbf{Evaluation datasets}: NLP techniques evaluated on datasets in dialects have peculiar observations. Language identification classifiers produce lower performance when the text is in a dialect of a language. The performance of LLMs on \redtext{dialectal} datasets is positively correlated with socio-economic factors.
\end{enumerate}

\begin{figure}
  \centering
  \includegraphics[width=0.6\textwidth]{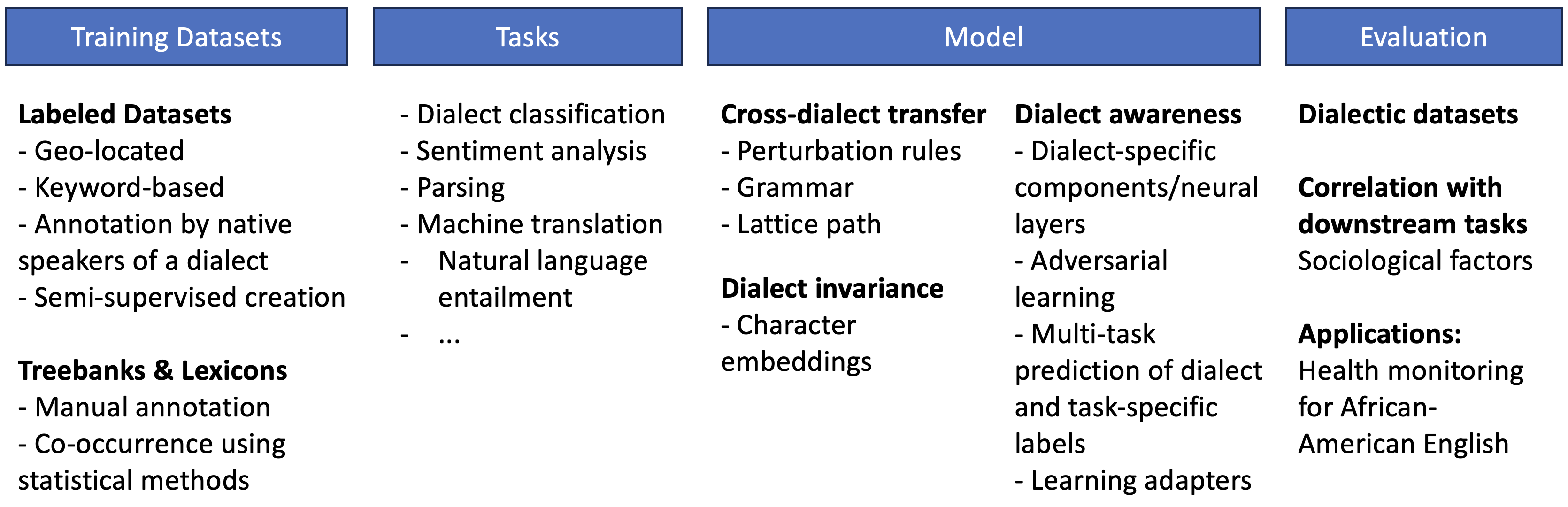}
  \caption{An Overview of Approaches in terms of NLP for Dialects.}
  \label{fig:overview1}
\end{figure}

Figure~\ref{fig:overview1} shows an overview of the approaches in terms of NLP for dialects. There have been different approaches to create labeled datasets, tree-banks and lexicons. In terms of models, past work varies in terms of NLP tasks and the way \redtext{dialectal} adaptation is handled: dialect transformation (where data is translated between dialects for the purpose of processing), dialect invariance (where models are made invariant to dialects) and dialect awareness (where models include dialect-specific components). Finally, we also describe \redtext{dialectal} datasets and resultant evaluations on downstream tasks including applications such as health monitoring. 

\section{Resources}
Being a data-driven field, NLP techniques rely on resources such as lexicons and textual datasets. In this section, we describe ways in which \redtext{dialectal} datasets have been created.
\label{sec:resources}
\subsection{\redtext{Dialectal} Lexicons}
\redtext{Dialectal} lexicons correspond to word lists or word mappings about a dialect. \redtext{Although lexicons were popular in early approaches of NLP, a recent paper by ~\citet{artemova-plank-2023-low} highlights the potential of dialectal lexicons and describes an approach to create such lexicons using large language models. Prior to this, research} in the creation of \redtext{dialectal} lexicons lies in three categories: the use of online dictionaries, and the use of textual corpora.

\subsubsection{Online dictionaries}
~\citet{azouaou2017alg} create a lexicon of words mapping French and its Algerian dialect. They use online dictionaries along with a combination of manual and automatic methods to enhance the lexicon. This includes many-to-one mapping of words in the two sets. Similarly, ~\citet{boujelbane-etal-2013-building} build bilingual lexicons to create Tunisian \redtext{dialectal} corpora to adapt n-gram models for statistical machine translation.

\subsubsection{Textual corpora}
~\citet{abdul2014sana} present a lexicon of words in dialects of Arabic including Levantine and Egyptian dialects. They present two lexicons: adjectives \redtext{in} news articles and common words in online chat forums. The words are labeled with a combination of manual and automatic techniques, the latter based on statistical techniques such as pointwise mutual information. Similarly,~\citet{burghardt-etal-2016-creating} use the web as a corpus to create a lexicon for the Bavarian dialect of German. Starting with a corpus of Facebook comments, they provide a rule-based algorithm to create the lexicon. They first extract unique text forms, and then filter non-dialect words based on the Dortmund chat corpus. ~\citet{younes2020language,harrat2018maghrebi} discuss various existing resources for the Maghrebi Arabic dialects (MAD) including annotated corpora for language identification, and morpho-syntactic analysis. MAD include principally Algerian Arabic, Moroccan Arabic and Tunisian Arabic.

\subsubsection{\orangetext{Lexicon Induction Using LLMs}}
\orangetext{A recent approach by ~\citet{artemova-plank-2023-low} performs German dialect lexicon induction using LLMs.}
\subsection{\redtext{Dialectal} Datasets}
\redtext{Datasets based on different} data sources (such as social media, and conversation transcripts) and dialects have been reported. In terms of procuring and labeling these datasets, the following methods have been used:
\subsubsection{Recruit native speakers \bluetext{of specific language varieties}}~\citet{estival-etal-2014-austalk} create a dataset of audio-visual recordings of 1000 speakers of Australian English. The dataset is accompanied by a transcript, which was manually created for 100 speakers. \bluetext{~\citet{bouamor2018madar} \orangetext{present MADAR}: a manually curated parallel corpus of sentences in Arabic dialects along with English, French and Modern Standard Arabic. \footnote{\bluetext{~\citet{obeid2019adida} is a demonstration based on the dataset.}}} Similarly,~\citet{riley-etal-2023-frmt} create a parallel corpus of English sentences and two dialects each of Portuguese and Chinese with the help of native speakers of these dialects. \bluetext{~\citet{eisenstein23_interspeech} introduce MD-3 a dataset of conversations between speakers playing the game of taboo. The dataset consists of speech recordings as well as text transcripts.} \bluetext{\citet{seddah-etal-2020-building} focus on treebank creation for Algerian supplemented with monolingual data obtained from CommonCrawl. They highlight the inherent difficulty of finding annotators and the cost of the same indicating the challenges for \redtext{dialectal} data generation. \citet{riabi-etal-2023-enriching} further extend this with additonal layers of morpho-syntactic knowledge and correct errors in the same.}

\subsubsection{Perturbation}~\citet{ziems-etal-2022-value} evaluate natural language understanding for African-American English. They design rules to perturb the dataset from Standard American English to African-American English. They then get them validated by native speakers.~\citet{ziems-etal-2023-multi} present Multi-VALUE, a suite of resources to evaluate fairness of LLMs by creating \redtext{dialectal} variations of a dataset. The suite provides mechanisms to generate 50 dialects of English by applying a set of perturbations.\greentext{~\citet{messner-lippincott-2024-examining} present a dataset of 19th century American literary orthovariant tokens with a novel layer of human-annotated dialect group tags, to examine language modelling assumptions, and find evidence that choice of tokenization scheme meaningfully impact the type of orthographic information in a language model.}

\subsubsection{Keywords}~\citet{wang-etal-2017-universal} create a dataset of Singaporean English sentences by searching for typical Singaporean English terms in online forums. ~\citet{ramponi2023diatopit} take a \textit{complementary} approach to create a dataset of tweets in dialects of Italian \bluetext{along with other languages spoken in Italy (which are not necessarily derived from Italian)}. They use location-based search to obtain the set of tweets from different regions of interest from within Italy. Following this, they use out-of-vocabulary words to identify words that are indicative of geographical regions and, as a result, dialects. \bluetext{A related dataset is GeoLingIt~\cite{ramponi-casula-2023-geolingit}}. In the case of social media, \textit{hashtags} can be used to obtain datasets in certain dialects. \redtext{~\citet{kuparinen-2023-murreviikko}} take advantage of dialect awareness week in Finland. They use a hashtag indicating usage of dialects in order to collect tweets in different dialects of Finnish. \redtext{In the context of Arabic dialect tweets, ~\citet{boujou2021open} benchmark is a novel dataset of ~50,000 tweets for five dialects of Arabic-Algerian, Lebanon, Morocco, Tunisian, and Egyptian.} 

\subsubsection{Location} Data from particular geographics can be extracted using filters (where the location is known) or inference (where it is not known). 
~\citet{jurgens-etal-2017-incorporating} use location-based filters available on Twitter at the time. They use language identification classifiers to predict the language and identify \redtext{dialectal} users.~\citet{husain2022weak} obtain tweets from Kuwait to create a dataset of tweets in the Kuwaiti dialect of Arabic. ~\citet{coats-2022-corpus} create an unlabeled dataset of Youtube comments. They start with a list of councils in Australia, extract official Youtube channels and retrieve comments. They manually validate the correctness of the channels. When working with geographically dispersed dialects, \textit{sampling} may also be used.\redtext{~\citet{hovy-purschke-2018-capturing} use Doc2Vec on a large corpus of anonymous online posts to learn document representation of cities, and recover dialect areas using geographic information via retrofitting and agglomerative clustering.}~\citet{dunn-adams-2020-geographically} create a Web-based corpus in different dialects by sampling sentences from different countries. The goal is to build a Web-based \orangetext{corpus} where the number of instances \redtext{is} reflective of the population of speakers in a country. The paper states that such a geography-aware corpus can lead to geography-aware representations when language models are trained on them. \orangetext{A criticism to the location-based filtering mechanism is by ~\citet{goutte2016discriminating}.}
\subsubsection{Dialect-aware annotation}\orangetext{ One} such example is by~\citet{sap2019risk}. They examine racial bias towards African-American English in the case of hate speech detection. They propose race and dialect priming in order to improve the quality of annotation. In order to prime the annotators, they propose to ask two questions: (a) is the tweet offensive to them\redtext{?}, and (b) is the tweet offensive to anyone\redtext{?} The dialect and race of the speaker are shown to the annotators.

\bluetext{Several datasets exist for varieties of Arabic~\cite{diab2010colaba}, including Palestinian Arabic~\cite{jarrar2017curras, dibas2022maknuune}, Gulf Arabic~\cite{khalifa2016large}, Egyptian Arabic~\cite{maamouri2014developing}, and Bahraini Arabic~\cite{abdulrahim2022bahrain}. This is in stark contrast with the lack of availability of datasets for dialects of English or several other languages of the world.}
\section{Natural Language Understanding (NLU)}
\label{sec:nlu}
This section covers NLU approaches centered around dialects. This includes approaches for NLP tasks such as dialect identification, sentiment analysis, morphosyntactic analysis and parsing. We also describe approaches reported on NLU benchmarks\orangetext{,} which cover multiple tasks.

\subsection{Dialect Identification}
The most commonly researched task in the scope of this paper is dialect identification. Dialect identification deals with the prediction of the dialect of an input text. Early approaches to dialect identification employed distance-based metrics, namely, Levenshtein, Manhattan, and Euclidean distance with \redtext{different clustering techniques}~\cite{nerbonne1997measuring,nerbonne2002computational}. They indicate that feature representations are more sensitive, and that Manhattan distance and Euclidean distance are good measures of phonetic overlap.~\citet{elnagar2021systematic} is a systematic review of identification of dialects of Arabic. For dialects of Arabic, lexical resources such as lexicons and \bluetext{treebanks}, and models using SVM or sequential neural layers like BiLSTM have been reported.~\citet{jauhiainen2019automatic} is a survey of automatic language identification. They describe that dialect detection may be more difficult than language detection since dialects may have lexical or syntactic overlap. In doing so, the survey does not make a distinction between languages and dialects - and treats different dialects as different class labels, while still maintaining a classification approach. However, one sees challenges in this regard.
\redtext{~\citet{boujou2021open} present a baseline approach\orangetext{,} which utilises classical machine learning.} \bluetext{While the majority of past work defines dialect identification as a Boolean/multi-class classification, ~\citet{baimukan2022hierarchical} use a hierarchy of dialect labels based on geographical and linguistic proximity}. We now describe details of past work in dialect identification in terms of shared tasks, datasets, and pre-deep learning and deep learning-based approaches. 

\begin{table}[]
\begin{tabular}{lp{8cm}}
\toprule
\orangetext{Shared Task}                                    & \orangetext{Dialects/Languages}  \\ \midrule
\orangetext{\cite{zampieri2014report}}        & \orangetext{Brazilian Portuguese and European Portuguese; American and British English; and Argentinian Spanish and Castilian Spanish} \\
\orangetext{\cite{zampieri2015overview}} & \orangetext{American and British English; and Argentinian Spanish and Castilian Spanish} \\ 
\orangetext{\cite{malmasi2016discriminating}} & \orangetext{Dialects of English, Spanish, French and Arabic}\\ 
\orangetext{\cite{zampieri-etal-2019-report}} & \orangetext{Dialects of German, Chinese, Romanian}\\
\orangetext{\cite{gaman-etal-2020-report}} & \orangetext{Dialects of Romanian, Geolocation-based Varieties} \\
\orangetext{\cite{aepli-etal-2022-findings}} & \orangetext{Dialects of French and Italian} \\
\orangetext{\cite{aepli-etal-2023-findings}} & \orangetext{Dialects of Indo-European and Ural languages (and other tasks)} \\
\orangetext{\cite{abdul-mageed-etal-2023-nadi}} & \orangetext{Dialects of Arabic} \\
\bottomrule
\end{tabular}
\caption{\orangetext{Shared tasks related to dialect identification.}\label{tab:sharedtasks}}
\end{table}
\subsubsection{Shared tasks} Shared tasks have accelerated past work in dialect identification. These have been primarily led by Workshop on NLP for Similar Languages, Varieties and Dialects, also known as VarDial. \orangetext{The shared tasks are listed in Table~\ref{tab:sharedtasks}. ~\citet{goutte2016discriminating} summarise the findings from the past versions of the shared task from 2014-2016. Similarly, the Nuanced Arabic Dialect identification (NADI) shared task is held annually. In the 2024 edition of NADI, a new task has been introduced to estimate the Arabic level of \textit{dialectness} within Arabic sentences.}

\subsubsection{Datasets}
~\citet{aji-etal-2022-one} report a dataset for several languages and dialects spoken in Indonesia. They observe that language identification works well for certain dialects (Ngoko-Central dialect of Javanese, for example). The paper also discusses code-mixing and orthography variations in these languages.~\citet{dunn-2019-modeling} reports dialect identification on 14 national varieties of English. \redtext{He shows} that cross-domain classification (CommonCrawl versus Twitter) also performs poorly.~\citet{cotterell2014multi} present a multi-dialect, multi-genre corpus of news comments and tweets written in dialects of Arabic. The tweets are manually annotated for dialect identification on MTurk.~\citet{ramponi2023diatopit} present a benchmark dataset for dialects of Italian. The benchmark is named as DIATOPIT. There has been recent work on creating a corpus of Norwegian dialect~\cite{barnes-etal-2021-nordial}. Also,~\citet{alshutayri2018creating} present a large (200K+ instances) corpus for Arabic dialects and Standard Arabic. The data is sourced largely from tweets but also includes comments from newspapers, and Facebook. The data is also being annotated for dialect identification and contains 24K annotated documents. \greentext{Recently,~\citet{talafha-etal-2024-casablanca} introduced CASABLANCA, a large scale community-driven effort to collect and transcribe a multi-dialectal Arabic dataset, covering eight dialects for Arabic releasing $48$ hours of manually transcribed speech data including annotations for transcription, gender, dialect, and
code-switching.}~\citet{le-luu-2023-parallel} present a parallel corpus for dialects of Vietnamese. \greentext{Further,~\citet{dinh-etal-2024-multi} propose a dialect identification, and speech recognition dataset, and fine-tuned models for for $63$ provincial dialects of Vietnamese with $102.5$ hours of audio, and $19000$ spoken utterances.}

\subsubsection{Feature-based approaches}
We now highlight features used for dialect identification.
\begin{enumerate}
 \item \textbf{Phonological features}: \redtext{P}honological features are based on markers in the written scripts.~\citet{darwish-etal-2014-verifiably} use lexical \orangetext{along with a lexicon of dialectal Egyptian words}, morphological and phonological features in a random forest classifier to detect dialects of Arabic spoken in a geographical region. 
  \item \textbf{Linguistic features}:~\citet{dogruoz-nakov-2014-predicting} present a method to predict dialects of Turkish by using light verb constructions. They use a statistical classifier based on verb-based features (base word, verb order, affixes, etc.) for the task.\redtext{~\citet{xie-etal-2024-extracting} discuss an approach to extract distinguishing lexical features of dialects by utilising interpretable dialect classifiers. With focus on varieties of Mandarin, Italian, and Low Saxon, this approach shows promising results on all varieties.}
\end{enumerate}
The combinations of the above set of features have also been reported. While~\citet{hanani2020spoken} work on the detection of dialects from speech, they also use word-level n-gram features. \bluetext{\citet{salameh2018fine} perform fine-grained dialect identification for 25 dialects of Arabic, using Na\"{i}ve Bayes classifier and word and character n-grams as features.}

\redtext{While dialect detection of Arabic has been explored in detail, the} pre-deep learning work in the context of \redtext{dialects of} English \redtext{is comparatively limited, although English is the predominant language for NLP research}.~\citet{lui2013classifying} is an early work in the detection of dialects of English. Specifically, the paper focuses on Australian, British and Canadian English. Their baseline is the LangID classifier~\cite{lui-baldwin-2012-langid} where dialects are treated as individual languages. They experiment with classifiers using features such as n-grams and POS-n-grams. This includes a distribution over function words and those in a vocabulary, akin to a clustering algorithm.~\citet{simaki-etal-2017-identifying} use linguistic, POS-tag-based and lexicon-based features. 

\subsubsection{Deep learning-based Approaches}
Deep learning-based approaches for dialect classification span three alternatives: train embeddings to reflect \redtext{dialectal} variations, use end-to-end LLMs, or predict dialect as a result of inference over dialect features.

\noindent \textbf{Embeddings in focus}:~\citet{abdul2018you} label tweets with 10 dialects of Arabic. The city is considered the \redtext{dialectal} granularity. The analysis compares \redtext{dialectal} variants by looking at word embeddings of words across different dialects. They use word2vec representations to show how \redtext{dialectal} words are captured.~\citet{goswami2020unsupervised} build character-to-sentence embeddings to represent words of different dialects. Unsupervised loss is computed in order to generate clusters of representations. While they also test on language identification, the dialect identification part is done on Swiss German dialect.~\citet{jurgens-etal-2017-incorporating} use a character-based \textit{seq2seq} model to map dialects. The models used for language identification are RNNs with GRU.~\citet{criscuolo2017discriminating} use character n-grams to identify language groups. This is followed by convolutional neural network-based dialect classifiers for each language group.

\noindent \textbf{Fine-tuning LLMs}:~\citet{ramponi2023diatopit} experiment with multiple models including statistical and neural. The fine-tuned AlBERTo model performs the best among umBERTo, mBERT and XLM-R.~\citet{obeid2020camel} present CAMeL: a python toolkit for Arabic language processing. It contains a dialect identifier that gives a distribution over multiple dialects. They use \redtext{dialectal} guidelines provided in ~\citet{elfardy2012simplified}.

\noindent \textbf{Detecting dialect features}: \citet{d-eta-2021-learning} introduce an approach for dialect classification using a novel multi-task approach that employs dialect feature detection. They train two multi-task learning-based approaches using a small number of minimal pairs. They evaluate the output based on 22 \redtext{dialectal} features based on Indian English and demonstrate that such models show the capability of learning to identify features with high accuracy. They show the efficacy of this task by applying it \orangetext{to} dialect identification, and by providing a measure of dialect density.
\subsection{Sentiment Analysis}
Sentiment analysis is the NLU task of prediction of sentiment polarity of a text. Sentiment analysis encompasses several related tasks, such as sarcasm classification and target-specific sentiment analysis.  We discuss past work in sentiment analysis along four directions: experiences from annotation (which highlights the challenge of dialects for sentiment analysis), \orangetext{dialect-aware models, dialect-invariant models}, and\orangetext{,} finally\orangetext{,} de-biasing of sentiment analysis models as a post-processing step. Table~\ref{tab:sa1} summarises approaches for sentiment analysis.

\noindent \orangetext{\textbf{Early guessing for dialects}: A recent advancement in dialect identification is early guessing~\cite{kanjirangat2022early}. The approach detects a dialect for an incremental input. \redtext{~\citet{salloum2022unsupervised} also break the input down into its components. Specifically, they present an unsupervised approach that uses unsupervised dialect segmentation for machine translation.}}
\subsubsection{Datasets \& Annotation}
Several datasets in Arabic sentiment analysis for dialects have been reported such as Moroccan~\cite{oussous2020asa} and Levantine~\cite{baly2019arsentd}.
Dialects can have an impact on annotation itself.~\citet{farha2022effect} show that dialect familiarity helps sarcasm annotation.~\citet{mdhaffar2017sentiment} create a dataset of 17000 Facebook comments labeled with sentiment in Tunisian dialect of Arabic.~\citet{assiri2018towards} present a sentiment-labeled lexicon of words in the Saudi dialect of Arabic, and use simple counting-based sentiment analysis.~\citet{husain2022weak} use weakly supervised labels for sentiment analysis of tweets in the Kuwaiti dialect of Arabic. The labels are then manually validated and updated. 

\subsubsection{Dialect-aware representations} \redtext{Given the high degree of similarity between dialects, there is a high likelihood for models to make inferences in the same way for different dialects and thus explicitly modeling dialect awareness into models is important. However, this same degree of similarity makes this dialect aware modeling challenging.} \citet{farha2022effect} train BERT-based models for sarcasm detection on data annotated by either of the two groups: those familiar with the dialect and those not. They show that familiarity of dialect improves the quality of the models trained on such a dataset. As a result, representations that capture dialects have been used for sentiment analysis.~\citet{mdhaffar2017sentiment} present models based on SVM and multi-layer perceptron (MLP).~\citet{mulki2019syntax} use a syntax-ignorant n-gram composition to create embeddings. The classifier model is a dense neural network that works on the addition of word embeddings, with a softmax at the end.~\citet{guellil-etal-2021-one} propose `one' model for sentiment classification in different dialects of Arabic. They use transliteration to map dialects to Standard Arabic. The sentiment analysis model itself uses word2vec features with statistical classifier. Finally,~\citet{husain2022weak} present statistical models based on SVM along with Transformers-based models like BERT. 

\subsubsection{Incorporating dialect information in sentiment prediction}~\citet{el-mekki-etal-2021-domain} use domain adaptation for sentiment analysis of dialects. Using representations from a BERT encoder, they use two classifiers: sentiment classifier and dialect classifier. The output of the two is later combined for the overall prediction. While this is a two-channel approach, the representation used for the task has also been used to predict dialect of the language. One such example is~\citet{okpala2022aaebert} who present an approach for hate speech detection using African-American English. In order to do so, they re-train BERT with AAE tweets. Finally, adversarial training \redtext{is needed} to regulate the debiasing of the hate speech classifier. Specifically, the adversary takes the final representation learned by the hate-speech classifier, and learns to predict the dialect from it.~\citet{kaseb2022saids} present a dialect-aware approach for sarcasm detection called the SAIDS model. \orangetext{SAIDS} uses MARBERT to detect dialect and sarcasm. Following that, MARBERT, along with sarcasm and dialect output, are used to detect sentiment. Evaluated on \redtext{Arabic dialects}, \orangetext{SAIDS} uses backpropagation only for prediction with respect to the BERT base model. It does not flow through sentiment<->sarcasm or sentiment<->dialect. 

\subsubsection{De-biasing sentiment analysis models} Making sentiment analysis agnostic to dialects involves removing \redtext{dialectal} biases in the resultant models. A work of this nature is by~\citet{ball2021differential} who apply adversarial debiasing \redtext{to} resampled data for harmful tweet detection of tweets written in African-American English. Resampling of the data uses a metric for margin of confidence which selects the set of tweets that are most likely to be mis-classified. Adversarial debiasing involves training an adversary network to debias the classifier by including the adversary network's loss. Similarly,~\citet{mozafari2020hate} report results on hate speech detection from African-American and Standard American tweets. They re-weight instances based on the presence of phrases that may highlight racial bias. They fine-tune BERT for the task. Finally,~\citet{zhang2021disentangling} present an approach to reduce spurious correlation between two attributes: toxicity and African-American Vernacular English. They construct triplets of sentences where the first \redtext{two} have the same toxicity label, and the \redtext{first and the third} have the same dialect label. The objective function of the model consists of a triplet loss over these triplets, and a disentanglement loss that ensures the masks for the true attributes are well-separated. Similarly, graphical models have been used to infer socio-cultural norms since they are closely associated with \redtext{dialectal} variations based on the language and cultural background of the speaker.~\citet{moghimifar-etal-2023-normmark} present a Markov model to discover socio-cultural norms in emotion classification.\greentext{~\citet{harris-etal-2024-modeling} evaluate the zero-shot performance of speech recognition systems across different genders and across four US-based English dialects: SAE, AAVE, Chicano English, and Spanglish, release a labeled dataset of 13 hours of podcast audio, transcribed by speakers of the represented dialects.}\redtext{While past research has only dealt with African-American English, there may indeed be other dialects\orangetext{,} which are considered aggressive and may result in sentiment analyzers producing biased output. \orangetext{This is significantly underexplored for dialects of other languages such as the Khariboli (Haryanvi group) dialect of Hindi~\cite{yadav1974interactions}.}}

\begin{table}
\centering
\begin{tblr}{
  width = \linewidth,
  colspec = {Q[290]Q[248]Q[650]},
  hline{1,10} = {-}{0.08em},
  hline{2} = {-}{0.05em},
}
\textbf{Paper} & \textbf{Dialects}                 & \textbf{Modelling Approach}                                                                             \\
~\cite{mdhaffar2017sentiment}  & Dialects of Arabic       & SVM/MLP                                                                                        \\
~\cite{mulki2019syntax} & Dialects of Arabic       & {Syntax-ignorant composition to learn\\word embeddings}                                           \\
~\cite{mozafari2020hate} & African-American English & Re-weight instances based on racially biased phrases for hate speech detection                 \\
~\cite{el-mekki-etal-2021-domain} & Dialects of Arabic       & {Infer dialect and sentiment label\\using two channels from BERT encoder}                         \\
~\cite{guellil-etal-2021-one} & Dialects of Arabic       & Transliteration to map to standard version                                                     \\
~\cite{kaseb2022saids} & Dialects of Arabic       & {Infer dialect and sarcasm label; limited\\backpropagation to maintain label dependency} \\
~\cite{okpala2022aaebert} & African-American English & {Adversarial training\\to debias dialectic variation}                                            \\
~\cite{moghimifar-etal-2023-normmark} & English                  & {Socio-cultural norms are inferred\\using a Markov model variation}                               
\end{tblr}
\caption{Sentiment Analysis Approaches for \textbf{Dialectal} Datasets.}
\label{tab:sa1}
\end{table}



\begin{table}
\centering
\begin{tblr}{
  width = \linewidth,
  colspec = {Q[290]Q[260]Q[580]},
  hline{1,8} = {-}{0.08em},
  hline{2} = {-}{0.06em},
}
\textbf{Paper} & \textbf{Dialects}        & \textbf{Highlight}                        \\
~\cite{habash2006magead}  & Dialects of Arabic                   & Rewrite rules to adapt morph analysers    \\
~\cite{eskander2016creating} & \redtext{Dialects of Arabic} & Dataset proportions for improved performance \\
~\cite{jorgensen-etal-2015-challenges} & African-American English & POS taggers perform worse for the dialect \\
~\cite{darwish2018multi} & Arabic                   & CRF-based POS tagger; Linguistic features \\
~\cite{inoue-etal-2022-morphosyntactic} & Arabic                   & Fine-tuned LLMs                           \\
~\cite{bafna-etal-2023-cross} & Indic Languages          & Fine-tuned LLMs\\
\end{tblr}
\caption{Approaches for Morphological Analysis Focusing on Dialects. Highlight-> Approach or Key Finding.}
\label{tab:morphanal}
\end{table}

\subsection{Morphosyntactic analysis}
Morphosyntactic analysis deals with linguistic tasks such as POS tagging and morphological analysis, and \bluetext{has been found to be useful for sense disambiguation, particularly in low-resource settings~\cite{khalifa2020morphological}.} We now describe past work that deals with \redtext{dialectal} variations, as summarised in Table~\ref{tab:morphanal}.
\subsubsection{Classical approaches}
~\citet{habash2006magead} is a seminal morphological analyser for dialects of Arabic called MAGEAD. Using morphological rewrite rules, they show how a morphological analyser can be adapted for dialects of a language.~\citet{jorgensen-etal-2015-challenges} evaluate on a dataset of African-American Vernacular English and show that the then-prevalent POS taggers perform significantly worse.~\citet{darwish2018multi} present a CRF-based POS tagger for dialects of Arabic. The POS tagger is trained on a small set of tweets using features derived from the dialects of interest. These features are progressive and negation particles. \bluetext{~\citet{eskander2016creating} adapt existing morphological analyzers to unseen dialects of Arabic by simulating the low-resource dialects.}

\subsubsection{Deep learning-based approaches}

\citet{inoue-etal-2022-morphosyntactic} use CamelBERT trained on Modern Standard Arabic fine-tuned on dialect-specific datasets for morphosyntactic analysis. They observe that training using high-resource dialects helps low-resource dialects as well. In the context of Indic languages,~\citet{bafna-etal-2023-cross} explore POS tagging for 5 Indic dialects by focusing on Hindi-aware LLM adaptation via small \redtext{dialectal} monolingual corpora. \redtext{~\citet{aepli-sennrich-2022-improving} propose improving cross-lingual transfer between closely related language varieties from the Finnic, West and North Germanic, and Western Romance language branches using character-level noise injection, and go on to show consistent improvements for POS tagging. Their approach is further applied to seven languages from three families and a total of eighteen dialects~\cite{blaschke-etal-2023-manipulating} with results showing improvements by varying the  level of noise injected during the cross-lingual transfer.}

\subsection{Parsing}
Parsing involves the creation of \redtext{syntactic} parse trees from text. Past work in parsing texts written in dialects of a language lies in three categories. The first category uses an existing parser on a dataset in a dialect of interest. The focus of such work is to create a baseline performance of popular parsers. The second category provides approaches to \orangetext{adapt existing parsers} towards texts in the dialect of a language. The third category creates a new parser for the dialect.

\subsubsection{Use of existing parsers}~\citet{eggleston-oconnor-2022-cross} parse tweets in Standard American English and African-American English and use it to analyse social attributes of an entity, as per sentiment expressed in the tweets.~\citet{kasen-etal-2022-norwegian} create a tree bank of sentences in the Bokmål variety of Norwegian \redtext{dialects}. They present their results on the UUParser, an existing parser for Norwegian.~\citet{roy2020parsing} present an analysis using Stanford parser and Allen NLP parser on parsing of news headlines in Indian English.~\citet{scannell-2020-universal} create a treebank for Manx \redtext{Gaelic} and compare the performance of existing classifiers with Irish \redtext{Gaelic} and Scottish \redtext{Gaelic}.


\subsubsection{Adaptation of an existing parser}
~\citet{chiang2006parsing} show how parsing of Arabic dialects can be done by a sentence transduction approach. This approach parses the standardised version of a \redtext{dialectal} sentence, and then \redtext{links} it to the original sentence. The standardisation is achieved using transduction, akin to n-gram decoding. However,~\citet{blodgett-etal-2018-twitter} \redtext{use} neural networks and \redtext{present} an approach to dependency parsing for African-American English. This approach uses two neural parsers, which are modified with the word embeddings used for initialisation. The word embeddings are trained on the standard and the dialect-specific datasets. Further,~\citet{wang-etal-2017-universal} create a dependency parser for Singaporean English. This approach uses a base parser for standard English and stacks it with a series of BiLSTM layers known as the `feature stack' to extract relevant features, and an MLP with an output layer \redtext{to} help produce dependency-parsed output.~\citet{zhao-etal-2020-semantic} use a treebank of learner English sentences labeled with POS tags and dependency information. They propose a factorisation-based parser that first predicts nodes followed by edges in a dependency parse.~\citet{dou2023multispider} evaluates various parsers designed for converting text to SQL, focusing on a multilingual benchmark that covers dialects from seven different languages. This research is significant for its emphasis on semantic parsing, differing from the aforementioned dependency parsing works.

\subsubsection{Development of a new parser}:\redtext{~\citet{vaillant-2008-layered} propose a rule-based approach to construct a common syntactic description for a group of Creoles from Haiti, Guadeloupe, Martinique and French Guiana.}~\citet{bowers-etal-2017-morphological} present a finite-state machine-based parser for the endangered Odawa dialect of Ojibwe spoken in \redtext{Canada and northeastern United States}. This approach uses a phonological module composed \redtext{of} a morphological module where morphological strings are modified by the phonology until they match surface forms of the language.


\subsection{NLU Benchmarks}
Finally, benchmarks such as GLUE\orangetext{,} \redtext{which provide datasets for NLP tasks like \orangetext{semantic textual similarity prediction} (STS-B), \orangetext{sentiment classification (using the} Stanford sentiment treebank (SST-2)\orangetext{)}, natural language inference (NLI), textual entailment and so on, are an important part of language model evaluation pipeline.} \citet{ziems-etal-2022-value} show a drop in performance on 7 GLUE tasks including SST-2, S\redtext{T}S-B\orangetext{, when tested on dialectal English variations of the original Standard American English version}. \redtext{For example, for} SST-2, there is a 1.5-2\% drop using fine-tuned RoBERTa.

\begin{table}[]
\begin{tabular}{lll}
\toprule
\textbf{Paper}                       & \textbf{Dialects}                             &  \textbf{Approach}    \\ \midrule
~\cite{ziems-etal-2022-value} & African-American English & Perturbation to create variants\\
~\cite{dacon-etal-2022-evaluating} & African-American English & Adversarial learning \\
~\cite{held-etal-2023-tada} & Dialects of English & Contrastive loss, Morphosyntactic loss \\
~\cite{xiao2023task} & Dialects of English & Hypernetworks as LoRA adapters \\
\bottomrule
\end{tabular}
\caption{Dialect-aware approaches evaluated on NLU benchmarks.}
\label{tab:nlu1}
\end{table}

~\citet{dacon-etal-2022-evaluating} work with African-American English. They first propose CodeSwitch, a rule-based method of perturbing a sentence from Standard American English (SAE) to African-American English (AAE). They create perturbed versions of the dataset using CodeSwitch and manually evaluate it. They finally evaluate their method on NLI. In order to do so, they use adversarial learning that ensures that the predicted label is the same if either the \redtext{SAE} or \redtext{AAE} sentences are provided as the input. They refer to this as a disentanglement of language style. \bluetext{~\citet{tan2020mind} present base-inflection encoding: a mechanism to inject \redtext{dialectal} information into the encoder. They show that their encoding algorithm improves the performance of Vernacular African-American English for SQUAD and MNLI tasks.}

~\citet{held-etal-2023-tada} model natural language understanding for dialects as a dialect adaptation task. Using Multi-VALUE, they create African-American English variations of the GLUE benchmark (which is primarily written in Standard American English). Following that, they adapt a model pre-trained on Standard American English. To do so, they use: (a) a contrastive loss to ensure the representation of a standard sentence and its \redtext{dialectal} version is as close as possible; (b) a morphosyntactic loss based on word-level alignment between the standard and \redtext{dialectal} sentences. Their results show improved robustness on 4 dialects based on the GLUE benchmark.\greentext{~\citet{vidal-gorene-etal-2024-cross} provide a benchmark on lemmatization, POS-tagging, and morphological analysis for four Armenian varieties- Classical, Modern Eastern, Modern Western, and the under-documented Getashen dialect. They compare traditional RNN models, multilingual encoders, and large language models using supervised, transfer learning, and zero/few-shot learning approaches, and show how RNNs are strong at POS tagging, but LLMs handle unseen dialectal variations.}

A recent work by~\citet{xiao2023task} shows how low-rank adapters \orangetext{Low-Rank Adapters (LoRA) (a parameter-efficient fine-tuning or PEFT technique that allows fine-tuning LLMs faster by storing weight updates instead of updating all weights)} can use linguistic knowledge of dialects to improve zero-shot performance on NLU tasks. They integrate hypernetworks with LoRA adapters for dialect adaptation. Experts encode linguistic information in the form of feature vectors. A hypernetwork then learns to generate adapter weights for LoRA from the feature vectors. They demonstrate the impact of their fine-tuning approach on several GLUE tasks such as MNLI, RTE and so on. The dataset consists of variants of the GLUE benchmark for five dialects: African
American Vernacular English (AAVE), Indian English (IndE), Nigerian English (NgE), Colloquial Singaporean English (CollSgE), and Chicano English (ChcE). Similarly,~\citet{liu-etal-2023-dada} use dynamic aggregation of linguistic rules to adapt LLMs to multiple dialects. They first create a synthetic dataset of linguistic transformations using LLM probing. Following that, they train a set of feature adapters to generalise across multiple dialects of interest. They present their evaluation of multiple dialects of English.

\redtext{DIALECTBENCH~\cite{faisal2024dialectbenchnlpbenchmarkdialects} is a large-scale benchmark covering 10 NLP tasks focusing on 281 language varieties. Their evaluation shows substantial disparities in performance between the standard and non-standard language varieties, while also identifying language clusters with large performance divergence across tasks\orangetext{. Most} recently, the VarDial 2024 evaluation campaign~\cite{chifu-etal-2024-vardial} released dataset on the choice of plausible alternatives (COPA) task focusing on three micro-dialects namely, Cerkno dialect of Slovenian, Chakavian dialect of Croatian, and the Torlak dialect which is spoken across Serbia, Macedonia, and Bulgaria. This task requires a computational model to select one of two candidate statements which is more likely to be the cause or effect of a given premise statement. Collectively, training and test datasets from VarDial evaluation campaigns ($2014$ - $2024$) organised over the years should act as a good benchmark for LLM evaluation of dialects.}

\subsection{\bluetext{Others}}
\bluetext{\citet{erdmann2018addressing} investigate how word embeddings trained on dialect-specific or mixed-dialect corpora perform. In their experiments for text in dialects of Arabic, they show how dialect-specific embeddings can be helpful for dictionary induction. Dictionary induction here refers to alignment tables between dialects of a language. \citet{demszky2021learning} report models that predict dialect features using minimal pairs that represent linguistic properties of dialects. They do so for Indian English.}

\begin{table}
\centering
\begin{tblr}{
  width = \linewidth,
  colspec = {Q[285]Q[224]Q[490]},
  row{1} = {c},
  row{2} = {c},
  row{5} = {c},
  row{8} = {c},
  row{11} = {c},
  cell{2}{1} = {c=3}{0.941\linewidth},
  cell{5}{1} = {c=3}{0.941\linewidth},
  cell{8}{1} = {c=3}{0.941\linewidth},
  cell{11}{1} = {c=3}{0.941\linewidth},
  hline{1-3,5-6, 8-9, 11-13} = {-}{},
}
\textbf{Paper} & \textbf{Dialect} & \textbf{Approach}\\
\textbf{Summarisation} &  & \\
~\cite{olabisi2022analyzing} & {African-American /\\Hispanic English} & {Representations-based clustering \&\\ obtain summaries separately.}\\
~\cite{keswani2021dialect} & English Dialects & {Diversity-representative sentences \& \\weigh them for summary generation}\\
\orangetext{\textbf{Machine translation}} & & \\
\orangetext{~\cite{alam-etal-2024-codet}} & \orangetext{891 variations} & \orangetext{Contrastive dialect benchmark for MT evaluation} \\
\orangetext{~\cite{riley-etal-2023-frmt}} & \orangetext{Portuguese and Chinese Dialects \& English} & \orangetext{Few-shot approach for MT} \\
\orangetext{\textbf{Dialogue systems}} & & \\
\orangetext{~\cite{zhan2023socialdial}} & \orangetext{Dialectal norms in Chinese culture} & \orangetext{Benchmark dataset}\\
\orangetext{~\cite{artemova2024exploring}} & \orangetext{German dialects} & \orangetext{Perturbation-based evaluation for dialects}\\
\textbf{Text generation} &  & \\
~\cite{sun-etal-2023-dialect} & English Dialects & Towards dialect-robust metrics for text generation \\ 
\end{tblr}
\caption{Representative Examples for sequence-to-sequence NLP \orangetext{tasks}.}
\label{tab:seq2seq}
\end{table}


\section{Natural Language Generation (NLG)}
\label{sec:nlg}
The previous section showed that NLU for dialects has primarily focused on tasks like identification of dialects and sentiment analysis. We now present approaches in NLG.
NLG deals with sequence-to-sequence (\textit{seq2seq}) tasks in NLP\orangetext{,} which take a sequence as input and produce a sequence. \redtext{Challenges in the presence of dialects in a generation task can differ significantly given the task. The data and evaluation methods can be different for tasks, especially where dialectal text is being generated.} Some examples of such problems are summarisation, question answering and machine translation, and are described in Table~\ref{tab:seq2seq}. While the situation in the case of NLU was already dire, our survey indicates that for NLG, it is even worse. We will now discuss NLP approaches that deal with dialects of a language in the context of \textit{seq2seq} problems.

Two works reflect advances in the context of \textit{seq2seq} problems:
\begin{enumerate}
  \item \textbf{Making evaluation metrics dialect-aware}:~\citet{sun-etal-2023-dialect} state that metrics used to measure text generation may penalise outputs in certain dialects. They propose a metric named NANO\orangetext{,} which allows perturbations in the generated output. They show that models pretrained with NANO as the metric can be helpful for dialect-robustness.
  \item \textbf{Creating \redtext{dialectal} variants of datasets for benchmarking}:~\citet{ziems-etal-2023-multi} present Multi-VALUE, a library that creates \redtext{dialectal} variations of datasets based on a set of manually created rules. They create variants of benchmark datasets, and evaluate the variants for several \textit{seq2seq} tasks including machine translation, question answering and so on. The models for evaluation are based on modern LLMs such as BERT, ROBERTA, BART and T5. The library provides a useful resource as well as insights for dialect-aware benchmarking in the future.
\end{enumerate}
\subsection{Summarisation}
Past work in summarisation, although limited, states that dialect labels may not be explicitly necessary. However, a review of Arabic text summarisation by~\citet{elsaid2022comprehensive} state the use of ``dialect period frameworks'' to incorporate semantic information about dialects. In the case of multi-document summarisation, clustering of sentences in the input set is a predominant paradigm. Two such works are noteworthy: 
\begin{enumerate}
  \item~\citet{olabisi2022analyzing} analyse the diversity of dialects in multi-document summarisation of social media posts. They present a dataset that contains summaries of a collection \redtext{of} tweets written in three dialects: African-American English, Hispanic English, and White English. They use extractive summarisation using LONGFORMER-EXT and abstractive summarisation using BART and T5. In order to bring diversity-awareness in summarisation, they create automatic clusters of input documents based on semantic attributes. They follow a 2-stage approach where the summarisers are separately applied, and the resultant outputs are combined again using a summariser.
  \item~\citet{keswani2021dialect} examine the role of dialect diversity on multi-tweet summarisation. They use a variety of summarisers: typical traditional summarisers like TF-IDF, TextRank and LexRank, and SummaRunner (a neural summariser that treats summarisation as a sequence classifiation task). They create a control set: a subset of sentences that represent different dialects in the set of sentences. They introduce a bias mitigation procedure that introduces dialect-awareness in summaries using a parameter that is weighted to increase the score of dialect-diverse sentences in the dialect set.
\end{enumerate}

\subsection{Machine Translation}
Compared to summarisation, machine translation has been studied a bit more. 
Recent work is broadly divided into two categories: (i) translation between dialects of the same language, and (ii) translation between the dialect of a language and another language. In the rest of this section, we cover the approaches in these categories. 

\subsubsection{MT between dialects of the same language} The primary goal of inter-dialect translation is the dissemination of information available between a standard dialect and a non-standard one. In this context, the following works are relevant. Mapping from less used dialects to their most common versions is called \textbf{dialect normalisation}. One such work by \citet{kuparinen-etal-2023-dialect} provides a dialect normalisation dataset in Swiss German, Slovene, Finnish, and Norwegian. \bluetext{\citet{bouamor2014multidialectal} present a multi-dialectal dataset for various dialects of Arabic.}

\noindent \textbf{Harnessing pre-trained models:} \citet{le-luu-2023-parallel} show that models based on $BART_{pho}$ perform well for dialect normalisation in dialects of Vietnamese. This indicates that denoising-based pre-trained models can be a good source for dialect data generation owing to their infilling capabilities.

\noindent \textbf{Character level modeling:} ~\citet{abe-etal-2018-multi} conduct Japanese dialect translation where they use NMT to translate from dialect to standard Japanese using character RNN trained on small datasets collected as a part of their work.  ~\citet{honnet-etal-2018-machine} additionally suggest that normalisation is an important aspect for translating between Swiss German dialects, which is achievable via character-level models. \citet{kuparinen-etal-2023-dialect} further show that sliding-window-based approaches are useful since dialect translation does not need the entire sentence-level context.

\noindent\textbf{Perturbation-based regularisation}:~\citet{liu-etal-2022-singlish} present a \textit{seq2seq} approach for machine translation of Singaporean English to standard English. They use word perturbation and sentence perturbation to prevent overfitting of lexical features.~\citet{maurya2023utilizing} used a similar approach for Indian dialects.

\bluetext{\noindent\textbf{Harnessing Linguistic Features:} \citet{erdmann-etal-2017-low} focus on translation among Arabic dialects in a low-resource setting where they supplement small parallel corpora with morpho-syntactic information injected into the model for machine translation. In general, incorporating linguistic features into the MT framework is known to significantly boost translation quality in low-resource settings \cite{chakrabarty-etal-2022-featurebart,chakrabarty-etal-2020-improving}. Especially, pre-training by leveraging linguistic features, as done by \citet{chakrabarty-etal-2022-featurebart}, should be beneficial for \redtext{dialectal} translation\orangetext{,} which is typically a low-resource problem.}


\noindent\textbf{Code-mixed training}: \citet{lu-etal-2022-exploring} use XLM for Translation between \redtext{Hokkien}-Mandarin code-mixed text. They observe that continuous training with code-mixed data enables monolingual language models to provide better performance when applied to code-mixed tasks. 

\noindent \textbf{Data Creation for MT between dialects:} ~\citet{zbib-etal-2012-machine} and \citet{meftouh-etal-2015-machine} also focus on multi-dialect MT data collection for Arabic, which is, once again, to be noted as one of the most studied languages for dialects. \citet{xu-etal-2015-building} use a Hidden Markov-based model to create word alignment between dialects of Chinese: Mainland Chinese, Hong Kong Chinese, and Taiwan Chinese. The outcome is a monolingual corpus that contains corresponding words used in the three dialects. Their approach was shown to be effective for three different alignment mapping cases. Rather than use word alignment, \citet{hassani-2017-kurdish} works on Kurdish \redtext{dialectal} MT using dictionaries and show that having limited to no parallel corpora is not a significant barrier for inter-dialect translation. \greentext{YORULECT~\cite{ahia-etal-2024-voices} releases parallel text and speech corpus for four regional variants of Yorùbá, by collecting data from within native communities for each variant, and providing a benchmark for machine translation, speech recognition, and text-to-speech synthesis. Most recently, \citet{dabre-etal-2024-machine} worked on Kadodi, a dialect of Marathi, for which the local community was involved for parallel corpora creation, further highlighting the need for community involvement.}

All these works emphasize that a small amount of parallel data between dialects is always important; however, data synthesis and transfer learning from a high-resource dialect is always impactful, especially in conjunction with character and word level perturbation methods.

\subsubsection{MT between dialects and another language} The second category, involving the harder challenge of machine translation between a dialect and another language, has received far more attention. We cover notable works below. 

\noindent \textbf{Dialect pivoting:} 
An early work in this regard is by~\citet{paul2011dialect}. They present a pivot-based MT approach for the translation of four
dialects of Japanese, namely Kumamoto, Kyoto, Okinawa, and Osaka. In order to map sentences across dialects, they use a character-based generative graphical model. They then translate the dialects into four Indo-European languages, using standardised Japanese as the pivot language. 
~\citet{jeblee-etal-2014-domain} focus on using modern standard Arabic as a pivot when translating from English to the Egyptian Arabic dialect.

\noindent\textbf{Unsupervised segmentation}: Different from \citet{abe-etal-2011-example} who focus on characters, ~\citet{al-mannai-etal-2014-unsupervised,salloum2022unsupervised} work on Arabic \redtext{dialectal} translation\orangetext{,} which shows that unsupervised word segmentation \redtext{is just as effective if not better} for translation into English.

\noindent \textbf{Evaluating existing translators on \redtext{dialectal} datasets}:~\citet{kantharuban-etal-2023-quantifying} show the performance of MT between English and dialects of seven languages. Using state-of-the-art MT systems such as Google NMT and Meta NLLB, they evaluate MT in both directions (to and from English). They report a drastic drop in BLEU for dialects of German, Portuguese and Bengali. \bluetext{~\citet{de2023mt} train NMT systems to evaluate the performance of legal domain translation for Italian $\Leftrightarrow$ South Tyrolean German, where their models show better performance compared to Google Translate and DeepL for this niche use case.} \orangetext{Similarly, CODET~\cite{alam-etal-2024-codet} is a contrastive dialectal benchmark dataset for evaluating machine translation systems focusing on $891$ variations from $12$ languages.}

\noindent \textbf{Using inferred \redtext{dialectal} labels to guide translation}:~\citet{sun-etal-2023-dialect} add language-dialect information as predicted by a language identifier as an input when training an MT system. They further improve the metrics for  robust evaluation of text generation systems for different languages and dialects. They report their results on several dialects of English, Chinese, Portuguese and so on. They use few-shot prompting to create semantic perturbations to train T5. The results show that dialect-awareness improves the performance of translation.
\citet{shapiro-duh-2019-comparing} explore a multidialect system and identify when \redtext{dialectal} identification is useful. 
\citet{tahssin-etal-2020-identifying} focus on dialect identification itself using AraBERT models.
\citet{salloum-etal-2014-sentence} \orangetext{focus on} sentence-level dialect identification for \orangetext{an SMT model} selection where \orangetext{the model} is optimised for that dialect. 

\noindent \textbf{Learning through exemplars}: Few-shot learning involves the use of examples in a prompt to guide the generation through a language model.~\citet{riley-etal-2023-frmt} present a few-shot machine translation approach for translation between English and two variants of Portuguese and Chinese. The parallel corpus is manually created by native speakers. The \orangetext{exemplars from the specific dialect are} \redtext{used} to obtain translations of English sentences.

\noindent \textbf{User-generated content}: User-generated content (UGC) is often mistranslated on social media, especially, for low-resource languages like \redtext{dialectal} Arabic.~\citet{saadany-etal-2022-semi} train a Transformer to translate from \redtext{dialectal} Arabic to English where they focus on challenges in translation of UGC, and propose a sentiment-aware evaluation metric for translation. They discuss results on multiple test sets, including a hand-crafted test set, and analyse the performance for a semi-supervised approach compared to a baseline NMT system, a pivoting-based system, and Google Translate.  

\noindent \textbf{Use of multiple translation models}: Translation models that translate between the standard version and a dialect can assist machine translation.~\citet{kumar-etal-2021-machine} show an approach for MT from English to Ukrainian, Belarusian, Nynorsk, and Arabic dialects. They use two models: a dialect-to-standard translation model, and a  standard source-to-target language  translation model.

\noindent \textbf{Data creation for MT of dialect to another language:} 
\citet{hassan-etal-2017-synthetic} explore synthetic data creation using word pairs between dialects based on embeddings. They take seed data, transform it into its \redtext{dialectal} variant and now have a \redtext{dialectal} parallel corpus. 
Similarly,~\citet{almansor-al-ani-2017-translating} focus on using monolingual data and tiny parallel corpora in conjunction \redtext{with} cross-\redtext{dialectal} embeddings to improve MT between dialects. ~\citet{sajjad-etal-2020-arabench} take dialect MT evaluation further and focus on multi-domain coarse-grained analysis of dialects of Arabic via their AraBench benchmark. \bluetext{\citet{hamed-etal-2022-arzen} propose an Arabic-English code-switched speech translation dataset\orangetext{,} which represents a practical use case since a vast majority of dialects are often spoken. There is a significant dearth of code-mixed datasets and recommend researchers to focus on the same. \citet{alkheder-etal-2023-benchmarking}, recognising the increasing usage of Arabic in several regions of Turkey, expand the MADAR corpus \cite{obeid-etal-2018-madar} to enable benchmarking of translation between Arabic and Turkish.~\citet{contarino2021neural} curates LEXB, a parallel corpus between South Tyrolean German and Italian containing nearly $175,000$ parallel segments from the legal domain. To curate parallel data, they use the LexBrowser database\footnote{http://lexbrowser.provinz.bz.it/; \greentext{Accessed on 20th November, 2024.}} and $20$ national laws
and codes (Civil Code, Criminal Code) translated into German.}\greentext{~\citet{igarashi-miyagawa-2024-enhancing} curate parallel corpus for Ainu$\Longleftrightarrow$Japanese where text for Ainu was curated from after post-processing OCR ouput for books, and other online resources. Ainu is a critically endangered family of dialects from from Northern Japan without any native writing script.}
%






\subsubsection{Dialect MT in Shared Tasks} Given that most dialects are spoken and not written, the IWSLT workshop, which focuses on spoken language translation, has been conducting shared tasks on dialects under the banner of low-resource MT. The 2022\footnote{\url{https://iwslt.org/2022/dialect}; Accessed on 9th January, 2024.} and 2023\footnote{\url{https://iwslt.org/2023/low-resource}; Accessed on 9th January, 2024.} workshops featured \redtext{dialectal} speech translation, with resources for text-text as well as speech-text translation. The focus, as is typically the case, is on dialects of Arabic like Tunisian, Egyptian and Moroccan. The shared tasks are an excellent source of datasets and benchmarks for \redtext{dialectal} MT. \redtext{Most recently, the ArabicNLP 2023\footnote{\url{https://arabicnlp2023.sigarab.org/}; \greentext{Accessed on 20th November, 2024.}} conference offered a shared task\footnote{\url{https://nadi.dlnlp.ai/}; \greentext{Accessed on 20th November, 2024.}} on translation from 4 Arabic dialect to modern standard Arabic. \orangetext{Over the years, NADI has reported progress on country and province-level dialect identification, dialectal sentiment analysis, and dialect to MSA machine translation~\cite{abdul-mageed-etal-2020-nadi,abdul-mageed-etal-2021-nadi,abdul-mageed-etal-2022-nadi,abdul-mageed-etal-2023-nadi}}.  We should also note that the Workshop on Machine Translation (WMT\footnote{\url{https://www2.statmt.org/wmt24}; \greentext{Accessed on 20th November, 2024.}}) and the Workshop on Asian Translation (WAT\footnote{\url{https://lotus.kuee.kyoto-u.ac.jp/WAT/WAT2024/index.html}}) often feature shared tasks on closely related languages.} The 2024\footnote{https://iwslt.org/2024/low-resource; Accessed on 9th January, 2024.} edition of IWSLT is expected to focus on North Levantine Arabic.



\subsection{Dialogue Systems}
Dialogue systems, crucial in facilitating human-computer interaction, are categorised into task-oriented, chit-chat, and hybrid systems. These systems, especially when dialect-aware, face the added challenge of understanding and adapting to linguistic variations.

\noindent\textbf{Task-oriented dialogue system}:
Task-oriented systems are designed to accomplish specific tasks. They integrate NLU, a dialogue manager, and NLG components. The effectiveness of these systems in handling dialects is pivotal. For instance,~\citet{elmadany2018improving,joukhadar2019arabic} study the classification of dialogue acts in Arabic dialect utterances, demonstrating the system's capacity to adapt to \redtext{dialectal} variations. \citet{al2020nabiha} use the Artificial Intelligence Markup Language (AIML) to build a chatbot that assists students with academic enquiries in the Saudi Arabian dialect. \orangetext{The VarDial 2023 campaign~\cite{aepli-etal-2023-findings} reports progress on \textit{slot and intent detection for low-resource language varieties} such as Swiss German, Neapolitan, South Tyrolean.} \redtext{\citet{artemova2024exploring} investigate the robustness of task-oriented dialogue systems, specifically their intent classification and slot detection components, to German dialects by applying perturbations that transform standard German sentences into colloquial variants.}


\noindent\textbf{Chit-chat dialogue system}:
Chit-chat dialogue systems, also known as open-domain systems, primarily focus on daily chat and handle broader interactions. ~\citet{ali2016botta} employ AIML and rule-based systems to manage dialectal variation in Egyptian Arabic, incorporating features like short vowels and consonantal doubling. ~\citet{ahmed2020design} also use AIML for Kurdish dialogues. Additionally, ~\citet{alshareef2020seq2seq} train a Seq2Seq model on a tweet corpus to respond to open-domain Arabic questions.

A \redtext{specialised} subset of chit-chat systems are socially-aware dialogue systems, which pay close attention to the influence of social norms and factors. These systems are designed to adhere to the cultural and social norms prevalent in different societies~\cite{hovy2021importance}. In different cultures, social norms will no doubt incorporate \orangetext{sociolect}, including whatever discourse force it may carry. \citet{ziems-etal-2023-multi} propose a framework to evaluate dialect differences in cross-dialectal English. \redtext{In addition}, \citet{zhan2023socialdial,zhan2024renovi} propose the socially-aware dialogue corpus based on Chinese culture and relevant dialectal norms. \orangetext{Sociolect} in a dialogue will dramatically affect human's understanding and behaviours towards speakers. \citet{rajai2022dealing} summarise existing problems and strategies towards dealing with \orangetext{sociolect}  in dialogues.

\noindent\textbf{Hybrid system:}
Hybrid systems combine features of both task-oriented and open-domain systems. An example is the system developed by~\citet{ben2021multilingual}, which answers user queries in various Arabic dialects like Tunisian, Igbo, Yoruba, and Hausa. This chatbot addresses both official FAQs, especially related to COVID-19, and informal chit-chat, responding to questions in the local dialect.


Awareness of social and societal norms of behaviour is particularly important in dialogue systems that serve specific transactional goals, be it to book a doctor’s appointment, to ask questions about income tax or to make a customer service complaint. Research in interactional socio-linguistics~\cite{gumperz1982discourse} has, in a rich body of research in different social contexts such as employment interviews~\cite{roberts2021linguistic}, shown that people interpret communicative intent against their own background expectations of what is `normal' or `expected behaviour'. This has the potential to exacerbate inequality (for example, by restricting access to employment), in particular, for underrepresented groups such as migrants.

In the case of dialogue systems, a lack of representation of different dialects (\textit{e.g.,} due to the lack of diverse training data) has the potential to cause similar effects: \redtext{If dialogue systems are not aware of social norms inherent to different dialects, and if what is communicated does not match users’ expectations,} communicative intent can be misinterpreted, and underrepresented user groups might become disengaged from the system. \redtext{The xSID dataset~\cite{van2021masked} is a multilingual dataset for spoken language understanding, and includes \orangetext{a low-resource} Austro-Bavarian German dialect and other dialects.} Indeed, previous research on dialogue systems has confirmed the importance of alignment of system-style choices with user needs and preferences ~\cite{li2015hedonic,chaves2022chatbots,folstad2020users}. 

\section{Conclusion \& Future Directions}
\label{sec:concl}
Dialects are syntactic and lexical variations of a language, often associated with socially or geographically cohesive groups. This paper summarises NLP approaches for dialects of several languages. The need for NLP approaches focusing on dialects of a language rest on four motivations: dialects pose linguistic challenges, benchmarks may not have sufficient \redtext{dialectal} representation, dialect-awareness is important for fair NLP technology, and there has been growing recent work in this direction. The survey identified trends in terms of tasks (which shows shifting focus from dialect classification), languages (with more work in Arabic as compared to other languages), and \redtext{a} shifting trend towards mitigation (by either making models dialect-invariant or dialect-aware). Following that, we described different methods to create \redtext{dialectal} lexicons and datasets, ranging from location/keyword-based filtering \redtext{(of which location-based filtering has been found to be ineffective by ~\citet{goutte2016discriminating})} to manual (via recruitment of native speakers) and automatic (via automatic perturbation). We then viewed past work in the context of NLU and NLG. For NLU, we covered dialect identification, sentiment analysis, morphosyntactic analysis, parsing and more recent work in NLU benchmarks. We described how the availability of datasets in multiple languages has fuelled research in dialect identification\orangetext{,} which continues to date. Sentiment analysis techniques for dialects included peculiar de-biasing approaches, in addition to dialect invariance and dialect awareness. Approaches to parse \redtext{dialectal} datasets used or adapted existing parsers or \orangetext{developed} dialect-specific parsers. Finally, we described how recent work on NLU benchmarks highlight how adversarial learning and LoRA can be used to reduce the degradation in the performance of \redtext{dialectal} datasets as compared to the standard ones. In the case of NLG, we described work in summarisation, machine translation and dialogue systems. We described the limited, recent work in multi-document summarisation of \redtext{dialectal} documents. Following that, we discussed approaches for machine translation in the context of dialect normalisation and dialect pivoting depending on whether the translation is between dialects of a language or between dialects and another language. Finally, we described dialogue systems in the context of task-oriented, chit-chat and hybrid systems.

Based on our survey, we now identify future directions and social/ethical implications. We hope that the former will be helpful for NLP research for dialects, while the latter will get more researchers interested in this richly investigated yet emergent area of NLP. We believe that NLP researchers should adopt a socio-technical perspective~\cite{johnson2017reframing} on their role and consider not only their own possible biases influencing the selection of training data, the design of algorithms \textit{etc.} but also other social arrangements (\textit{e.g.,} users and their behaviours) relevant to specific systems. \redtext{In their survey of speakers of German dialects, ~\citet{blaschke2024dialect} also discuss in detail their needs as users of language technologies.} 

\subsection{Implications to NLP research}
In addition to the trends reported earlier in the paper, the following would be potential future directions in the context of NLP.

\subsubsection{Focusing on unexplored dialects of languages}: NLP for dialects face problems akin to low-resource languages, in terms of the availability of existing resources and tools. While some dialect families, such as English and Arabic, have seen consistent efforts, dialects for other languages need more focused large-scale efforts for data curation and annotation. While English is arguably the leading language for advances in NLP, efforts remain to be done to fully represent the full diversity of the English language itself through appropriate datasets and models that are curated for specific \redtext{dialectal} tasks. It is not always necessary to create new datasets, given that datasets specific to particular dialects are available. However, caution is advised for dialogue systems as many existing corpora – with the exception of those focusing on English as a \textit{lingua franca} (ELF) – are dominated by written texts, which may not represent the richness of \redtext{dialectal} variations of spoken language. 

\subsubsection{Rethinking the pre-training of LLMs}:~\citet{chow2022singlish} present a computational grammar for Singaporean English. Such \redtext{dialectal} representations can be useful to generalise the ability of LLMs. It would be beneficial for LLMs to be able to ingest other kinds of information such as dialect-specific grammatical structures. Ability to pre-train LLMs using data in different formats (not just modality, which is currently a popular paradigm) may improve their performance for diverse datasets such as dialects. \greentext{Similar impact of pre-training language data distribution is known for cross-lingual meaning transfer within NLP tasks~\cite{qian-etal-2024-large,zerva-etal-2024-findings} involving low-resource languages~\cite{nigatu-etal-2024-zenos}.}

\subsubsection{Dialect identification as an \redtext{auxiliary} task}: Multi-task learning is used to train models for multiple tasks. Dialect identification could be used as one of the tasks in order to train equitable models.~\citet{lent2023creoleval} present a multi-task, multi-lingual dataset of Creole languages. They report the baseline performance of NLU and MT tasks on the dataset using appropriate models. Availability of such large benchmarks will aid the development of new methods and models.
\subsubsection{Rethinking LLM Evaluation}:~\citet{xiao2023task} \orangetext{evaluate LoRA adapters for unseen English dialects}, and they say: ``a comprehensive examination of PEFT modules for dialects is
needed, which we leave for future work.". Similar evaluations can be performed for other NLP approaches. In addition, new evaluation techniques and metrics will be useful to measure \redtext{dialectal} variation and its potential correlation with the performance of NLP tasks. Two recent papers can be of value.~\citet{lameli2023measure} present a measure for spatial language variation. Using distance between locations as a heuristic for \redtext{dialectal} similarity, they examine variations in dialects of German. Also,~\citet{keleg-etal-2023-aldi} use a dataset in Arabic labeled with the degree of dialectness, to train a BERT-based regression model.


\subsection{Ethical \& Social Implications}
Overall, \redtext{dialectal} NLP presents an excellent avenue for research with huge social implications. We highlight three considerations of relevance.

\subsubsection{Social Implications}
While everyone speaks a standard dialect, most people tend to feel familiarity with people who speak specific dialects. Furthermore, certain traditions and practices are tied to localities\orangetext{,} which are\orangetext{,} in turn\orangetext{,} tied to dialects. If the goal of NLP research is to make communication seamless then \orangetext{the correct way to do so is via a strong emphasis on dialects, by effectively engaging with speakers of the dialects~\cite{blaschke2024dialect}}. \redtext{Most dialects around the world are under-represented in modern-day NLP, which can potentially disadvantage them or leave them out of the benefits of LLMs.}. There is also a growing concern among speakers of specific dialects that their language is dying either due to the pervasiveness of English via the internet, \redtext{another majority language, or a related dialect\orangetext{,} which has higher official support or recognition}. We should acknowledge these concerns and make headway into preserving as many dialects as possible, at the risk of losing valuable aspects of the vast tapestry of culture and history.

\subsubsection{\redtext{Dialectal} Research By Dialect Speakers}
Linguistic research colonisation is the process where researchers who do not speak specific languages nor have connections with them conduct research on said languages. Despite the negative connotation of colonization, this is not a bad thing, because no one should monopolise working on specific languages. However, it highlights that there are haves and have-nots, where the haves are researchers and organisations with funding who can work on dialects and the have-nots are the researchers who would like to work on dialects but simply lack funding. Recently, there has been a growing trend where language speakers are reclaiming dominion over research involving their own languages. For example, there has been an explosive growth in the number of researchers and groups like DeepLearning Indaba, Masakhane from African countries working solely on African languages and organisations like AI4Bharat in India working on Indian languages. Indeed, they have shown that a dedicated focus on language research by speakers of these languages \redtext{leads} to better NLP systems. We, therefore, propose that the organisations with funding leverage their privilege and support those without funding so as to ensure that work on dialects is led and owned by groups that are most connected \orangetext{to} and impacted by dialects. This will lead to true diversity, equality and inclusivity in NLP research\orangetext{,} which will strongly impact society. \redtext{Towards this, the emerging sentiment in recent thematic papers in NLP is that communities that speak the dialects must be involved in the development of language technologies for the \orangetext{communities~\cite{ramponi2024language, bird-2022-local}}.}

\subsubsection{Normalising Working on and Speaking Dialects}
One aspect that limits \redtext{dialectal} research is the concept of shame in speaking a certain language or a dialect, an aspect also known as \href{https://unravellingmag.com/articles/linguistic-self-hate/}{linguistic self-hatred}. For example, take the case of Mauritian Creole, whose speakers are dwindling by the day, mainly because the younger generation \href{https://lexpress.mu/blog/286734/creole-disillusion}{feels shame} in speaking their native language. The \href{https://hanvkonn.wordpress.com/2019/06/22/ashamed-of-speaking-in-konkani/}{same exists for Konkani}. While there are no official reports highlighting the same for dialects, it is not far-fetched to consider that linguistic self-hatred will exist here as well. It is time to end this self-hatred and normalise speaking dialects. By doing so, people speaking dialects will become more enthusiastic about preserving their dialects and this will inevitably aid research on dialects, thereby positively impacting society. Dialects are closely tied to culture and such differences have not been captured explicitly beyond the works described in this paper.

\bibliography{sample-base, raj-citations} 


\begin{thebibliography}{227}


\ifx \showCODEN    \undefined \def \showCODEN     #1{\unskip}     \fi
\ifx \showDOI      \undefined \def \showDOI       #1{#1}\fi
\ifx \showISBNx    \undefined \def \showISBNx     #1{\unskip}     \fi
\ifx \showISBNxiii \undefined \def \showISBNxiii  #1{\unskip}     \fi
\ifx \showISSN     \undefined \def \showISSN      #1{\unskip}     \fi
\ifx \showLCCN     \undefined \def \showLCCN      #1{\unskip}     \fi
\ifx \shownote     \undefined \def \shownote      #1{#1}          \fi
\ifx \showarticletitle \undefined \def \showarticletitle #1{#1}   \fi
\ifx \showURL      \undefined \def \showURL       {\relax}        \fi
\providecommand\bibfield[2]{#2}
\providecommand\bibinfo[2]{#2}
\providecommand\natexlab[1]{#1}
\providecommand\showeprint[2][]{arXiv:#2}

\bibitem[Abdul-Mageed et~al\mbox{.}(2018)]%
        {abdul2018you}
\bibfield{author}{\bibinfo{person}{Muhammad Abdul-Mageed},
  \bibinfo{person}{Hassan Alhuzali}, {and} \bibinfo{person}{Mohamed Elaraby}.}
  \bibinfo{year}{2018}\natexlab{}.
\newblock \showarticletitle{You tweet what you speak: A city-level dataset of
  arabic dialects}. In \bibinfo{booktitle}{\emph{LREC}}.
\newblock


\bibitem[Abdul-Mageed and Diab(2014)]%
        {abdul2014sana}
\bibfield{author}{\bibinfo{person}{Muhammad Abdul-Mageed} {and}
  \bibinfo{person}{Mona~T Diab}.} \bibinfo{year}{2014}\natexlab{}.
\newblock \showarticletitle{Sana: A large scale multi-genre, multi-dialect
  lexicon for arabic subjectivity and sentiment analysis.}. In
  \bibinfo{booktitle}{\emph{LREC}}. \bibinfo{pages}{1162--1169}.
\newblock


\bibitem[Abdul-Mageed et~al\mbox{.}(2023)]%
        {abdul-mageed-etal-2023-nadi}
\bibfield{author}{\bibinfo{person}{Muhammad Abdul-Mageed},
  \bibinfo{person}{AbdelRahim Elmadany}, \bibinfo{person}{Chiyu Zhang},
  \bibinfo{person}{El~Moatez~Billah Nagoudi}, \bibinfo{person}{Houda Bouamor},
  {and} \bibinfo{person}{Nizar Habash}.} \bibinfo{year}{2023}\natexlab{}.
\newblock \showarticletitle{{NADI} 2023: The Fourth Nuanced {A}rabic Dialect
  Identification Shared Task}. In \bibinfo{booktitle}{\emph{ArabicNLP}}.
  \bibinfo{publisher}{Association for Computational Linguistics},
  \bibinfo{address}{Singapore (Hybrid)}, \bibinfo{pages}{600--613}.
\newblock
\urldef\tempurl%
\url{https://doi.org/10.18653/v1/2023.arabicnlp-1.62}
\showDOI{\tempurl}


\bibitem[Abdul-Mageed et~al\mbox{.}(2020)]%
        {abdul-mageed-etal-2020-nadi}
\bibfield{author}{\bibinfo{person}{Muhammad Abdul-Mageed},
  \bibinfo{person}{Chiyu Zhang}, \bibinfo{person}{Houda Bouamor}, {and}
  \bibinfo{person}{Nizar Habash}.} \bibinfo{year}{2020}\natexlab{}.
\newblock \showarticletitle{{NADI} 2020: The First Nuanced {A}rabic Dialect
  Identification Shared Task}. In \bibinfo{booktitle}{\emph{Fifth Arabic
  Natural Language Processing Workshop}}. \bibinfo{publisher}{ACL},
  \bibinfo{address}{Barcelona, Spain (Online)}, \bibinfo{pages}{97--110}.
\newblock
\urldef\tempurl%
\url{https://aclanthology.org/2020.wanlp-1.9}
\showURL{%
\tempurl}


\bibitem[Abdul-Mageed et~al\mbox{.}(2021)]%
        {abdul-mageed-etal-2021-nadi}
\bibfield{author}{\bibinfo{person}{Muhammad Abdul-Mageed},
  \bibinfo{person}{Chiyu Zhang}, \bibinfo{person}{AbdelRahim Elmadany},
  \bibinfo{person}{Houda Bouamor}, {and} \bibinfo{person}{Nizar Habash}.}
  \bibinfo{year}{2021}\natexlab{}.
\newblock \showarticletitle{{NADI} 2021: The Second Nuanced {A}rabic Dialect
  Identification Shared Task}. In \bibinfo{booktitle}{\emph{Sixth Arabic
  Natural Language Processing Workshop}}. \bibinfo{publisher}{ACL},
  \bibinfo{address}{Kyiv, Ukraine (Virtual)}, \bibinfo{pages}{244--259}.
\newblock
\urldef\tempurl%
\url{https://aclanthology.org/2021.wanlp-1.28}
\showURL{%
\tempurl}


\bibitem[Abdul-Mageed et~al\mbox{.}(2022)]%
        {abdul-mageed-etal-2022-nadi}
\bibfield{author}{\bibinfo{person}{Muhammad Abdul-Mageed},
  \bibinfo{person}{Chiyu Zhang}, \bibinfo{person}{AbdelRahim Elmadany},
  \bibinfo{person}{Houda Bouamor}, {and} \bibinfo{person}{Nizar Habash}.}
  \bibinfo{year}{2022}\natexlab{}.
\newblock \showarticletitle{{NADI} 2022: The Third Nuanced {A}rabic Dialect
  Identification Shared Task}. In \bibinfo{booktitle}{\emph{Seventh Arabic
  Natural Language Processing Workshop (WANLP)}}.
  \bibinfo{publisher}{Association for Computational Linguistics},
  \bibinfo{address}{Abu Dhabi, United Arab Emirates (Hybrid)},
  \bibinfo{pages}{85--97}.
\newblock
\urldef\tempurl%
\url{https://doi.org/10.18653/v1/2022.wanlp-1.9}
\showDOI{\tempurl}


\bibitem[Abdulrahim et~al\mbox{.}(2022)]%
        {abdulrahim2022bahrain}
\bibfield{author}{\bibinfo{person}{Dana Abdulrahim}, \bibinfo{person}{Go
  Inoue}, \bibinfo{person}{Latifa Shamsan}, \bibinfo{person}{Salam Khalifa},
  {and} \bibinfo{person}{Nizar Habash}.} \bibinfo{year}{2022}\natexlab{}.
\newblock \showarticletitle{The Bahrain Corpus: A Multi-genre Corpus of
  Bahraini Arabic}. In \bibinfo{booktitle}{\emph{LREC}}.
  \bibinfo{pages}{2345--2352}.
\newblock


\bibitem[Abe et~al\mbox{.}(2018)]%
        {abe-etal-2018-multi}
\bibfield{author}{\bibinfo{person}{Kaori Abe}, \bibinfo{person}{Yuichiroh
  Matsubayashi}, \bibinfo{person}{Naoaki Okazaki}, {and}
  \bibinfo{person}{Kentaro Inui}.} \bibinfo{year}{2018}\natexlab{}.
\newblock \showarticletitle{Multi-dialect Neural Machine Translation and
  Dialectometry}. In \bibinfo{booktitle}{\emph{PACLIC}}.
  \bibinfo{publisher}{ACL}, \bibinfo{address}{Hong Kong}.
\newblock
\urldef\tempurl%
\url{https://aclanthology.org/Y18-1001}
\showURL{%
\tempurl}


\bibitem[Abe et~al\mbox{.}(2011)]%
        {abe-etal-2011-example}
\bibfield{author}{\bibinfo{person}{Yusuke Abe}, \bibinfo{person}{Takafumi
  Suzuki}, \bibinfo{person}{Bing Liang}, \bibinfo{person}{Takehito Utsuro},
  \bibinfo{person}{Mikio Yamamoto}, \bibinfo{person}{Suguru Matsuyoshi}, {and}
  \bibinfo{person}{Yasuhide Kawada}.} \bibinfo{year}{2011}\natexlab{}.
\newblock \showarticletitle{Example-based Translation of {J}apanese Functional
  Expressions utilizing Semantic Equivalence Classes}. In
  \bibinfo{booktitle}{\emph{4th Workshop on Patent Translation}}.
  \bibinfo{address}{Xiamen, China}.
\newblock
\urldef\tempurl%
\url{https://aclanthology.org/2011.mtsummit-wpt.10}
\showURL{%
\tempurl}


\bibitem[Aepli et~al\mbox{.}(2022)]%
        {aepli-etal-2022-findings}
\bibfield{author}{\bibinfo{person}{No{\"e}mi Aepli}, \bibinfo{person}{Antonios
  Anastasopoulos}, \bibinfo{person}{Adrian-Gabriel Chifu},
  \bibinfo{person}{William Domingues}, \bibinfo{person}{Fahim Faisal},
  \bibinfo{person}{Mihaela Gaman}, \bibinfo{person}{Radu~Tudor Ionescu}, {and}
  \bibinfo{person}{Yves Scherrer}.} \bibinfo{year}{2022}\natexlab{}.
\newblock \showarticletitle{Findings of the {V}ar{D}ial Evaluation Campaign
  2022}. In \bibinfo{booktitle}{\emph{Ninth Workshop on NLP for Similar
  Languages, Varieties and Dialects}}. \bibinfo{publisher}{Association for
  Computational Linguistics}, \bibinfo{address}{Gyeongju, Republic of Korea},
  \bibinfo{pages}{1--13}.
\newblock


\bibitem[Aepli et~al\mbox{.}(2023)]%
        {aepli-etal-2023-findings}
\bibfield{author}{\bibinfo{person}{No{\"e}mi Aepli},
  \bibinfo{person}{{\c{C}}a{\u{g}}r{\i} {\c{C}}{\"o}ltekin},
  \bibinfo{person}{Rob Van Der~Goot}, \bibinfo{person}{Tommi Jauhiainen},
  \bibinfo{person}{Mourhaf Kazzaz}, \bibinfo{person}{Nikola
  Ljube{\v{s}}i{\'c}}, \bibinfo{person}{Kai North}, \bibinfo{person}{Barbara
  Plank}, \bibinfo{person}{Yves Scherrer}, {and} \bibinfo{person}{Marcos
  Zampieri}.} \bibinfo{year}{2023}\natexlab{}.
\newblock \showarticletitle{Findings of the {V}ar{D}ial Evaluation Campaign
  2023}. In \bibinfo{booktitle}{\emph{VarDial}}.
  \bibinfo{publisher}{Association for Computational Linguistics},
  \bibinfo{address}{Dubrovnik, Croatia}, \bibinfo{pages}{251--261}.
\newblock


\bibitem[Aepli and Sennrich(2022)]%
        {aepli-sennrich-2022-improving}
\bibfield{author}{\bibinfo{person}{No{\"e}mi Aepli} {and} \bibinfo{person}{Rico
  Sennrich}.} \bibinfo{year}{2022}\natexlab{}.
\newblock \showarticletitle{Improving Zero-Shot Cross-lingual Transfer Between
  Closely Related Languages by Injecting Character-Level Noise}. In
  \bibinfo{booktitle}{\emph{Findings of ACL}}. \bibinfo{pages}{4074--4083}.
\newblock
\urldef\tempurl%
\url{https://doi.org/10.18653/v1/2022.findings-acl.321}
\showDOI{\tempurl}


\bibitem[Ahia et~al\mbox{.}(2024)]%
        {ahia-etal-2024-voices}
\bibfield{author}{\bibinfo{person}{Orevaoghene Ahia},
  \bibinfo{person}{Anuoluwapo Aremu}, \bibinfo{person}{Diana Abagyan},
  \bibinfo{person}{Hila Gonen}, \bibinfo{person}{David~Ifeoluwa Adelani},
  \bibinfo{person}{Daud Abolade}, \bibinfo{person}{Noah~A. Smith}, {and}
  \bibinfo{person}{Yulia Tsvetkov}.} \bibinfo{year}{2024}\natexlab{}.
\newblock \showarticletitle{Voices Unheard: {NLP} Resources and Models for
  {Y}or{\`u}b{\'a} Regional Dialects}. In \bibinfo{booktitle}{\emph{EMNLP}}.
  \bibinfo{publisher}{Association for Computational Linguistics},
  \bibinfo{address}{Miami, Florida, USA}, \bibinfo{pages}{4392--4409}.
\newblock
\urldef\tempurl%
\url{https://aclanthology.org/2024.emnlp-main.251}
\showURL{%
\tempurl}


\bibitem[Ahia et~al\mbox{.}(2023)]%
        {ahia2023all}
\bibfield{author}{\bibinfo{person}{Orevaoghene Ahia}, \bibinfo{person}{Sachin
  Kumar}, \bibinfo{person}{Hila Gonen}, \bibinfo{person}{Jungo Kasai},
  \bibinfo{person}{David~R Mortensen}, \bibinfo{person}{Noah~A Smith}, {and}
  \bibinfo{person}{Yulia Tsvetkov}.} \bibinfo{year}{2023}\natexlab{}.
\newblock \showarticletitle{Do All Languages Cost the Same? Tokenization in the
  Era of Commercial Language Models}. In \bibinfo{booktitle}{\emph{EMNLP}}.
  \bibinfo{pages}{9904--9923}.
\newblock


\bibitem[Ahmed and Hussein(2020)]%
        {ahmed2020design}
\bibfield{author}{\bibinfo{person}{Hemn~Karim Ahmed} {and}
  \bibinfo{person}{Jamal~Ali Hussein}.} \bibinfo{year}{2020}\natexlab{}.
\newblock \showarticletitle{Design and Implementation of a Chatbot for Kurdish
  Language Speakers Using Chatfuel Platform}.
\newblock \bibinfo{journal}{\emph{Kurdistan Journal of Applied Research}}
  (\bibinfo{year}{2020}), \bibinfo{pages}{117--135}.
\newblock


\bibitem[Aji et~al\mbox{.}(2022)]%
        {aji-etal-2022-one}
\bibfield{author}{\bibinfo{person}{Alham~Fikri Aji},
  \bibinfo{person}{Genta~Indra Winata}, \bibinfo{person}{Fajri Koto},
  \bibinfo{person}{Samuel Cahyawijaya}, \bibinfo{person}{Ade Romadhony},
  \bibinfo{person}{Rahmad Mahendra}, \bibinfo{person}{Kemal Kurniawan},
  \bibinfo{person}{David Moeljadi}, \bibinfo{person}{Radityo~Eko Prasojo},
  \bibinfo{person}{Timothy Baldwin}, \bibinfo{person}{Jey~Han Lau}, {and}
  \bibinfo{person}{Sebastian Ruder}.} \bibinfo{year}{2022}\natexlab{}.
\newblock \showarticletitle{One Country, 700+ Languages: {NLP} Challenges for
  Underrepresented Languages and Dialects in {I}ndonesia}. In
  \bibinfo{booktitle}{\emph{ACL}}. \bibinfo{address}{Dublin, Ireland},
  \bibinfo{pages}{7226--7249}.
\newblock
\urldef\tempurl%
\url{https://doi.org/10.18653/v1/2022.acl-long.500}
\showDOI{\tempurl}


\bibitem[Al-Ghadhban and Al-Twairesh(2020)]%
        {al2020nabiha}
\bibfield{author}{\bibinfo{person}{Dana Al-Ghadhban} {and}
  \bibinfo{person}{Nora Al-Twairesh}.} \bibinfo{year}{2020}\natexlab{}.
\newblock \showarticletitle{Nabiha: an Arabic dialect chatbot}.
\newblock \bibinfo{journal}{\emph{International Journal of Advanced Computer
  Science and Applications}} \bibinfo{volume}{11}, \bibinfo{number}{3}
  (\bibinfo{year}{2020}).
\newblock


\bibitem[Al-Mannai et~al\mbox{.}(2014)]%
        {al-mannai-etal-2014-unsupervised}
\bibfield{author}{\bibinfo{person}{Kamla Al-Mannai}, \bibinfo{person}{Hassan
  Sajjad}, \bibinfo{person}{Alaa Khader}, \bibinfo{person}{Fahad Al~Obaidli},
  \bibinfo{person}{Preslav Nakov}, {and} \bibinfo{person}{Stephan Vogel}.}
  \bibinfo{year}{2014}\natexlab{}.
\newblock \showarticletitle{Unsupervised Word Segmentation Improves Dialectal
  {A}rabic to {E}nglish Machine Translation}. In
  \bibinfo{booktitle}{\emph{Workshop on Arabic Natural Language Processing}}.
  \bibinfo{publisher}{ACL}, \bibinfo{address}{Doha, Qatar},
  \bibinfo{pages}{207--216}.
\newblock
\urldef\tempurl%
\url{https://doi.org/10.3115/v1/W14-3628}
\showDOI{\tempurl}


\bibitem[Alam et~al\mbox{.}(2024)]%
        {alam-etal-2024-codet}
\bibfield{author}{\bibinfo{person}{Md~Mahfuz~Ibn Alam}, \bibinfo{person}{Sina
  Ahmadi}, {and} \bibinfo{person}{Antonios Anastasopoulos}.}
  \bibinfo{year}{2024}\natexlab{}.
\newblock \showarticletitle{{CODET}: A Benchmark for Contrastive Dialectal
  Evaluation of Machine Translation}. In \bibinfo{booktitle}{\emph{Findings of
  EACL}}. \bibinfo{pages}{1790--1859}.
\newblock
\urldef\tempurl%
\url{https://aclanthology.org/2024.findings-eacl.125}
\showURL{%
\tempurl}


\bibitem[Ali and Habash(2016)]%
        {ali2016botta}
\bibfield{author}{\bibinfo{person}{Dana~Abu Ali} {and} \bibinfo{person}{Nizar
  Habash}.} \bibinfo{year}{2016}\natexlab{}.
\newblock \showarticletitle{Botta: An arabic dialect chatbot}. In
  \bibinfo{booktitle}{\emph{COLING (System Demonstrations)}}.
  \bibinfo{pages}{208--212}.
\newblock


\bibitem[Alkheder et~al\mbox{.}(2023)]%
        {alkheder-etal-2023-benchmarking}
\bibfield{author}{\bibinfo{person}{Hasan Alkheder}, \bibinfo{person}{Houda
  Bouamor}, \bibinfo{person}{Nizar Habash}, {and} \bibinfo{person}{Ahmet
  Zengin}.} \bibinfo{year}{2023}\natexlab{}.
\newblock \showarticletitle{Benchmarking Dialectal {A}rabic-{T}urkish Machine
  Translation}. In \bibinfo{booktitle}{\emph{Machine Translation Summit XIX}}.
  \bibinfo{publisher}{Asia-Pacific Association for Machine Translation},
  \bibinfo{address}{Macau SAR, China}, \bibinfo{pages}{261--271}.
\newblock
\urldef\tempurl%
\url{https://aclanthology.org/2023.mtsummit-research.22}
\showURL{%
\tempurl}


\bibitem[Almansor and Al-Ani(2017)]%
        {almansor-al-ani-2017-translating}
\bibfield{author}{\bibinfo{person}{Ebtesam~H Almansor} {and}
  \bibinfo{person}{Ahmed Al-Ani}.} \bibinfo{year}{2017}\natexlab{}.
\newblock \showarticletitle{Translating Dialectal {A}rabic as Low Resource
  Language using Word Embedding}. In \bibinfo{booktitle}{\emph{RANLP}}.
  \bibinfo{publisher}{INCOMA Ltd.}, \bibinfo{address}{Varna, Bulgaria},
  \bibinfo{pages}{52--57}.
\newblock
\urldef\tempurl%
\url{https://doi.org/10.26615/978-954-452-049-6_008}
\showDOI{\tempurl}


\bibitem[Alshareef and Siddiqui(2020)]%
        {alshareef2020seq2seq}
\bibfield{author}{\bibinfo{person}{Tahani Alshareef} {and}
  \bibinfo{person}{Muazzam~Ahmed Siddiqui}.} \bibinfo{year}{2020}\natexlab{}.
\newblock \showarticletitle{A seq2seq neural network based conversational agent
  for gulf arabic dialect}. In \bibinfo{booktitle}{\emph{2020 21st
  International Arab Conference on Information Technology (ACIT)}}. IEEE,
  \bibinfo{pages}{1--7}.
\newblock


\bibitem[Alshutayri and Atwell(2018)]%
        {alshutayri2018creating}
\bibfield{author}{\bibinfo{person}{Areej Alshutayri} {and}
  \bibinfo{person}{Eric Atwell}.} \bibinfo{year}{2018}\natexlab{}.
\newblock \showarticletitle{Creating an Arabic Dialect Text Corpus by Exploring
  Twitter, Facebook, and Online Newspapers}. In \bibinfo{booktitle}{\emph{OSACT
  3: The 3rd Workshop on Open-Source Arabic Corpora and Processing Tools}}.
  \bibinfo{pages}{54}.
\newblock


\bibitem[Artemova et~al\mbox{.}(2024)]%
        {artemova2024exploring}
\bibfield{author}{\bibinfo{person}{Ekaterina Artemova}, \bibinfo{person}{Verena
  Blaschke}, {and} \bibinfo{person}{Barbara Plank}.}
  \bibinfo{year}{2024}\natexlab{}.
\newblock \showarticletitle{Exploring the Robustness of Task-oriented Dialogue
  Systems for Colloquial German Varieties}. In
  \bibinfo{booktitle}{\emph{EACL)}}. \bibinfo{pages}{445--468}.
\newblock


\bibitem[Artemova and Plank(2023)]%
        {artemova-plank-2023-low}
\bibfield{author}{\bibinfo{person}{Katya Artemova} {and}
  \bibinfo{person}{Barbara Plank}.} \bibinfo{year}{2023}\natexlab{}.
\newblock \showarticletitle{Low-resource Bilingual Dialect Lexicon Induction
  with Large Language Models}. In \bibinfo{booktitle}{\emph{24th Nordic
  Conference on Computational Linguistics (NoDaLiDa)}}.
  \bibinfo{publisher}{University of Tartu Library},
  \bibinfo{address}{T{\'o}rshavn, Faroe Islands}, \bibinfo{pages}{371--385}.
\newblock
\urldef\tempurl%
\url{https://aclanthology.org/2023.nodalida-1.39}
\showURL{%
\tempurl}


\bibitem[Assiri et~al\mbox{.}(2018)]%
        {assiri2018towards}
\bibfield{author}{\bibinfo{person}{Adel Assiri}, \bibinfo{person}{Ahmed Emam},
  {and} \bibinfo{person}{Hmood Al-Dossari}.} \bibinfo{year}{2018}\natexlab{}.
\newblock \showarticletitle{Towards enhancement of a lexicon-based approach for
  Saudi dialect sentiment analysis}.
\newblock \bibinfo{journal}{\emph{Journal of information science}}
  \bibinfo{volume}{44}, \bibinfo{number}{2} (\bibinfo{year}{2018}),
  \bibinfo{pages}{184--202}.
\newblock


\bibitem[Azouaou and Guellil(2017)]%
        {azouaou2017alg}
\bibfield{author}{\bibinfo{person}{Faical Azouaou} {and} \bibinfo{person}{Imane
  Guellil}.} \bibinfo{year}{2017}\natexlab{}.
\newblock \showarticletitle{Alg/fr: A step by step construction of a lexicon
  between algerian dialect and french}. In \bibinfo{booktitle}{\emph{PACLIC}},
  Vol.~\bibinfo{volume}{31}.
\newblock


\bibitem[Bafna et~al\mbox{.}(2023)]%
        {bafna-etal-2023-cross}
\bibfield{author}{\bibinfo{person}{Niyati Bafna}, \bibinfo{person}{Cristina
  Espa{\~n}a-Bonet}, \bibinfo{person}{Josef Van~Genabith},
  \bibinfo{person}{Beno{\^\i}t Sagot}, {and} \bibinfo{person}{Rachel Bawden}.}
  \bibinfo{year}{2023}\natexlab{}.
\newblock \showarticletitle{Cross-lingual Strategies for Low-resource Language
  Modeling: A Study on Five {I}ndic Dialects}. In
  \bibinfo{booktitle}{\emph{Actes de CORIA-TALN 2023. Actes de la 30e
  Conf{\'e}rence sur le Traitement Automatique des Langues Naturelles (TALN),
  volume 1 : travaux de recherche originaux -- articles longs}}.
  \bibinfo{address}{Paris, France}, \bibinfo{pages}{28--42}.
\newblock


\bibitem[Baimukan et~al\mbox{.}(2022)]%
        {baimukan2022hierarchical}
\bibfield{author}{\bibinfo{person}{Nurpeiis Baimukan}, \bibinfo{person}{Houda
  Bouamor}, {and} \bibinfo{person}{Nizar Habash}.}
  \bibinfo{year}{2022}\natexlab{}.
\newblock \showarticletitle{Hierarchical aggregation of dialectal data for
  Arabic dialect identification}. In \bibinfo{booktitle}{\emph{LREC}}.
  \bibinfo{pages}{4586--4596}.
\newblock


\bibitem[Ball-Burack et~al\mbox{.}(2021)]%
        {ball2021differential}
\bibfield{author}{\bibinfo{person}{Ari Ball-Burack}, \bibinfo{person}{Michelle
  Seng~Ah Lee}, \bibinfo{person}{Jennifer Cobbe}, {and}
  \bibinfo{person}{Jatinder Singh}.} \bibinfo{year}{2021}\natexlab{}.
\newblock \showarticletitle{Differential tweetment: Mitigating racial dialect
  bias in harmful tweet detection}. In \bibinfo{booktitle}{\emph{2021 ACM
  Conference on Fairness, Accountability, and Transparency}}.
  \bibinfo{pages}{116--128}.
\newblock


\bibitem[Baly et~al\mbox{.}(2019)]%
        {baly2019arsentd}
\bibfield{author}{\bibinfo{person}{Ramy Baly}, \bibinfo{person}{Alaa Khaddaj},
  \bibinfo{person}{Hazem Hajj}, \bibinfo{person}{Wassim El-Hajj}, {and}
  \bibinfo{person}{Khaled~Bashir Shaban}.} \bibinfo{year}{2019}\natexlab{}.
\newblock \showarticletitle{Arsentd-lev: A multi-topic corpus for target-based
  sentiment analysis in arabic levantine tweets}.
\newblock \bibinfo{journal}{\emph{arXiv preprint arXiv:1906.01830}}
  (\bibinfo{year}{2019}).
\newblock


\bibitem[Barnes et~al\mbox{.}(2021)]%
        {barnes-etal-2021-nordial}
\bibfield{author}{\bibinfo{person}{Jeremy Barnes}, \bibinfo{person}{Petter
  M{\ae}hlum}, {and} \bibinfo{person}{Samia Touileb}.}
  \bibinfo{year}{2021}\natexlab{}.
\newblock \showarticletitle{{N}or{D}ial: A Preliminary Corpus of Written
  {N}orwegian Dialect Use}. In \bibinfo{booktitle}{\emph{23rd Nordic Conference
  on Computational Linguistics (NoDaLiDa)}}. \bibinfo{publisher}{Link{\"o}ping
  University Electronic Press, Sweden}, \bibinfo{address}{Reykjavik, Iceland
  (Online)}, \bibinfo{pages}{445--451}.
\newblock
\urldef\tempurl%
\url{https://aclanthology.org/2021.nodalida-main.51}
\showURL{%
\tempurl}


\bibitem[Ben Elhaj~Mabrouk et~al\mbox{.}(2021)]%
        {ben2021multilingual}
\bibfield{author}{\bibinfo{person}{Aymen Ben Elhaj~Mabrouk},
  \bibinfo{person}{Moez Ben Haj~Hmida}, \bibinfo{person}{Chayma Fourati},
  \bibinfo{person}{Hatem Haddad}, {and} \bibinfo{person}{Abir Messaoudi}.}
  \bibinfo{year}{2021}\natexlab{}.
\newblock \showarticletitle{A Multilingual African Embedding for FAQ Chatbots}.
\newblock \bibinfo{journal}{\emph{arXiv e-prints}} (\bibinfo{year}{2021}),
  \bibinfo{pages}{arXiv--2103}.
\newblock


\bibitem[Bird(2022)]%
        {bird-2022-local}
\bibfield{author}{\bibinfo{person}{Steven Bird}.}
  \bibinfo{year}{2022}\natexlab{}.
\newblock \showarticletitle{Local Languages, Third Spaces, and other
  High-Resource Scenarios}. In \bibinfo{booktitle}{\emph{ACL}}.
  \bibinfo{pages}{7817--7829}.
\newblock
\urldef\tempurl%
\url{https://doi.org/10.18653/v1/2022.acl-long.539}
\showDOI{\tempurl}


\bibitem[Blaschke et~al\mbox{.}(2024)]%
        {blaschke2024dialect}
\bibfield{author}{\bibinfo{person}{Verena Blaschke}, \bibinfo{person}{Christoph
  Purschke}, \bibinfo{person}{Hinrich Sch{\"u}tze}, {and}
  \bibinfo{person}{Barbara Plank}.} \bibinfo{year}{2024}\natexlab{}.
\newblock \showarticletitle{What do dialect speakers want? a survey of
  attitudes towards language technology for german dialects}.
\newblock \bibinfo{journal}{\emph{arXiv preprint arXiv:2402.11968}}
  (\bibinfo{year}{2024}).
\newblock


\bibitem[Blaschke et~al\mbox{.}(2023)]%
        {blaschke-etal-2023-manipulating}
\bibfield{author}{\bibinfo{person}{Verena Blaschke}, \bibinfo{person}{Hinrich
  Sch{\"u}tze}, {and} \bibinfo{person}{Barbara Plank}.}
  \bibinfo{year}{2023}\natexlab{}.
\newblock \showarticletitle{Does Manipulating Tokenization Aid Cross-Lingual
  Transfer? A Study on {POS} Tagging for Non-Standardized Languages}. In
  \bibinfo{booktitle}{\emph{VarDial}}. \bibinfo{pages}{40--54}.
\newblock
\urldef\tempurl%
\url{https://doi.org/10.18653/v1/2023.vardial-1.5}
\showDOI{\tempurl}


\bibitem[Blodgett et~al\mbox{.}(2020)]%
        {blodgett2020language}
\bibfield{author}{\bibinfo{person}{Su~Lin Blodgett}, \bibinfo{person}{Solon
  Barocas}, \bibinfo{person}{Hal Daum{\'e}~III}, {and} \bibinfo{person}{Hanna
  Wallach}.} \bibinfo{year}{2020}\natexlab{}.
\newblock \showarticletitle{Language (technology) is power: A critical survey
  of "bias" in nlp}.
\newblock \bibinfo{journal}{\emph{arXiv preprint arXiv:2005.14050}}
  (\bibinfo{year}{2020}).
\newblock


\bibitem[Blodgett et~al\mbox{.}(2016)]%
        {blodgett-etal-2016-demographic}
\bibfield{author}{\bibinfo{person}{Su~Lin Blodgett}, \bibinfo{person}{Lisa
  Green}, {and} \bibinfo{person}{Brendan O{'}Connor}.}
  \bibinfo{year}{2016}\natexlab{}.
\newblock \showarticletitle{Demographic Dialectal Variation in Social Media: A
  Case Study of {A}frican-{A}merican {E}nglish}. In
  \bibinfo{booktitle}{\emph{EMNLP}}. \bibinfo{address}{Austin, Texas},
  \bibinfo{pages}{1119--1130}.
\newblock
\urldef\tempurl%
\url{https://doi.org/10.18653/v1/D16-1120}
\showDOI{\tempurl}


\bibitem[Blodgett et~al\mbox{.}(2018)]%
        {blodgett-etal-2018-twitter}
\bibfield{author}{\bibinfo{person}{Su~Lin Blodgett}, \bibinfo{person}{Johnny
  Wei}, {and} \bibinfo{person}{Brendan O{'}Connor}.}
  \bibinfo{year}{2018}\natexlab{}.
\newblock \showarticletitle{{T}witter {U}niversal {D}ependency Parsing for
  {A}frican-{A}merican and Mainstream {A}merican {E}nglish}. In
  \bibinfo{booktitle}{\emph{ACL}}. \bibinfo{pages}{1415--1425}.
\newblock
\urldef\tempurl%
\url{https://doi.org/10.18653/v1/P18-1131}
\showDOI{\tempurl}


\bibitem[Bouamor et~al\mbox{.}(2014)]%
        {bouamor2014multidialectal}
\bibfield{author}{\bibinfo{person}{Houda Bouamor}, \bibinfo{person}{Nizar
  Habash}, {and} \bibinfo{person}{Kemal Oflazer}.}
  \bibinfo{year}{2014}\natexlab{}.
\newblock \showarticletitle{A multidialectal parallel corpus of Arabic}. In
  \bibinfo{booktitle}{\emph{LREC 2014}}. European Language Resources
  Association (ELRA), \bibinfo{pages}{1240--1245}.
\newblock


\bibitem[Bouamor et~al\mbox{.}(2018)]%
        {bouamor2018madar}
\bibfield{author}{\bibinfo{person}{Houda Bouamor}, \bibinfo{person}{Nizar
  Habash}, \bibinfo{person}{Mohammad Salameh}, \bibinfo{person}{Wajdi
  Zaghouani}, \bibinfo{person}{Owen Rambow}, \bibinfo{person}{Dana Abdulrahim},
  \bibinfo{person}{Ossama Obeid}, \bibinfo{person}{Salam Khalifa},
  \bibinfo{person}{Fadhl Eryani}, \bibinfo{person}{Alexander Erdmann},
  {et~al\mbox{.}}} \bibinfo{year}{2018}\natexlab{}.
\newblock \showarticletitle{The MADAR arabic dialect corpus and lexicon}. In
  \bibinfo{booktitle}{\emph{LREC}}.
\newblock


\bibitem[Boujelbane et~al\mbox{.}(2013)]%
        {boujelbane-etal-2013-building}
\bibfield{author}{\bibinfo{person}{Rahma Boujelbane}, \bibinfo{person}{Mariem
  Ellouze~khemekhem}, \bibinfo{person}{Siwar BenAyed}, {and}
  \bibinfo{person}{Lamia Hadrich~Belguith}.} \bibinfo{year}{2013}\natexlab{}.
\newblock \showarticletitle{Building bilingual lexicon to create Dialect
  {T}unisian corpora and adapt language model}. In
  \bibinfo{booktitle}{\emph{Second Workshop on Hybrid Approaches to
  Translation}}. \bibinfo{address}{Sofia, Bulgaria}, \bibinfo{pages}{88--93}.
\newblock
\urldef\tempurl%
\url{https://aclanthology.org/W13-2813}
\showURL{%
\tempurl}


\bibitem[Boujou et~al\mbox{.}(2021)]%
        {boujou2021open}
\bibfield{author}{\bibinfo{person}{ElMehdi Boujou}, \bibinfo{person}{Hamza
  Chataoui}, \bibinfo{person}{Abdellah~El Mekki}, \bibinfo{person}{Saad
  Benjelloun}, \bibinfo{person}{Ikram Chairi}, {and} \bibinfo{person}{Ismail
  Berrada}.} \bibinfo{year}{2021}\natexlab{}.
\newblock \showarticletitle{An open access nlp dataset for arabic dialects:
  Data collection, labeling, and model construction}.
\newblock \bibinfo{journal}{\emph{arXiv preprint arXiv:2102.11000}}
  (\bibinfo{year}{2021}).
\newblock


\bibitem[Bowers et~al\mbox{.}(2017)]%
        {bowers-etal-2017-morphological}
\bibfield{author}{\bibinfo{person}{Dustin Bowers}, \bibinfo{person}{Antti
  Arppe}, \bibinfo{person}{Jordan Lachler}, \bibinfo{person}{Sjur Moshagen},
  {and} \bibinfo{person}{Trond Trosterud}.} \bibinfo{year}{2017}\natexlab{}.
\newblock \showarticletitle{A Morphological Parser for Odawa}. In
  \bibinfo{booktitle}{\emph{Workshop on the Use of Computational Methods in the
  Study of Endangered Languages}}. \bibinfo{pages}{1--9}.
\newblock
\urldef\tempurl%
\url{https://doi.org/10.18653/v1/W17-0101}
\showDOI{\tempurl}


\bibitem[Burghardt et~al\mbox{.}(2016)]%
        {burghardt-etal-2016-creating}
\bibfield{author}{\bibinfo{person}{Manuel Burghardt}, \bibinfo{person}{Daniel
  Granvogl}, {and} \bibinfo{person}{Christian Wolff}.}
  \bibinfo{year}{2016}\natexlab{}.
\newblock \showarticletitle{Creating a Lexicon of {B}avarian Dialect by Means
  of {F}acebook Language Data and Crowdsourcing}. In
  \bibinfo{booktitle}{\emph{LREC}}. \bibinfo{pages}{2029--2033}.
\newblock
\urldef\tempurl%
\url{https://aclanthology.org/L16-1321}
\showURL{%
\tempurl}


\bibitem[Chakrabarty et~al\mbox{.}({[n.\,d.]})]%
        {chakrabarty-etal-2022-featurebart}
\bibfield{author}{\bibinfo{person}{Abhisek Chakrabarty}, \bibinfo{person}{Raj
  Dabre}, \bibinfo{person}{Chenchen Ding}, \bibinfo{person}{Hideki Tanaka},
  \bibinfo{person}{Masao Utiyama}, {and} \bibinfo{person}{Eiichiro Sumita}.}
  \bibinfo{year}{[n.\,d.]}\natexlab{}.
\newblock \showarticletitle{{F}eature{BART}: Feature Based Sequence-to-Sequence
  Pre-Training for Low-Resource {NMT}}. In \bibinfo{booktitle}{\emph{COLING}}.
\newblock


\bibitem[Chakrabarty et~al\mbox{.}(2020)]%
        {chakrabarty-etal-2020-improving}
\bibfield{author}{\bibinfo{person}{Abhisek Chakrabarty}, \bibinfo{person}{Raj
  Dabre}, \bibinfo{person}{Chenchen Ding}, \bibinfo{person}{Masao Utiyama},
  {and} \bibinfo{person}{Eiichiro Sumita}.} \bibinfo{year}{2020}\natexlab{}.
\newblock \showarticletitle{Improving Low-Resource {NMT} through Relevance
  Based Linguistic Features Incorporation}. In
  \bibinfo{booktitle}{\emph{COLING}}. \bibinfo{publisher}{International
  Committee on Computational Linguistics}, \bibinfo{address}{Barcelona, Spain
  (Online)}, \bibinfo{pages}{4263--4274}.
\newblock
\urldef\tempurl%
\url{https://doi.org/10.18653/v1/2020.coling-main.376}
\showDOI{\tempurl}


\bibitem[Chaves et~al\mbox{.}(2022)]%
        {chaves2022chatbots}
\bibfield{author}{\bibinfo{person}{Ana~Paula Chaves}, \bibinfo{person}{Jesse
  Egbert}, \bibinfo{person}{Toby Hocking}, \bibinfo{person}{Eck Doerry}, {and}
  \bibinfo{person}{Marco~Aurelio Gerosa}.} \bibinfo{year}{2022}\natexlab{}.
\newblock \showarticletitle{Chatbots language design: The influence of language
  variation on user experience with tourist assistant chatbots}.
\newblock \bibinfo{journal}{\emph{ACM Transactions on Computer-Human
  Interaction}} \bibinfo{volume}{29}, \bibinfo{number}{2}
  (\bibinfo{year}{2022}), \bibinfo{pages}{1--38}.
\newblock


\bibitem[Chiang et~al\mbox{.}(2006)]%
        {chiang2006parsing}
\bibfield{author}{\bibinfo{person}{David Chiang}, \bibinfo{person}{Mona Diab},
  \bibinfo{person}{Nizar Habash}, \bibinfo{person}{Owen Rambow}, {and}
  \bibinfo{person}{Safiullah Shareef}.} \bibinfo{year}{2006}\natexlab{}.
\newblock \showarticletitle{Parsing arabic dialects}. In
  \bibinfo{booktitle}{\emph{11th Conference of the European Chapter of the
  Association for Computational Linguistics}}. \bibinfo{pages}{369--376}.
\newblock


\bibitem[Chifu et~al\mbox{.}(2024)]%
        {chifu-etal-2024-vardial}
\bibfield{author}{\bibinfo{person}{Adrian-Gabriel Chifu},
  \bibinfo{person}{Goran Glava{\v{s}}}, \bibinfo{person}{Radu~Tudor Ionescu},
  \bibinfo{person}{Nikola Ljube{\v{s}}i{\'c}}, \bibinfo{person}{Aleksandra
  Mileti{\'c}}, \bibinfo{person}{Filip Mileti{\'c}}, \bibinfo{person}{Yves
  Scherrer}, {and} \bibinfo{person}{Ivan Vuli{\'c}}.}
  \bibinfo{year}{2024}\natexlab{}.
\newblock \showarticletitle{{V}ar{D}ial Evaluation Campaign 2024: Commonsense
  Reasoning in Dialects and Multi-Label Similar Language Identification}. In
  \bibinfo{booktitle}{\emph{VarDial}}. \bibinfo{publisher}{Association for
  Computational Linguistics}, \bibinfo{address}{Mexico City, Mexico},
  \bibinfo{pages}{1--15}.
\newblock


\bibitem[Chitturi and Hansen(2008)]%
        {chitturi-hansen-2008-dialect}
\bibfield{author}{\bibinfo{person}{Rahul Chitturi} {and} \bibinfo{person}{John
  Hansen}.} \bibinfo{year}{2008}\natexlab{}.
\newblock \showarticletitle{Dialect Classification for Online Podcasts Fusing
  Acoustic and Language Based Structural and Semantic Information}. In
  \bibinfo{booktitle}{\emph{ACL}}. \bibinfo{pages}{21--24}.
\newblock
\urldef\tempurl%
\url{https://aclanthology.org/P08-2006}
\showURL{%
\tempurl}


\bibitem[Chow and Bond(2022)]%
        {chow2022singlish}
\bibfield{author}{\bibinfo{person}{Siew~Yeng Chow} {and}
  \bibinfo{person}{Francis Bond}.} \bibinfo{year}{2022}\natexlab{}.
\newblock \showarticletitle{Singlish where got rules one? constructing a
  computational grammar for Singlish}. In \bibinfo{booktitle}{\emph{LREC}}.
  \bibinfo{pages}{5243--5250}.
\newblock


\bibitem[Coats(2022)]%
        {coats-2022-corpus}
\bibfield{author}{\bibinfo{person}{Steven Coats}.}
  \bibinfo{year}{2022}\natexlab{}.
\newblock \showarticletitle{The Corpus of {A}ustralian and {N}ew {Z}ealand
  Spoken {E}nglish: A new resource of naturalistic speech transcripts}. In
  \bibinfo{booktitle}{\emph{Australasian Language Technology Association
  Workshop}}. \bibinfo{pages}{1--5}.
\newblock
\urldef\tempurl%
\url{https://aclanthology.org/2022.alta-1.1}
\showURL{%
\tempurl}


\bibitem[Coats(2023)]%
        {coats2023double}
\bibfield{author}{\bibinfo{person}{Steven Coats}.}
  \bibinfo{year}{2023}\natexlab{}.
\newblock \showarticletitle{Double modals in contemporary British and Irish
  speech}.
\newblock \bibinfo{journal}{\emph{English Language \& Linguistics}}
  \bibinfo{volume}{27}, \bibinfo{number}{4} (\bibinfo{year}{2023}),
  \bibinfo{pages}{693--718}.
\newblock


\bibitem[Contarino(2021)]%
        {contarino2021neural}
\bibfield{author}{\bibinfo{person}{Antonio Contarino}.}
  \bibinfo{year}{2021}\natexlab{}.
\newblock \bibinfo{title}{Neural machine translation adaptation and automatic
  terminology evaluation: a case study on Italian and South Tyrolean German
  legal texts}.
\newblock
\newblock


\bibitem[Cotterell and Callison-Burch(2014)]%
        {cotterell2014multi}
\bibfield{author}{\bibinfo{person}{Ryan Cotterell} {and} \bibinfo{person}{Chris
  Callison-Burch}.} \bibinfo{year}{2014}\natexlab{}.
\newblock \showarticletitle{A Multi-Dialect, Multi-Genre Corpus of Informal
  Written Arabic.}. In \bibinfo{booktitle}{\emph{LREC}}.
  \bibinfo{pages}{241--245}.
\newblock


\bibitem[Cox(2006)]%
        {cox2006acoustic}
\bibfield{author}{\bibinfo{person}{Felicity Cox}.}
  \bibinfo{year}{2006}\natexlab{}.
\newblock \showarticletitle{The acoustic characteristics of/hVd/vowels in the
  speech of some Australian teenagers}.
\newblock \bibinfo{journal}{\emph{Australian journal of linguistics}}
  \bibinfo{volume}{26}, \bibinfo{number}{2} (\bibinfo{year}{2006}),
  \bibinfo{pages}{147--179}.
\newblock


\bibitem[Cox and Palethorpe(2007)]%
        {cox2007australian}
\bibfield{author}{\bibinfo{person}{Felicity Cox} {and}
  \bibinfo{person}{Sallyanne Palethorpe}.} \bibinfo{year}{2007}\natexlab{}.
\newblock \showarticletitle{Australian English}.
\newblock \bibinfo{journal}{\emph{Journal of the International Phonetic
  Association}} \bibinfo{volume}{37}, \bibinfo{number}{3}
  (\bibinfo{year}{2007}), \bibinfo{pages}{341--350}.
\newblock


\bibitem[Criscuolo and Aluisio(2017)]%
        {criscuolo2017discriminating}
\bibfield{author}{\bibinfo{person}{Marcelo Criscuolo} {and}
  \bibinfo{person}{Sandra Aluisio}.} \bibinfo{year}{2017}\natexlab{}.
\newblock \showarticletitle{Discriminating between similar languages with
  word-level convolutional neural networks}. In
  \bibinfo{booktitle}{\emph{VarDial}}. \bibinfo{pages}{124--130}.
\newblock


\bibitem[Dabre et~al\mbox{.}(2024)]%
        {dabre-etal-2024-machine}
\bibfield{author}{\bibinfo{person}{Raj Dabre}, \bibinfo{person}{Mary Dabre},
  {and} \bibinfo{person}{Teresa Pereira}.} \bibinfo{year}{2024}\natexlab{}.
\newblock \showarticletitle{Machine Translation Of {M}arathi Dialects: A Case
  Study Of Kadodi}. In \bibinfo{booktitle}{\emph{Eleventh Workshop on Asian
  Translation (WAT 2024)}}. \bibinfo{publisher}{Association for Computational
  Linguistics}, \bibinfo{address}{Miami, Florida, USA},
  \bibinfo{pages}{36--44}.
\newblock
\urldef\tempurl%
\url{https://aclanthology.org/2024.wat-1.3}
\showURL{%
\tempurl}


\bibitem[Dacon et~al\mbox{.}(2022)]%
        {dacon-etal-2022-evaluating}
\bibfield{author}{\bibinfo{person}{Jamell Dacon}, \bibinfo{person}{Haochen
  Liu}, {and} \bibinfo{person}{Jiliang Tang}.} \bibinfo{year}{2022}\natexlab{}.
\newblock \showarticletitle{Evaluating and Mitigating Inherent Linguistic Bias
  of {A}frican {A}merican {E}nglish through Inference}. In
  \bibinfo{booktitle}{\emph{COLING}}. \bibinfo{pages}{1442--1454}.
\newblock
\urldef\tempurl%
\url{https://aclanthology.org/2022.coling-1.124}
\showURL{%
\tempurl}


\bibitem[Darwish et~al\mbox{.}(2021)]%
        {darwish2021panoramic}
\bibfield{author}{\bibinfo{person}{Kareem Darwish}, \bibinfo{person}{Nizar
  Habash}, \bibinfo{person}{Mourad Abbas}, \bibinfo{person}{Hend Al-Khalifa},
  \bibinfo{person}{Huseein~T Al-Natsheh}, \bibinfo{person}{Houda Bouamor},
  \bibinfo{person}{Karim Bouzoubaa}, \bibinfo{person}{Violetta Cavalli-Sforza},
  \bibinfo{person}{Samhaa~R El-Beltagy}, \bibinfo{person}{Wassim El-Hajj},
  {et~al\mbox{.}}} \bibinfo{year}{2021}\natexlab{}.
\newblock \showarticletitle{A panoramic survey of natural language processing
  in the Arab world}.
\newblock \bibinfo{journal}{\emph{Commun. ACM}} \bibinfo{volume}{64},
  \bibinfo{number}{4} (\bibinfo{year}{2021}), \bibinfo{pages}{72--81}.
\newblock


\bibitem[Darwish et~al\mbox{.}(2018)]%
        {darwish2018multi}
\bibfield{author}{\bibinfo{person}{Kareem Darwish}, \bibinfo{person}{Hamdy
  Mubarak}, \bibinfo{person}{Mohamed Eldesouki}, \bibinfo{person}{Ahmed
  Abdelali}, \bibinfo{person}{Younes Samih}, \bibinfo{person}{Randah Alharbi},
  \bibinfo{person}{Mohammed Attia}, \bibinfo{person}{Walid Magdy}, {and}
  \bibinfo{person}{Laura Kallmeyer}.} \bibinfo{year}{2018}\natexlab{}.
\newblock \showarticletitle{Multi-dialect Arabic POS tagging: a CRF approach}.
  In \bibinfo{booktitle}{\emph{LREC}}. European Language Resources Association
  (ELRA), \bibinfo{pages}{93--98}.
\newblock


\bibitem[Darwish et~al\mbox{.}(2014)]%
        {darwish-etal-2014-verifiably}
\bibfield{author}{\bibinfo{person}{Kareem Darwish}, \bibinfo{person}{Hassan
  Sajjad}, {and} \bibinfo{person}{Hamdy Mubarak}.}
  \bibinfo{year}{2014}\natexlab{}.
\newblock \showarticletitle{Verifiably Effective {A}rabic Dialect
  Identification}. In \bibinfo{booktitle}{\emph{EMNLP}}.
  \bibinfo{pages}{1465--1468}.
\newblock
\urldef\tempurl%
\url{https://doi.org/10.3115/v1/D14-1154}
\showDOI{\tempurl}


\bibitem[De~Camillis et~al\mbox{.}(2023)]%
        {de2023mt}
\bibfield{author}{\bibinfo{person}{Flavia De~Camillis},
  \bibinfo{person}{Egon~Waldemar Stemle}, \bibinfo{person}{Elena Chiocchetti},
  {and} \bibinfo{person}{Francesco Fernicola}.}
  \bibinfo{year}{2023}\natexlab{}.
\newblock \showarticletitle{The MT@ BZ Corpus: machine translation \& legal
  language}. In \bibinfo{booktitle}{\emph{Annual Conference of the European
  Association for Machine Translation}}. \bibinfo{pages}{171--180}.
\newblock


\bibitem[Demszky et~al\mbox{.}(2020)]%
        {demszky2020learning}
\bibfield{author}{\bibinfo{person}{Dorottya Demszky}, \bibinfo{person}{Devyani
  Sharma}, \bibinfo{person}{Jonathan~H Clark}, \bibinfo{person}{Vinodkumar
  Prabhakaran}, {and} \bibinfo{person}{Jacob Eisenstein}.}
  \bibinfo{year}{2020}\natexlab{}.
\newblock \showarticletitle{Learning to recognize dialect features}.
\newblock \bibinfo{journal}{\emph{arXiv preprint arXiv:2010.12707}}
  (\bibinfo{year}{2020}).
\newblock


\bibitem[Demszky et~al\mbox{.}(2021)]%
        {demszky2021learning}
\bibfield{author}{\bibinfo{person}{Dorottya Demszky}, \bibinfo{person}{Devyani
  Sharma}, \bibinfo{person}{Jonathan~H Clark}, \bibinfo{person}{Vinodkumar
  Prabhakaran}, {and} \bibinfo{person}{Jacob Eisenstein}.}
  \bibinfo{year}{2021}\natexlab{}.
\newblock \showarticletitle{Learning to Recognize Dialect Features}. In
  \bibinfo{booktitle}{\emph{NAACL}}. \bibinfo{pages}{2315--2338}.
\newblock


\bibitem[Diab et~al\mbox{.}(2010)]%
        {diab2010colaba}
\bibfield{author}{\bibinfo{person}{Mona Diab}, \bibinfo{person}{Nizar Habash},
  \bibinfo{person}{Owen Rambow}, \bibinfo{person}{Mohamed Altantawy}, {and}
  \bibinfo{person}{Yassine Benajiba}.} \bibinfo{year}{2010}\natexlab{}.
\newblock \showarticletitle{COLABA: Arabic dialect annotation and processing}.
  In \bibinfo{booktitle}{\emph{Workshop on semitic language processing}}.
  \bibinfo{pages}{66--74}.
\newblock


\bibitem[Dibas et~al\mbox{.}(2022)]%
        {dibas2022maknuune}
\bibfield{author}{\bibinfo{person}{Shahd Dibas}, \bibinfo{person}{Christian
  Khairallah}, \bibinfo{person}{Nizar Habash}, \bibinfo{person}{Omar~Fayez
  Sadi}, \bibinfo{person}{Tariq Sairafy}, \bibinfo{person}{Karmel Sarabta},
  {and} \bibinfo{person}{Abrar Ardah}.} \bibinfo{year}{2022}\natexlab{}.
\newblock \showarticletitle{Maknuune: A Large Open Palestinian Arabic Lexicon}.
  In \bibinfo{booktitle}{\emph{Arabic Natural Language Processing Workshop}}.
  Association for Computational Linguistics (ACL), \bibinfo{pages}{131--141}.
\newblock


\bibitem[Dinh et~al\mbox{.}(2024)]%
        {dinh-etal-2024-multi}
\bibfield{author}{\bibinfo{person}{Nguyen~Van Dinh}, \bibinfo{person}{Thanh~Chi
  Dang}, \bibinfo{person}{Luan Thanh~Nguyen}, {and} \bibinfo{person}{Kiet~Van
  Nguyen}.} \bibinfo{year}{2024}\natexlab{}.
\newblock \showarticletitle{Multi-Dialect {V}ietnamese: Task, Dataset, Baseline
  Models and Challenges}. In \bibinfo{booktitle}{\emph{EMNLP}}.
  \bibinfo{publisher}{Association for Computational Linguistics},
  \bibinfo{address}{Miami, Florida, USA}, \bibinfo{pages}{7476--7498}.
\newblock
\urldef\tempurl%
\url{https://aclanthology.org/2024.emnlp-main.426}
\showURL{%
\tempurl}


\bibitem[Do{\u{g}}ru{\"o}z and Nakov(2014)]%
        {dogruoz-nakov-2014-predicting}
\bibfield{author}{\bibinfo{person}{A.~Seza Do{\u{g}}ru{\"o}z} {and}
  \bibinfo{person}{Preslav Nakov}.} \bibinfo{year}{2014}\natexlab{}.
\newblock \showarticletitle{Predicting Dialect Variation in Immigrant Contexts
  Using Light Verb Constructions}. In \bibinfo{booktitle}{\emph{EMNLP}}.
  \bibinfo{pages}{1391--1395}.
\newblock
\urldef\tempurl%
\url{https://doi.org/10.3115/v1/D14-1145}
\showDOI{\tempurl}


\bibitem[Dou et~al\mbox{.}(2023)]%
        {dou2023multispider}
\bibfield{author}{\bibinfo{person}{Longxu Dou}, \bibinfo{person}{Yan Gao},
  \bibinfo{person}{Mingyang Pan}, \bibinfo{person}{Dingzirui Wang},
  \bibinfo{person}{Wanxiang Che}, \bibinfo{person}{Dechen Zhan}, {and}
  \bibinfo{person}{Jian-Guang Lou}.} \bibinfo{year}{2023}\natexlab{}.
\newblock \showarticletitle{MultiSpider: towards benchmarking multilingual
  text-to-SQL semantic parsing}. In \bibinfo{booktitle}{\emph{AAAI}},
  Vol.~\bibinfo{volume}{37}. \bibinfo{pages}{12745--12753}.
\newblock


\bibitem[Ducel et~al\mbox{.}(2022)]%
        {ducel2022we}
\bibfield{author}{\bibinfo{person}{Fanny Ducel}, \bibinfo{person}{Kar{\"e}n
  Fort}, \bibinfo{person}{Ga{\"e}l Lejeune}, {and} \bibinfo{person}{Yves
  Lepage}.} \bibinfo{year}{2022}\natexlab{}.
\newblock \showarticletitle{Do we Name the Languages we Study? The\# BenderRule
  in LREC and ACL articles}. In \bibinfo{booktitle}{\emph{LREC}}.
  \bibinfo{pages}{564--573}.
\newblock


\bibitem[Dunn(2019)]%
        {dunn-2019-modeling}
\bibfield{author}{\bibinfo{person}{Jonathan Dunn}.}
  \bibinfo{year}{2019}\natexlab{}.
\newblock \showarticletitle{Modeling Global Syntactic Variation in {E}nglish
  Using Dialect Classification}. In \bibinfo{booktitle}{\emph{VarDial}}.
  \bibinfo{publisher}{Association for Computational Linguistics},
  \bibinfo{pages}{42--53}.
\newblock
\urldef\tempurl%
\url{https://doi.org/10.18653/v1/W19-1405}
\showDOI{\tempurl}


\bibitem[Dunn and Adams(2020)]%
        {dunn-adams-2020-geographically}
\bibfield{author}{\bibinfo{person}{Jonathan Dunn} {and} \bibinfo{person}{Ben
  Adams}.} \bibinfo{year}{2020}\natexlab{}.
\newblock \showarticletitle{Geographically-Balanced {G}igaword Corpora for 50
  Language Varieties}. In \bibinfo{booktitle}{\emph{LREC}}.
  \bibinfo{publisher}{European Language Resources Association},
  \bibinfo{address}{Marseille, France}, \bibinfo{pages}{2528--2536}.
\newblock
\showISBNx{979-10-95546-34-4}
\urldef\tempurl%
\url{https://aclanthology.org/2020.lrec-1.308}
\showURL{%
\tempurl}


\bibitem[Eggleston and O{'}Connor(2022)]%
        {eggleston-oconnor-2022-cross}
\bibfield{author}{\bibinfo{person}{Chloe Eggleston} {and}
  \bibinfo{person}{Brendan O{'}Connor}.} \bibinfo{year}{2022}\natexlab{}.
\newblock \showarticletitle{Cross-Dialect Social Media Dependency Parsing for
  Social Scientific Entity Attribute Analysis}. In
  \bibinfo{booktitle}{\emph{Workshop on Noisy User-generated Text (W-NUT
  2022)}}. \bibinfo{publisher}{Association for Computational Linguistics},
  \bibinfo{pages}{38--50}.
\newblock
\urldef\tempurl%
\url{https://aclanthology.org/2022.wnut-1.4}
\showURL{%
\tempurl}


\bibitem[Eisenstein et~al\mbox{.}(2023)]%
        {eisenstein23_interspeech}
\bibfield{author}{\bibinfo{person}{Jacob Eisenstein},
  \bibinfo{person}{Vinodkumar Prabhakaran}, \bibinfo{person}{Clara Rivera},
  \bibinfo{person}{Dorottya Demszky}, {and} \bibinfo{person}{Devyani Sharma}.}
  \bibinfo{year}{2023}\natexlab{}.
\newblock \showarticletitle{{MD3: The Multi-Dialect Dataset of Dialogues}}. In
  \bibinfo{booktitle}{\emph{Proc. INTERSPEECH 2023}}.
  \bibinfo{pages}{4059--4063}.
\newblock
\urldef\tempurl%
\url{https://doi.org/10.21437/Interspeech.2023-2150}
\showDOI{\tempurl}


\bibitem[El~Mekki et~al\mbox{.}(2021)]%
        {el-mekki-etal-2021-domain}
\bibfield{author}{\bibinfo{person}{Abdellah El~Mekki},
  \bibinfo{person}{Abdelkader El~Mahdaouy}, \bibinfo{person}{Ismail Berrada},
  {and} \bibinfo{person}{Ahmed Khoumsi}.} \bibinfo{year}{2021}\natexlab{}.
\newblock \showarticletitle{Domain Adaptation for {A}rabic Cross-Domain and
  Cross-Dialect Sentiment Analysis from Contextualized Word Embedding}. In
  \bibinfo{booktitle}{\emph{NAACL}}. \bibinfo{pages}{2824--2837}.
\newblock
\urldef\tempurl%
\url{https://doi.org/10.18653/v1/2021.naacl-main.226}
\showDOI{\tempurl}


\bibitem[Elfardy and Diab(2012)]%
        {elfardy2012simplified}
\bibfield{author}{\bibinfo{person}{Heba Elfardy} {and} \bibinfo{person}{Mona~T
  Diab}.} \bibinfo{year}{2012}\natexlab{}.
\newblock \showarticletitle{Simplified guidelines for the creation of Large
  Scale Dialectal Arabic Annotations.}. In \bibinfo{booktitle}{\emph{LREC}}.
  \bibinfo{pages}{371--378}.
\newblock


\bibitem[Elmadany et~al\mbox{.}(2018a)]%
        {elmadany2018improving}
\bibfield{author}{\bibinfo{person}{AbdelRahim Elmadany},
  \bibinfo{person}{Sherif Abdou}, {and} \bibinfo{person}{Mervat Gheith}.}
  \bibinfo{year}{2018}\natexlab{a}.
\newblock \showarticletitle{Improving Dialogue Act Classification for
  Spontaneous Arabic Speech and Instant Messages at Utterance Level}. In
  \bibinfo{booktitle}{\emph{LREC}}.
\newblock


\bibitem[Elmadany et~al\mbox{.}(2018b)]%
        {elmadany2018arsas}
\bibfield{author}{\bibinfo{person}{A Elmadany}, \bibinfo{person}{Hamdy
  Mubarak}, {and} \bibinfo{person}{Walid Magdy}.}
  \bibinfo{year}{2018}\natexlab{b}.
\newblock \showarticletitle{Arsas: An arabic speech-act and sentiment corpus of
  tweets}.
\newblock \bibinfo{journal}{\emph{OSACT}}  \bibinfo{volume}{3}
  (\bibinfo{year}{2018}), \bibinfo{pages}{20}.
\newblock


\bibitem[Elnagar et~al\mbox{.}(2021)]%
        {elnagar2021systematic}
\bibfield{author}{\bibinfo{person}{Ashraf Elnagar}, \bibinfo{person}{Sane~M
  Yagi}, \bibinfo{person}{Ali~Bou Nassif}, \bibinfo{person}{Ismail Shahin},
  {and} \bibinfo{person}{Said~A Salloum}.} \bibinfo{year}{2021}\natexlab{}.
\newblock \showarticletitle{Systematic literature review of dialectal Arabic:
  identification and detection}.
\newblock \bibinfo{journal}{\emph{IEEE Access}}  \bibinfo{volume}{9}
  (\bibinfo{year}{2021}), \bibinfo{pages}{31010--31042}.
\newblock


\bibitem[Elsaid et~al\mbox{.}(2022)]%
        {elsaid2022comprehensive}
\bibfield{author}{\bibinfo{person}{Asmaa Elsaid}, \bibinfo{person}{Ammar
  Mohammed}, \bibinfo{person}{Lamiaa~Fattouh Ibrahim}, {and}
  \bibinfo{person}{Mohammed~M Sakre}.} \bibinfo{year}{2022}\natexlab{}.
\newblock \showarticletitle{A comprehensive review of arabic text
  summarization}.
\newblock \bibinfo{journal}{\emph{IEEE Access}}  \bibinfo{volume}{10}
  (\bibinfo{year}{2022}), \bibinfo{pages}{38012--38030}.
\newblock


\bibitem[Erdmann et~al\mbox{.}(2017)]%
        {erdmann-etal-2017-low}
\bibfield{author}{\bibinfo{person}{Alexander Erdmann}, \bibinfo{person}{Nizar
  Habash}, \bibinfo{person}{Dima Taji}, {and} \bibinfo{person}{Houda Bouamor}.}
  \bibinfo{year}{2017}\natexlab{}.
\newblock \showarticletitle{Low Resourced Machine Translation via
  Morpho-syntactic Modeling: The Case of Dialectal {A}rabic}. In
  \bibinfo{booktitle}{\emph{Machine Translation Summit XVI: Research Track}}.
  \bibinfo{address}{Nagoya Japan}, \bibinfo{pages}{185--200}.
\newblock
\urldef\tempurl%
\url{https://aclanthology.org/2017.mtsummit-papers.15}
\showURL{%
\tempurl}


\bibitem[Erdmann et~al\mbox{.}(2018)]%
        {erdmann2018addressing}
\bibfield{author}{\bibinfo{person}{Alexander Erdmann}, \bibinfo{person}{Nasser
  Zalmout}, {and} \bibinfo{person}{Nizar Habash}.}
  \bibinfo{year}{2018}\natexlab{}.
\newblock \showarticletitle{Addressing noise in multidialectal word
  embeddings}. In \bibinfo{booktitle}{\emph{ACL}}. \bibinfo{pages}{558--565}.
\newblock


\bibitem[Eskander et~al\mbox{.}(2016)]%
        {eskander2016creating}
\bibfield{author}{\bibinfo{person}{Ramy Eskander}, \bibinfo{person}{Nizar
  Habash}, \bibinfo{person}{Owen Rambow}, {and} \bibinfo{person}{Arfath
  Pasha}.} \bibinfo{year}{2016}\natexlab{}.
\newblock \showarticletitle{Creating resources for Dialectal Arabic from a
  single annotation: A case study on Egyptian and Levantine}. In
  \bibinfo{booktitle}{\emph{COLING}}. \bibinfo{pages}{3455--3465}.
\newblock


\bibitem[Estival et~al\mbox{.}(2014)]%
        {estival-etal-2014-austalk}
\bibfield{author}{\bibinfo{person}{Dominique Estival}, \bibinfo{person}{Steve
  Cassidy}, \bibinfo{person}{Felicity Cox}, {and} \bibinfo{person}{Denis
  Burnham}.} \bibinfo{year}{2014}\natexlab{}.
\newblock \showarticletitle{{A}us{T}alk: an audio-visual corpus of {A}ustralian
  {E}nglish}. In \bibinfo{booktitle}{\emph{Ninth International Conference on
  Language Resources and Evaluation ({LREC}'14)}}. \bibinfo{publisher}{European
  Language Resources Association (ELRA)}, \bibinfo{pages}{3105--3109}.
\newblock
\urldef\tempurl%
\url{http://www.lrec-conf.org/proceedings/lrec2014/pdf/520_Paper.pdf}
\showURL{%
\tempurl}


\bibitem[Fadhil et~al\mbox{.}(2019)]%
        {fadhil2019ollobot}
\bibfield{author}{\bibinfo{person}{Ahmed Fadhil} {et~al\mbox{.}}}
  \bibinfo{year}{2019}\natexlab{}.
\newblock \showarticletitle{OlloBot-towards a text-based arabic health
  conversational agent: Evaluation and results}. In
  \bibinfo{booktitle}{\emph{RANLP}}. \bibinfo{pages}{295--303}.
\newblock


\bibitem[Faisal et~al\mbox{.}(2024)]%
        {faisal2024dialectbenchnlpbenchmarkdialects}
\bibfield{author}{\bibinfo{person}{Fahim Faisal}, \bibinfo{person}{Orevaoghene
  Ahia}, \bibinfo{person}{Aarohi Srivastava}, \bibinfo{person}{Kabir Ahuja},
  \bibinfo{person}{David Chiang}, \bibinfo{person}{Yulia Tsvetkov}, {and}
  \bibinfo{person}{Antonios Anastasopoulos}.} \bibinfo{year}{2024}\natexlab{}.
\newblock \showarticletitle{DIALECTBENCH: A NLP Benchmark for Dialects,
  Varieties, and Closely-Related Languages}. In
  \bibinfo{booktitle}{\emph{ACL}}.
\newblock


\bibitem[Farha and Magdy(2022)]%
        {farha2022effect}
\bibfield{author}{\bibinfo{person}{Ibrahim~Abu Farha} {and}
  \bibinfo{person}{Walid Magdy}.} \bibinfo{year}{2022}\natexlab{}.
\newblock \showarticletitle{The Effect of Arabic Dialect Familiarity on Data
  Annotation}. In \bibinfo{booktitle}{\emph{Arabic Natural Language Processing
  Workshop}}. \bibinfo{pages}{399--408}.
\newblock


\bibitem[Fleisig et~al\mbox{.}(2024)]%
        {fleisig-etal-2024-linguistic}
\bibfield{author}{\bibinfo{person}{Eve Fleisig}, \bibinfo{person}{Genevieve
  Smith}, \bibinfo{person}{Madeline Bossi}, \bibinfo{person}{Ishita Rustagi},
  \bibinfo{person}{Xavier Yin}, {and} \bibinfo{person}{Dan Klein}.}
  \bibinfo{year}{2024}\natexlab{}.
\newblock \showarticletitle{Linguistic Bias in {C}hat{GPT}: Language Models
  Reinforce Dialect Discrimination}. In \bibinfo{booktitle}{\emph{EMNLP}}.
  \bibinfo{publisher}{Association for Computational Linguistics},
  \bibinfo{address}{Miami, Florida, USA}, \bibinfo{pages}{13541--13564}.
\newblock
\urldef\tempurl%
\url{https://aclanthology.org/2024.emnlp-main.750}
\showURL{%
\tempurl}


\bibitem[F{\o}lstad and Brandtzaeg(2020)]%
        {folstad2020users}
\bibfield{author}{\bibinfo{person}{Asbj{\o}rn F{\o}lstad} {and}
  \bibinfo{person}{Petter~Bae Brandtzaeg}.} \bibinfo{year}{2020}\natexlab{}.
\newblock \showarticletitle{Users' experiences with chatbots: findings from a
  questionnaire study}.
\newblock \bibinfo{journal}{\emph{Quality and User Experience}}
  \bibinfo{volume}{5}, \bibinfo{number}{1} (\bibinfo{year}{2020}),
  \bibinfo{pages}{3}.
\newblock


\bibitem[Fuad and Al-Yahya(2022)]%
        {fuad2022araconv}
\bibfield{author}{\bibinfo{person}{Ahlam Fuad} {and} \bibinfo{person}{Maha
  Al-Yahya}.} \bibinfo{year}{2022}\natexlab{}.
\newblock \showarticletitle{AraConv: Developing an Arabic task-oriented
  dialogue system using multi-lingual transformer model mT5}.
\newblock \bibinfo{journal}{\emph{Applied Sciences}} \bibinfo{volume}{12},
  \bibinfo{number}{4} (\bibinfo{year}{2022}), \bibinfo{pages}{1881}.
\newblock


\bibitem[Gaman et~al\mbox{.}(2020)]%
        {gaman-etal-2020-report}
\bibfield{author}{\bibinfo{person}{Mihaela Gaman}, \bibinfo{person}{Dirk Hovy},
  \bibinfo{person}{Radu~Tudor Ionescu}, \bibinfo{person}{Heidi Jauhiainen},
  \bibinfo{person}{Tommi Jauhiainen}, \bibinfo{person}{Krister Lind{\'e}n},
  \bibinfo{person}{Nikola Ljube{\v{s}}i{\'c}}, \bibinfo{person}{Niko Partanen},
  \bibinfo{person}{Christoph Purschke}, \bibinfo{person}{Yves Scherrer}, {and}
  \bibinfo{person}{Marcos Zampieri}.} \bibinfo{year}{2020}\natexlab{}.
\newblock \showarticletitle{A Report on the {V}ar{D}ial Evaluation Campaign
  2020}. In \bibinfo{booktitle}{\emph{VarDial}}.
  \bibinfo{publisher}{International Committee on Computational Linguistics
  (ICCL)}, \bibinfo{address}{Barcelona, Spain (Online)},
  \bibinfo{pages}{1--14}.
\newblock
\urldef\tempurl%
\url{https://aclanthology.org/2020.vardial-1.1}
\showURL{%
\tempurl}


\bibitem[Goebl(1993)]%
        {goebl1993dialectometry}
\bibfield{author}{\bibinfo{person}{Hans Goebl}.}
  \bibinfo{year}{1993}\natexlab{}.
\newblock \showarticletitle{Dialectometry: a short overview of the principles
  and practice of quantitative classification of linguistic atlas data}. In
  \bibinfo{booktitle}{\emph{Contributions to Quantitative Linguistics: First
  International Conference on Quantitative Linguistics, QUALICO, Trier, 1991}}.
  Springer, \bibinfo{pages}{277--315}.
\newblock


\bibitem[Goswami et~al\mbox{.}(2020)]%
        {goswami2020unsupervised}
\bibfield{author}{\bibinfo{person}{Koustava Goswami}, \bibinfo{person}{Rajdeep
  Sarkar}, \bibinfo{person}{Bharathi~Raja Chakravarthi},
  \bibinfo{person}{Theodorus Fransen}, {and} \bibinfo{person}{John~Philip
  McCrae}.} \bibinfo{year}{2020}\natexlab{}.
\newblock \showarticletitle{Unsupervised deep language and dialect
  identification for short texts}. In \bibinfo{booktitle}{\emph{International
  Conference on Computational Linguistics}}. \bibinfo{pages}{1606--1617}.
\newblock


\bibitem[Goutte et~al\mbox{.}(2016)]%
        {goutte2016discriminating}
\bibfield{author}{\bibinfo{person}{Cyril Goutte}, \bibinfo{person}{Serge
  L{\'e}ger}, \bibinfo{person}{Shervin Malmasi}, {and} \bibinfo{person}{Marcos
  Zampieri}.} \bibinfo{year}{2016}\natexlab{}.
\newblock \showarticletitle{Discriminating similar languages: Evaluations and
  explorations}. In \bibinfo{booktitle}{\emph{LREC}}.
\newblock


\bibitem[Guellil et~al\mbox{.}(2021)]%
        {guellil-etal-2021-one}
\bibfield{author}{\bibinfo{person}{Imane Guellil}, \bibinfo{person}{Faical
  Azouaou}, \bibinfo{person}{Fodil Benali}, {and} \bibinfo{person}{Hachani
  Ala-Eddine}.} \bibinfo{year}{2021}\natexlab{}.
\newblock \showarticletitle{{ONE}: Toward {ONE} model, {ONE} algorithm, {ONE}
  corpus dedicated to sentiment analysis of {A}rabic/{A}rabizi and its
  dialects}. In \bibinfo{booktitle}{\emph{Workshop on Computational Approaches
  to Subjectivity, Sentiment and Social Media Analysis}}.
  \bibinfo{publisher}{Association for Computational Linguistics},
  \bibinfo{address}{Online}, \bibinfo{pages}{236--249}.
\newblock


\bibitem[Gumperz(1982)]%
        {gumperz1982discourse}
\bibfield{author}{\bibinfo{person}{John~J Gumperz}.}
  \bibinfo{year}{1982}\natexlab{}.
\newblock \bibinfo{booktitle}{\emph{Discourse strategies}}.
\newblock Number~1. \bibinfo{publisher}{Cambridge University Press}.
\newblock


\bibitem[Habash and Rambow(2006)]%
        {habash2006magead}
\bibfield{author}{\bibinfo{person}{Nizar Habash} {and} \bibinfo{person}{Owen
  Rambow}.} \bibinfo{year}{2006}\natexlab{}.
\newblock \showarticletitle{MAGEAD: A Morphological Analyzer and Generator for
  the Arabic Dialects}. In \bibinfo{booktitle}{\emph{ACL}}.
  \bibinfo{pages}{681--688}.
\newblock


\bibitem[Hamed et~al\mbox{.}(2022)]%
        {hamed-etal-2022-arzen}
\bibfield{author}{\bibinfo{person}{Injy Hamed}, \bibinfo{person}{Nizar Habash},
  \bibinfo{person}{Slim Abdennadher}, {and} \bibinfo{person}{Ngoc~Thang Vu}.}
  \bibinfo{year}{2022}\natexlab{}.
\newblock \showarticletitle{{A}rz{E}n-{ST}: A Three-way Speech Translation
  Corpus for Code-Switched {E}gyptian {A}rabic-{E}nglish}. In
  \bibinfo{booktitle}{\emph{Seventh Arabic Natural Language Processing Workshop
  (WANLP)}}. \bibinfo{publisher}{Association for Computational Linguistics},
  \bibinfo{address}{Abu Dhabi, United Arab Emirates (Hybrid)},
  \bibinfo{pages}{119--130}.
\newblock
\urldef\tempurl%
\url{https://doi.org/10.18653/v1/2022.wanlp-1.12}
\showDOI{\tempurl}


\bibitem[Hanani and Naser(2020)]%
        {hanani2020spoken}
\bibfield{author}{\bibinfo{person}{Abualsoud Hanani} {and}
  \bibinfo{person}{Rabee Naser}.} \bibinfo{year}{2020}\natexlab{}.
\newblock \showarticletitle{Spoken Arabic dialect recognition using X-vectors}.
\newblock \bibinfo{journal}{\emph{Natural Language Engineering}}
  \bibinfo{volume}{26}, \bibinfo{number}{6} (\bibinfo{year}{2020}),
  \bibinfo{pages}{691--700}.
\newblock


\bibitem[Harrat et~al\mbox{.}(2018)]%
        {harrat2018maghrebi}
\bibfield{author}{\bibinfo{person}{Salima Harrat}, \bibinfo{person}{Karima
  Meftouh}, {and} \bibinfo{person}{Kamel Sma{\"\i}li}.}
  \bibinfo{year}{2018}\natexlab{}.
\newblock \showarticletitle{Maghrebi Arabic dialect processing: an overview}.
\newblock \bibinfo{journal}{\emph{Journal of International Science and General
  Applications}}  \bibinfo{volume}{1} (\bibinfo{year}{2018}).
\newblock


\bibitem[Harris et~al\mbox{.}(2022)]%
        {harris2022exploring}
\bibfield{author}{\bibinfo{person}{Camille Harris}, \bibinfo{person}{Matan
  Halevy}, \bibinfo{person}{Ayanna Howard}, \bibinfo{person}{Amy Bruckman},
  {and} \bibinfo{person}{Diyi Yang}.} \bibinfo{year}{2022}\natexlab{}.
\newblock \showarticletitle{Exploring the role of grammar and word choice in
  bias toward african american english (aae) in hate speech classification}. In
  \bibinfo{booktitle}{\emph{2022 ACM Conference on Fairness, Accountability,
  and Transparency}}. \bibinfo{pages}{789--798}.
\newblock


\bibitem[Harris et~al\mbox{.}(2024)]%
        {harris-etal-2024-modeling}
\bibfield{author}{\bibinfo{person}{Camille Harris}, \bibinfo{person}{Chijioke
  Mgbahurike}, \bibinfo{person}{Neha Kumar}, {and} \bibinfo{person}{Diyi
  Yang}.} \bibinfo{year}{2024}\natexlab{}.
\newblock \showarticletitle{Modeling Gender and Dialect Bias in Automatic
  Speech Recognition}. In \bibinfo{booktitle}{\emph{Findings of EMNLP}}.
  \bibinfo{publisher}{Association for Computational Linguistics},
  \bibinfo{address}{Miami, Florida, USA}, \bibinfo{pages}{15166--15184}.
\newblock
\urldef\tempurl%
\url{https://aclanthology.org/2024.findings-emnlp.890}
\showURL{%
\tempurl}


\bibitem[Hassan et~al\mbox{.}(2017)]%
        {hassan-etal-2017-synthetic}
\bibfield{author}{\bibinfo{person}{Hany Hassan}, \bibinfo{person}{Mostafa
  Elaraby}, {and} \bibinfo{person}{Ahmed~Y. Tawfik}.}
  \bibinfo{year}{2017}\natexlab{}.
\newblock \showarticletitle{Synthetic Data for Neural Machine Translation of
  Spoken-Dialects}. In \bibinfo{booktitle}{\emph{14th International Conference
  on Spoken Language Translation}}. \bibinfo{publisher}{International Workshop
  on Spoken Language Translation}, \bibinfo{address}{Tokyo, Japan},
  \bibinfo{pages}{82--89}.
\newblock
\urldef\tempurl%
\url{https://aclanthology.org/2017.iwslt-1.12}
\showURL{%
\tempurl}


\bibitem[Hassani(2017)]%
        {hassani-2017-kurdish}
\bibfield{author}{\bibinfo{person}{Hossein Hassani}.}
  \bibinfo{year}{2017}\natexlab{}.
\newblock \showarticletitle{{K}urdish Interdialect Machine Translation}. In
  \bibinfo{booktitle}{\emph{Fourth Workshop on {NLP} for Similar Languages,
  Varieties and Dialects ({V}ar{D}ial)}}. \bibinfo{publisher}{ACL},
  \bibinfo{address}{Valencia, Spain}, \bibinfo{pages}{63--72}.
\newblock
\urldef\tempurl%
\url{https://doi.org/10.18653/v1/W17-1208}
\showDOI{\tempurl}


\bibitem[Haugen(1966)]%
        {haugen1966dialect}
\bibfield{author}{\bibinfo{person}{Einar Haugen}.}
  \bibinfo{year}{1966}\natexlab{}.
\newblock \showarticletitle{Dialect, language, nation 1}.
\newblock \bibinfo{journal}{\emph{American anthropologist}}
  \bibinfo{volume}{68}, \bibinfo{number}{4} (\bibinfo{year}{1966}),
  \bibinfo{pages}{922--935}.
\newblock


\bibitem[Haugh and Schneider(2012)]%
        {haugh2012politeness}
\bibfield{author}{\bibinfo{person}{Michael Haugh} {and}
  \bibinfo{person}{Klaus~P Schneider}.} \bibinfo{year}{2012}\natexlab{}.
\newblock \bibinfo{title}{Im/politeness across Englishes}.
\newblock , \bibinfo{numpages}{1017--1021}~pages.
\newblock


\bibitem[Held et~al\mbox{.}(2023)]%
        {held-etal-2023-tada}
\bibfield{author}{\bibinfo{person}{William Held}, \bibinfo{person}{Caleb
  Ziems}, {and} \bibinfo{person}{Diyi Yang}.} \bibinfo{year}{2023}\natexlab{}.
\newblock \showarticletitle{{TADA} : Task Agnostic Dialect Adapters for
  {E}nglish}. In \bibinfo{booktitle}{\emph{Findings of ACL}}.
  \bibinfo{pages}{813--824}.
\newblock
\urldef\tempurl%
\url{https://doi.org/10.18653/v1/2023.findings-acl.51}
\showDOI{\tempurl}


\bibitem[Hofmann et~al\mbox{.}(2024)]%
        {hofmann2024dialect}
\bibfield{author}{\bibinfo{person}{Valentin Hofmann},
  \bibinfo{person}{Pratyusha~Ria Kalluri}, \bibinfo{person}{Dan Jurafsky},
  {and} \bibinfo{person}{Sharese King}.} \bibinfo{year}{2024}\natexlab{}.
\newblock \showarticletitle{Dialect prejudice predicts AI decisions about
  people's character, employability, and criminality}.
\newblock \bibinfo{journal}{\emph{arXiv preprint arXiv:2403.00742}}
  (\bibinfo{year}{2024}).
\newblock


\bibitem[Honnet et~al\mbox{.}(2018)]%
        {honnet-etal-2018-machine}
\bibfield{author}{\bibinfo{person}{Pierre-Edouard Honnet},
  \bibinfo{person}{Andrei Popescu-Belis}, \bibinfo{person}{Claudiu Musat},
  {and} \bibinfo{person}{Michael Baeriswyl}.} \bibinfo{year}{2018}\natexlab{}.
\newblock \showarticletitle{Machine Translation of Low-Resource Spoken
  Dialects: Strategies for Normalizing {S}wiss {G}erman}. In
  \bibinfo{booktitle}{\emph{LREC}}. \bibinfo{publisher}{European Language
  Resources Association (ELRA)}, \bibinfo{address}{Miyazaki, Japan}.
\newblock
\urldef\tempurl%
\url{https://aclanthology.org/L18-1597}
\showURL{%
\tempurl}


\bibitem[Hou and Huang(2020)]%
        {hou2020classification}
\bibfield{author}{\bibinfo{person}{Renkui Hou} {and} \bibinfo{person}{Chu-Ren
  Huang}.} \bibinfo{year}{2020}\natexlab{}.
\newblock \showarticletitle{Classification of regional and genre varieties of
  Chinese: A correspondence analysis approach based on comparable balanced
  corpora}.
\newblock \bibinfo{journal}{\emph{Natural Language Engineering}}
  \bibinfo{volume}{26}, \bibinfo{number}{6} (\bibinfo{year}{2020}),
  \bibinfo{pages}{613--640}.
\newblock


\bibitem[Hovy and Purschke(2018)]%
        {hovy-purschke-2018-capturing}
\bibfield{author}{\bibinfo{person}{Dirk Hovy} {and} \bibinfo{person}{Christoph
  Purschke}.} \bibinfo{year}{2018}\natexlab{}.
\newblock \showarticletitle{Capturing Regional Variation with Distributed Place
  Representations and Geographic Retrofitting}. In
  \bibinfo{booktitle}{\emph{EMNLP}}. \bibinfo{publisher}{ACL},
  \bibinfo{address}{Brussels, Belgium}, \bibinfo{pages}{4383--4394}.
\newblock
\urldef\tempurl%
\url{https://doi.org/10.18653/v1/D18-1469}
\showDOI{\tempurl}


\bibitem[Hovy and Yang(2021)]%
        {hovy2021importance}
\bibfield{author}{\bibinfo{person}{Dirk Hovy} {and} \bibinfo{person}{Diyi
  Yang}.} \bibinfo{year}{2021}\natexlab{}.
\newblock \showarticletitle{The importance of modeling social factors of
  language: Theory and practice}. In \bibinfo{booktitle}{\emph{NAACL-HLT}}.
  \bibinfo{pages}{588--602}.
\newblock


\bibitem[Husain et~al\mbox{.}(2022)]%
        {husain2022weak}
\bibfield{author}{\bibinfo{person}{Fatemah Husain}, \bibinfo{person}{Hana
  Al-Ostad}, {and} \bibinfo{person}{Halima Omar}.}
  \bibinfo{year}{2022}\natexlab{}.
\newblock \showarticletitle{A weak supervised transfer learning approach for
  sentiment analysis to the Kuwaiti dialect}. In \bibinfo{booktitle}{\emph{The
  Seventh Arabic Natural Language Processing Workshop (WANLP)}}.
  \bibinfo{pages}{161--173}.
\newblock


\bibitem[Igarashi and Miyagawa(2024)]%
        {igarashi-miyagawa-2024-enhancing}
\bibfield{author}{\bibinfo{person}{Ryo Igarashi} {and} \bibinfo{person}{So
  Miyagawa}.} \bibinfo{year}{2024}\natexlab{}.
\newblock \showarticletitle{Enhancing Neural Machine Translation for
  {A}inu-{J}apanese: A Comprehensive Study on the Impact of Domain and Dialect
  Integration}. In \bibinfo{booktitle}{\emph{4th International Conference on
  Natural Language Processing for Digital Humanities}}.
  \bibinfo{publisher}{Association for Computational Linguistics},
  \bibinfo{address}{Miami, USA}, \bibinfo{pages}{413--422}.
\newblock
\urldef\tempurl%
\url{https://aclanthology.org/2024.nlp4dh-1.40}
\showURL{%
\tempurl}


\bibitem[Inoue et~al\mbox{.}(2021)]%
        {inoue2021interplay}
\bibfield{author}{\bibinfo{person}{Go Inoue}, \bibinfo{person}{Bashar Alhafni},
  \bibinfo{person}{Nurpeiis Baimukan}, \bibinfo{person}{Houda Bouamor}, {and}
  \bibinfo{person}{Nizar Habash}.} \bibinfo{year}{2021}\natexlab{}.
\newblock \showarticletitle{The Interplay of Variant, Size, and Task Type in
  Arabic Pre-trained Language Models}. In \bibinfo{booktitle}{\emph{6th Arabic
  Natural Language Processing Workshop, WANLP 2021}}. Association for
  Computational Linguistics (ACL), \bibinfo{pages}{92--104}.
\newblock


\bibitem[Inoue et~al\mbox{.}(2022)]%
        {inoue-etal-2022-morphosyntactic}
\bibfield{author}{\bibinfo{person}{Go Inoue}, \bibinfo{person}{Salam Khalifa},
  {and} \bibinfo{person}{Nizar Habash}.} \bibinfo{year}{2022}\natexlab{}.
\newblock \showarticletitle{Morphosyntactic Tagging with Pre-trained Language
  Models for {A}rabic and its Dialects}. In \bibinfo{booktitle}{\emph{Findings
  of ACL}}. \bibinfo{pages}{1708--1719}.
\newblock
\urldef\tempurl%
\url{https://doi.org/10.18653/v1/2022.findings-acl.135}
\showDOI{\tempurl}


\bibitem[Jarrar et~al\mbox{.}(2017)]%
        {jarrar2017curras}
\bibfield{author}{\bibinfo{person}{Mustafa Jarrar}, \bibinfo{person}{Nizar
  Habash}, \bibinfo{person}{Faeq Alrimawi}, \bibinfo{person}{Diyam Akra}, {and}
  \bibinfo{person}{Nasser Zalmout}.} \bibinfo{year}{2017}\natexlab{}.
\newblock \showarticletitle{Curras: an annotated corpus for the Palestinian
  Arabic dialect}.
\newblock \bibinfo{journal}{\emph{Language Resources and Evaluation}}
  \bibinfo{volume}{51} (\bibinfo{year}{2017}), \bibinfo{pages}{745--775}.
\newblock


\bibitem[Jauhiainen et~al\mbox{.}(2019)]%
        {jauhiainen2019automatic}
\bibfield{author}{\bibinfo{person}{Tommi Jauhiainen}, \bibinfo{person}{Marco
  Lui}, \bibinfo{person}{Marcos Zampieri}, \bibinfo{person}{Timothy Baldwin},
  {and} \bibinfo{person}{Krister Lind{\'e}n}.} \bibinfo{year}{2019}\natexlab{}.
\newblock \showarticletitle{Automatic language identification in texts: A
  Survey}.
\newblock \bibinfo{journal}{\emph{Journal of Artificial Intelligence Research}}
   \bibinfo{volume}{65} (\bibinfo{year}{2019}), \bibinfo{pages}{675--782}.
\newblock


\bibitem[Jeblee et~al\mbox{.}(2014)]%
        {jeblee-etal-2014-domain}
\bibfield{author}{\bibinfo{person}{Serena Jeblee}, \bibinfo{person}{Weston
  Feely}, \bibinfo{person}{Houda Bouamor}, \bibinfo{person}{Alon Lavie},
  \bibinfo{person}{Nizar Habash}, {and} \bibinfo{person}{Kemal Oflazer}.}
  \bibinfo{year}{2014}\natexlab{}.
\newblock \showarticletitle{Domain and Dialect Adaptation for Machine
  Translation into {E}gyptian {A}rabic}. In \bibinfo{booktitle}{\emph{Workshop
  on {A}rabic Natural Language Processing}}. \bibinfo{publisher}{Association
  for Computational Linguistics}, \bibinfo{address}{Doha, Qatar},
  \bibinfo{pages}{196--206}.
\newblock
\urldef\tempurl%
\url{https://doi.org/10.3115/v1/W14-3627}
\showDOI{\tempurl}


\bibitem[Jenkins(2009)]%
        {jenkins2009english}
\bibfield{author}{\bibinfo{person}{Jennifer Jenkins}.}
  \bibinfo{year}{2009}\natexlab{}.
\newblock \showarticletitle{English as a lingua franca: Interpretations and
  attitudes}.
\newblock \bibinfo{journal}{\emph{World Englishes}} \bibinfo{volume}{28},
  \bibinfo{number}{2} (\bibinfo{year}{2009}), \bibinfo{pages}{200--207}.
\newblock


\bibitem[Johnson and Verdicchio(2017)]%
        {johnson2017reframing}
\bibfield{author}{\bibinfo{person}{Deborah~G Johnson} {and}
  \bibinfo{person}{Mario Verdicchio}.} \bibinfo{year}{2017}\natexlab{}.
\newblock \showarticletitle{Reframing AI discourse}.
\newblock \bibinfo{journal}{\emph{Minds and Machines}}  \bibinfo{volume}{27}
  (\bibinfo{year}{2017}), \bibinfo{pages}{575--590}.
\newblock


\bibitem[J{\o}rgensen et~al\mbox{.}(2015)]%
        {jorgensen-etal-2015-challenges}
\bibfield{author}{\bibinfo{person}{Anna J{\o}rgensen}, \bibinfo{person}{Dirk
  Hovy}, {and} \bibinfo{person}{Anders S{\o}gaard}.}
  \bibinfo{year}{2015}\natexlab{}.
\newblock \showarticletitle{Challenges of studying and processing dialects in
  social media}. In \bibinfo{booktitle}{\emph{Workshop on Noisy User-generated
  Text}}. \bibinfo{pages}{9--18}.
\newblock
\urldef\tempurl%
\url{https://doi.org/10.18653/v1/W15-4302}
\showDOI{\tempurl}


\bibitem[Joukhadar et~al\mbox{.}(2019)]%
        {joukhadar2019arabic}
\bibfield{author}{\bibinfo{person}{Alaa Joukhadar}, \bibinfo{person}{Huda
  Saghergy}, \bibinfo{person}{Leen Kweider}, {and} \bibinfo{person}{Nada
  Ghneim}.} \bibinfo{year}{2019}\natexlab{}.
\newblock \showarticletitle{Arabic dialogue act recognition for textual chatbot
  systems}. In \bibinfo{booktitle}{\emph{First International Workshop on NLP
  Solutions for Under Resourced Languages}}. \bibinfo{pages}{43--49}.
\newblock


\bibitem[Jurgens et~al\mbox{.}(2017)]%
        {jurgens-etal-2017-incorporating}
\bibfield{author}{\bibinfo{person}{David Jurgens}, \bibinfo{person}{Yulia
  Tsvetkov}, {and} \bibinfo{person}{Dan Jurafsky}.}
  \bibinfo{year}{2017}\natexlab{}.
\newblock \showarticletitle{Incorporating Dialectal Variability for Socially
  Equitable Language Identification}. In \bibinfo{booktitle}{\emph{ACL}}.
  \bibinfo{address}{Vancouver, Canada}, \bibinfo{pages}{51--57}.
\newblock
\urldef\tempurl%
\url{https://doi.org/10.18653/v1/P17-2009}
\showDOI{\tempurl}


\bibitem[Kachru(1992)]%
        {kachru1992other}
\bibfield{author}{\bibinfo{person}{Braj~B Kachru}.}
  \bibinfo{year}{1992}\natexlab{}.
\newblock \showarticletitle{The other tongue: English across cultures}.
\newblock \bibinfo{journal}{\emph{Urbana}} (\bibinfo{year}{1992}).
\newblock


\bibitem[Kanjirangat et~al\mbox{.}(2022)]%
        {kanjirangat2022early}
\bibfield{author}{\bibinfo{person}{Vani Kanjirangat}, \bibinfo{person}{Tanja
  Samardzic}, \bibinfo{person}{Fabio Rinaldi}, {and} \bibinfo{person}{Ljiljana
  Dolamic}.} \bibinfo{year}{2022}\natexlab{}.
\newblock \showarticletitle{Early Guessing for Dialect Identification}. In
  \bibinfo{booktitle}{\emph{Findings of EMNLP}}. \bibinfo{pages}{6417--6426}.
\newblock


\bibitem[Kantharuban et~al\mbox{.}(2023)]%
        {kantharuban-etal-2023-quantifying}
\bibfield{author}{\bibinfo{person}{Anjali Kantharuban}, \bibinfo{person}{Ivan
  Vuli{\'c}}, {and} \bibinfo{person}{Anna Korhonen}.}
  \bibinfo{year}{2023}\natexlab{}.
\newblock \showarticletitle{Quantifying the Dialect Gap and its Correlates
  Across Languages}. In \bibinfo{booktitle}{\emph{Findings of EMNLP}}.
  \bibinfo{pages}{7226--7245}.
\newblock
\urldef\tempurl%
\url{https://aclanthology.org/2023.findings-emnlp.481}
\showURL{%
\tempurl}


\bibitem[Kaseb and Farouk(2022)]%
        {kaseb2022saids}
\bibfield{author}{\bibinfo{person}{Abdelrahman Kaseb} {and}
  \bibinfo{person}{Mona Farouk}.} \bibinfo{year}{2022}\natexlab{}.
\newblock \showarticletitle{SAIDS: A Novel Approach for Sentiment Analysis
  Informed of Dialect and Sarcasm}.
\newblock \bibinfo{journal}{\emph{WANLP 2022}} (\bibinfo{year}{2022}),
  \bibinfo{pages}{22}.
\newblock


\bibitem[K{\aa}sen et~al\mbox{.}(2022)]%
        {kasen-etal-2022-norwegian}
\bibfield{author}{\bibinfo{person}{Andre K{\aa}sen}, \bibinfo{person}{Kristin
  Hagen}, \bibinfo{person}{Anders N{\o}klestad}, \bibinfo{person}{Joel
  Priestly}, \bibinfo{person}{Per~Erik Solberg}, {and} \bibinfo{person}{Dag
  Trygve~Truslew Haug}.} \bibinfo{year}{2022}\natexlab{}.
\newblock \showarticletitle{The {N}orwegian Dialect Corpus Treebank}. In
  \bibinfo{booktitle}{\emph{LREC}}. \bibinfo{pages}{4827--4832}.
\newblock
\urldef\tempurl%
\url{https://aclanthology.org/2022.lrec-1.516}
\showURL{%
\tempurl}


\bibitem[Keleg et~al\mbox{.}(2023)]%
        {keleg-etal-2023-aldi}
\bibfield{author}{\bibinfo{person}{Amr Keleg}, \bibinfo{person}{Sharon
  Goldwater}, {and} \bibinfo{person}{Walid Magdy}.}
  \bibinfo{year}{2023}\natexlab{}.
\newblock \showarticletitle{{ALD}i: Quantifying the {A}rabic Level of
  Dialectness of Text}. In \bibinfo{booktitle}{\emph{EMNLP}}.
  \bibinfo{pages}{10597--10611}.
\newblock
\urldef\tempurl%
\url{https://aclanthology.org/2023.emnlp-main.655}
\showURL{%
\tempurl}


\bibitem[Keswani and Celis(2021)]%
        {keswani2021dialect}
\bibfield{author}{\bibinfo{person}{Vijay Keswani} {and}
  \bibinfo{person}{L~Elisa Celis}.} \bibinfo{year}{2021}\natexlab{}.
\newblock \showarticletitle{Dialect diversity in text summarization on
  twitter}. In \bibinfo{booktitle}{\emph{The Web Conference 2021}}.
  \bibinfo{pages}{3802--3814}.
\newblock


\bibitem[Khalifa et~al\mbox{.}(2016)]%
        {khalifa2016large}
\bibfield{author}{\bibinfo{person}{Salam Khalifa}, \bibinfo{person}{Nizar
  Habash}, \bibinfo{person}{Dana Abdulrahim}, {and} \bibinfo{person}{Sara
  Hassan}.} \bibinfo{year}{2016}\natexlab{}.
\newblock \showarticletitle{A large scale corpus of Gulf Arabic}. In
  \bibinfo{booktitle}{\emph{LREC}}. European Language Resources Association
  (ELRA), \bibinfo{pages}{4282--4289}.
\newblock


\bibitem[Khalifa et~al\mbox{.}(2020)]%
        {khalifa2020morphological}
\bibfield{author}{\bibinfo{person}{Salam Khalifa}, \bibinfo{person}{Nasser
  Zalmout}, {and} \bibinfo{person}{Nizar Habash}.}
  \bibinfo{year}{2020}\natexlab{}.
\newblock \showarticletitle{Morphological analysis and disambiguation for Gulf
  Arabic: The interplay between resources and methods}. In
  \bibinfo{booktitle}{\emph{LREC}}. \bibinfo{pages}{3895--3904}.
\newblock


\bibitem[Kroch(1986)]%
        {kroch1986toward}
\bibfield{author}{\bibinfo{person}{Anthony~S Kroch}.}
  \bibinfo{year}{1986}\natexlab{}.
\newblock \showarticletitle{Toward a theory of social dialect variation}.
\newblock In \bibinfo{booktitle}{\emph{Dialect and Language Variation}}.
  \bibinfo{publisher}{Elsevier}, \bibinfo{pages}{344--366}.
\newblock


\bibitem[Kumar et~al\mbox{.}(2021)]%
        {kumar-etal-2021-machine}
\bibfield{author}{\bibinfo{person}{Sachin Kumar}, \bibinfo{person}{Antonios
  Anastasopoulos}, \bibinfo{person}{Shuly Wintner}, {and}
  \bibinfo{person}{Yulia Tsvetkov}.} \bibinfo{year}{2021}\natexlab{}.
\newblock \showarticletitle{Machine Translation into Low-resource Language
  Varieties}. In \bibinfo{booktitle}{\emph{ACL-IJCNLP}}.
  \bibinfo{publisher}{Association for Computational Linguistics},
  \bibinfo{address}{Online}, \bibinfo{pages}{110--121}.
\newblock
\urldef\tempurl%
\url{https://doi.org/10.18653/v1/2021.acl-short.16}
\showDOI{\tempurl}


\bibitem[Kuparinen(2023)]%
        {kuparinen-2023-murreviikko}
\bibfield{author}{\bibinfo{person}{Olli Kuparinen}.}
  \bibinfo{year}{2023}\natexlab{}.
\newblock \showarticletitle{Murreviikko - A Dialectologically Annotated and
  Normalized Dataset of {F}innish Tweets}. In
  \bibinfo{booktitle}{\emph{VarDial}}. \bibinfo{publisher}{Association for
  Computational Linguistics}, \bibinfo{address}{Dubrovnik, Croatia},
  \bibinfo{pages}{31--39}.
\newblock
\urldef\tempurl%
\url{https://doi.org/10.18653/v1/2023.vardial-1.3}
\showDOI{\tempurl}


\bibitem[Kuparinen et~al\mbox{.}(2023)]%
        {kuparinen-etal-2023-dialect}
\bibfield{author}{\bibinfo{person}{Olli Kuparinen}, \bibinfo{person}{Aleksandra
  Mileti{\'c}}, {and} \bibinfo{person}{Yves Scherrer}.}
  \bibinfo{year}{2023}\natexlab{}.
\newblock \showarticletitle{Dialect-to-Standard Normalization: A Large-Scale
  Multilingual Evaluation}. In \bibinfo{booktitle}{\emph{Findings of EMNLP}}.
  \bibinfo{publisher}{Association for Computational Linguistics},
  \bibinfo{pages}{13814--13828}.
\newblock
\urldef\tempurl%
\url{https://aclanthology.org/2023.findings-emnlp.923}
\showURL{%
\tempurl}


\bibitem[Lameli and Sch{\"o}nberg(2023)]%
        {lameli2023measure}
\bibfield{author}{\bibinfo{person}{Alfred Lameli} {and}
  \bibinfo{person}{Andreas Sch{\"o}nberg}.} \bibinfo{year}{2023}\natexlab{}.
\newblock \showarticletitle{A Measure for Linguistic Coherence in Spatial
  Language Variation}. In \bibinfo{booktitle}{\emph{VarDIAL}}.
  \bibinfo{pages}{133--141}.
\newblock


\bibitem[Le and Luu(2023)]%
        {le-luu-2023-parallel}
\bibfield{author}{\bibinfo{person}{Thang Le} {and} \bibinfo{person}{Anh Luu}.}
  \bibinfo{year}{2023}\natexlab{}.
\newblock \showarticletitle{A Parallel Corpus for {V}ietnamese Central-Northern
  Dialect Text Transfer}. In \bibinfo{booktitle}{\emph{Findings of EMNLP}}.
  \bibinfo{pages}{13839--13855}.
\newblock
\urldef\tempurl%
\url{https://aclanthology.org/2023.findings-emnlp.925}
\showURL{%
\tempurl}


\bibitem[Lent et~al\mbox{.}(2024)]%
        {lent2023creoleval}
\bibfield{author}{\bibinfo{person}{Heather Lent}, \bibinfo{person}{Kushal
  Tatariya}, \bibinfo{person}{Raj Dabre}, \bibinfo{person}{Yiyi Chen},
  \bibinfo{person}{Marcell Fekete}, \bibinfo{person}{Esther Ploeger},
  \bibinfo{person}{Li Zhou}, \bibinfo{person}{Ruth-Ann Armstrong},
  \bibinfo{person}{Abee Eijansantos}, \bibinfo{person}{Catriona Malau},
  \bibinfo{person}{Hans~Erik Heje}, \bibinfo{person}{Ernests Lavrinovics},
  \bibinfo{person}{Diptesh Kanojia}, \bibinfo{person}{Paul Belony},
  \bibinfo{person}{Marcel Bollmann}, \bibinfo{person}{Lo{\"\i}c Grobol},
  \bibinfo{person}{Miryam~de Lhoneux}, \bibinfo{person}{Daniel Hershcovich},
  \bibinfo{person}{Michel DeGraff}, \bibinfo{person}{Anders S{\o}gaard}, {and}
  \bibinfo{person}{Johannes Bjerva}.} \bibinfo{year}{2024}\natexlab{}.
\newblock \showarticletitle{{C}reole{V}al: Multilingual Multitask Benchmarks
  for Creoles}.
\newblock \bibinfo{journal}{\emph{Transactions of the Association for
  Computational Linguistics}}  \bibinfo{volume}{12} (\bibinfo{year}{2024}),
  \bibinfo{pages}{950--978}.
\newblock
\urldef\tempurl%
\url{https://doi.org/10.1162/tacl_a_00682}
\showDOI{\tempurl}


\bibitem[Li and Mao(2015)]%
        {li2015hedonic}
\bibfield{author}{\bibinfo{person}{Manning Li} {and} \bibinfo{person}{Jiye
  Mao}.} \bibinfo{year}{2015}\natexlab{}.
\newblock \showarticletitle{Hedonic or utilitarian? Exploring the impact of
  communication style alignment on user's perception of virtual health advisory
  services}.
\newblock \bibinfo{journal}{\emph{International Journal of Information
  Management}} \bibinfo{volume}{35}, \bibinfo{number}{2}
  (\bibinfo{year}{2015}), \bibinfo{pages}{229--243}.
\newblock


\bibitem[Liu et~al\mbox{.}(2023)]%
        {liu-etal-2023-dada}
\bibfield{author}{\bibinfo{person}{Yanchen Liu}, \bibinfo{person}{William
  Held}, {and} \bibinfo{person}{Diyi Yang}.} \bibinfo{year}{2023}\natexlab{}.
\newblock \showarticletitle{{DADA}: Dialect Adaptation via Dynamic Aggregation
  of Linguistic Rules}. In \bibinfo{booktitle}{\emph{EMNLP}}.
  \bibinfo{address}{Singapore}, \bibinfo{pages}{13776--13793}.
\newblock


\bibitem[Liu et~al\mbox{.}(2022)]%
        {liu-etal-2022-singlish}
\bibfield{author}{\bibinfo{person}{Zhengyuan Liu}, \bibinfo{person}{Shikang
  Ni}, \bibinfo{person}{Ai~Ti Aw}, {and} \bibinfo{person}{Nancy~F. Chen}.}
  \bibinfo{year}{2022}\natexlab{}.
\newblock \showarticletitle{{S}inglish Message Paraphrasing: A Joint Task of
  Creole Translation and Text Normalization}. In
  \bibinfo{booktitle}{\emph{COLING}}. \bibinfo{pages}{3924--3936}.
\newblock
\urldef\tempurl%
\url{https://aclanthology.org/2022.coling-1.345}
\showURL{%
\tempurl}


\bibitem[Lu et~al\mbox{.}(2022)]%
        {lu-etal-2022-exploring}
\bibfield{author}{\bibinfo{person}{Sin-En Lu}, \bibinfo{person}{Bo-Han Lu},
  \bibinfo{person}{Chao-Yi Lu}, {and} \bibinfo{person}{Richard Tzong-Han
  Tsai}.} \bibinfo{year}{2022}\natexlab{}.
\newblock \showarticletitle{Exploring Methods for Building Dialects-{M}andarin
  Code-Mixing Corpora: A Case Study in {T}aiwanese Hokkien}. In
  \bibinfo{booktitle}{\emph{Findings of EMNLP}}. \bibinfo{pages}{6287--6305}.
\newblock
\urldef\tempurl%
\url{https://doi.org/10.18653/v1/2022.findings-emnlp.469}
\showDOI{\tempurl}


\bibitem[Lui and Baldwin(2012)]%
        {lui-baldwin-2012-langid}
\bibfield{author}{\bibinfo{person}{Marco Lui} {and} \bibinfo{person}{Timothy
  Baldwin}.} \bibinfo{year}{2012}\natexlab{}.
\newblock \showarticletitle{langid.py: An Off-the-shelf Language Identification
  Tool}. In \bibinfo{booktitle}{\emph{ACL System Demonstrations}}.
  \bibinfo{address}{Jeju Island, Korea}, \bibinfo{pages}{25--30}.
\newblock


\bibitem[Lui and Cook(2013)]%
        {lui2013classifying}
\bibfield{author}{\bibinfo{person}{Marco Lui} {and} \bibinfo{person}{Paul
  Cook}.} \bibinfo{year}{2013}\natexlab{}.
\newblock \showarticletitle{Classifying English documents by national dialect}.
  In \bibinfo{booktitle}{\emph{Australasian Language Technology Association
  Workshop}}. \bibinfo{pages}{5--15}.
\newblock


\bibitem[Maamouri et~al\mbox{.}(2014)]%
        {maamouri2014developing}
\bibfield{author}{\bibinfo{person}{Mohamed Maamouri}, \bibinfo{person}{Ann
  Bies}, \bibinfo{person}{Seth Kulick}, \bibinfo{person}{Michael Ciul},
  \bibinfo{person}{Nizar Habash}, {and} \bibinfo{person}{Ramy Eskander}.}
  \bibinfo{year}{2014}\natexlab{}.
\newblock \showarticletitle{Developing an Egyptian Arabic Treebank: Impact of
  Dialectal Morphology on Annotation and Tool Development.}. In
  \bibinfo{booktitle}{\emph{LREC}}. \bibinfo{pages}{2348--2354}.
\newblock


\bibitem[Malmasi et~al\mbox{.}(2016)]%
        {malmasi2016discriminating}
\bibfield{author}{\bibinfo{person}{Shervin Malmasi}, \bibinfo{person}{Marcos
  Zampieri}, \bibinfo{person}{Nikola Ljube{\v{s}}i{\'c}},
  \bibinfo{person}{Preslav Nakov}, \bibinfo{person}{Ahmed Ali}, {and}
  \bibinfo{person}{J{\"o}rg Tiedemann}.} \bibinfo{year}{2016}\natexlab{}.
\newblock \showarticletitle{Discriminating between similar languages and arabic
  dialect identification: A report on the third dsl shared task}. In
  \bibinfo{booktitle}{\emph{VarDial}}. \bibinfo{pages}{1--14}.
\newblock


\bibitem[Maurya et~al\mbox{.}(2023)]%
        {maurya2023utilizing}
\bibfield{author}{\bibinfo{person}{Kaushal~Kumar Maurya},
  \bibinfo{person}{Rahul Kejriwal}, \bibinfo{person}{Maunendra~Sankar
  Desarkar}, {and} \bibinfo{person}{Anoop Kunchukuttan}.}
  \bibinfo{year}{2023}\natexlab{}.
\newblock \showarticletitle{Utilizing Lexical Similarity to Enable Zero-Shot
  Machine Translation for Extremely Low-resource Languages}.
\newblock \bibinfo{journal}{\emph{arXiv preprint arXiv:2305.05214}}
  (\bibinfo{year}{2023}).
\newblock


\bibitem[Mdhaffar et~al\mbox{.}(2017)]%
        {mdhaffar2017sentiment}
\bibfield{author}{\bibinfo{person}{Salima Mdhaffar}, \bibinfo{person}{Fethi
  Bougares}, \bibinfo{person}{Yannick Esteve}, {and} \bibinfo{person}{Lamia
  Hadrich-Belguith}.} \bibinfo{year}{2017}\natexlab{}.
\newblock \showarticletitle{Sentiment analysis of tunisian dialects: Linguistic
  ressources and experiments}. In \bibinfo{booktitle}{\emph{Arabic Natural
  Language Processing Workshop}}. \bibinfo{pages}{55--61}.
\newblock


\bibitem[Meftouh et~al\mbox{.}(2015)]%
        {meftouh-etal-2015-machine}
\bibfield{author}{\bibinfo{person}{Karima Meftouh}, \bibinfo{person}{Salima
  Harrat}, \bibinfo{person}{Salma Jamoussi}, \bibinfo{person}{Mourad Abbas},
  {and} \bibinfo{person}{Kamel Smaili}.} \bibinfo{year}{2015}\natexlab{}.
\newblock \showarticletitle{Machine Translation Experiments on {PADIC}: A
  Parallel {A}rabic {DI}alect Corpus}. In \bibinfo{booktitle}{\emph{PACLIC}}.
  \bibinfo{address}{Shanghai, China}, \bibinfo{pages}{26--34}.
\newblock
\urldef\tempurl%
\url{https://aclanthology.org/Y15-1004}
\showURL{%
\tempurl}


\bibitem[Merrison et~al\mbox{.}(2012)]%
        {merrison2012getting}
\bibfield{author}{\bibinfo{person}{Andrew~John Merrison},
  \bibinfo{person}{Jack~J Wilson}, \bibinfo{person}{Bethan~L Davies}, {and}
  \bibinfo{person}{Michael Haugh}.} \bibinfo{year}{2012}\natexlab{}.
\newblock \showarticletitle{Getting stuff done: Comparing e-mail requests from
  students in higher education in Britain and Australia}.
\newblock \bibinfo{journal}{\emph{Journal of pragmatics}} \bibinfo{volume}{44},
  \bibinfo{number}{9} (\bibinfo{year}{2012}), \bibinfo{pages}{1077--1098}.
\newblock


\bibitem[Messner and Lippincott(2024)]%
        {messner-lippincott-2024-examining}
\bibfield{author}{\bibinfo{person}{Craig Messner} {and} \bibinfo{person}{Thomas
  Lippincott}.} \bibinfo{year}{2024}\natexlab{}.
\newblock \showarticletitle{Examining Language Modeling Assumptions Using an
  Annotated Literary Dialect Corpus}. In \bibinfo{booktitle}{\emph{4th
  International Conference on Natural Language Processing for Digital
  Humanities}}. \bibinfo{publisher}{Association for Computational Linguistics},
  \bibinfo{address}{Miami, USA}, \bibinfo{pages}{325--330}.
\newblock
\urldef\tempurl%
\url{https://aclanthology.org/2024.nlp4dh-1.32}
\showURL{%
\tempurl}


\bibitem[Meyer(2014)]%
        {meyer2014culture}
\bibfield{author}{\bibinfo{person}{Erin Meyer}.}
  \bibinfo{year}{2014}\natexlab{}.
\newblock \bibinfo{booktitle}{\emph{The culture map: Breaking through the
  invisible boundaries of global business}}.
\newblock \bibinfo{publisher}{Public Affairs}.
\newblock


\bibitem[Moghimifar et~al\mbox{.}(2023)]%
        {moghimifar-etal-2023-normmark}
\bibfield{author}{\bibinfo{person}{Farhad Moghimifar}, \bibinfo{person}{Shilin
  Qu}, \bibinfo{person}{Tongtong Wu}, \bibinfo{person}{Yuan-Fang Li}, {and}
  \bibinfo{person}{Gholamreza Haffari}.} \bibinfo{year}{2023}\natexlab{}.
\newblock \showarticletitle{{N}orm{M}ark: A Weakly Supervised {M}arkov Model
  for Socio-cultural Norm Discovery}. In \bibinfo{booktitle}{\emph{Findings of
  ACL}}. \bibinfo{address}{Toronto, Canada}, \bibinfo{pages}{5081--5089}.
\newblock


\bibitem[Moore(1999)]%
        {moore1999vocabulary}
\bibfield{author}{\bibinfo{person}{Bruce Moore}.}
  \bibinfo{year}{1999}\natexlab{}.
\newblock \showarticletitle{The Vocabulary of Australian English}.
\newblock \bibinfo{journal}{\emph{Australian National Dictionary Centre,
  Australian National University. URL: http://andc. anu. edu.
  au/sites/default/files/vocab\_aussie\_eng. pdf}} (\bibinfo{year}{1999}).
\newblock


\bibitem[Morin and Coats(2023)]%
        {morin2023double}
\bibfield{author}{\bibinfo{person}{Cameron Morin} {and} \bibinfo{person}{Steven
  Coats}.} \bibinfo{year}{2023}\natexlab{}.
\newblock \showarticletitle{Double modals in Australian and New Zealand
  English}.
\newblock \bibinfo{journal}{\emph{World Englishes}} (\bibinfo{year}{2023}).
\newblock


\bibitem[Mozafari et~al\mbox{.}(2020)]%
        {mozafari2020hate}
\bibfield{author}{\bibinfo{person}{Marzieh Mozafari}, \bibinfo{person}{Reza
  Farahbakhsh}, {and} \bibinfo{person}{No{\"e}l Crespi}.}
  \bibinfo{year}{2020}\natexlab{}.
\newblock \showarticletitle{Hate speech detection and racial bias mitigation in
  social media based on BERT model}.
\newblock \bibinfo{journal}{\emph{PloS one}} \bibinfo{volume}{15},
  \bibinfo{number}{8} (\bibinfo{year}{2020}), \bibinfo{pages}{e0237861}.
\newblock


\bibitem[Mulki et~al\mbox{.}(2019)]%
        {mulki2019syntax}
\bibfield{author}{\bibinfo{person}{Hala Mulki}, \bibinfo{person}{Hatem Haddad},
  \bibinfo{person}{Mourad Gridach}, {and} \bibinfo{person}{Ismail
  Babao{\u{g}}lu}.} \bibinfo{year}{2019}\natexlab{}.
\newblock \showarticletitle{Syntax-ignorant N-gram embeddings for sentiment
  analysis of Arabic dialects}. In \bibinfo{booktitle}{\emph{Arabic Natural
  Language Processing Workshop}}. \bibinfo{pages}{30--39}.
\newblock


\bibitem[Nagata(2014)]%
        {nagata-2014-language}
\bibfield{author}{\bibinfo{person}{Ryo Nagata}.}
  \bibinfo{year}{2014}\natexlab{}.
\newblock \showarticletitle{Language Family Relationship Preserved in
  Non-native {E}nglish}. In \bibinfo{booktitle}{\emph{COLING}}.
  \bibinfo{publisher}{Dublin City University and Association for Computational
  Linguistics}, \bibinfo{pages}{1940--1949}.
\newblock
\urldef\tempurl%
\url{https://aclanthology.org/C14-1183}
\showURL{%
\tempurl}


\bibitem[Naveed et~al\mbox{.}(2023)]%
        {naveed2023comprehensive}
\bibfield{author}{\bibinfo{person}{Humza Naveed}, \bibinfo{person}{Asad~Ullah
  Khan}, \bibinfo{person}{Shi Qiu}, \bibinfo{person}{Muhammad Saqib},
  \bibinfo{person}{Saeed Anwar}, \bibinfo{person}{Muhammad Usman},
  \bibinfo{person}{Naveed Akhtar}, \bibinfo{person}{Nick Barnes}, {and}
  \bibinfo{person}{Ajmal Mian}.} \bibinfo{year}{2023}\natexlab{}.
\newblock \showarticletitle{A comprehensive overview of large language models}.
\newblock \bibinfo{journal}{\emph{arXiv preprint arXiv:2307.06435}}
  (\bibinfo{year}{2023}).
\newblock


\bibitem[Nerbonne and Heeringa(1997)]%
        {nerbonne1997measuring}
\bibfield{author}{\bibinfo{person}{John Nerbonne} {and}
  \bibinfo{person}{Wilbert Heeringa}.} \bibinfo{year}{1997}\natexlab{}.
\newblock \showarticletitle{Measuring dialect distance phonetically}. In
  \bibinfo{booktitle}{\emph{Computational phonology: third meeting of the acl
  special interest group in computational phonology}}.
\newblock


\bibitem[Nerbonne and Heeringa(2001)]%
        {nerbonne2001computational}
\bibfield{author}{\bibinfo{person}{John Nerbonne} {and}
  \bibinfo{person}{Wilbert Heeringa}.} \bibinfo{year}{2001}\natexlab{}.
\newblock \showarticletitle{Computational comparison and classification of
  dialects}.
\newblock  (\bibinfo{year}{2001}).
\newblock


\bibitem[Nerbonne and Heeringa(2002)]%
        {nerbonne2002computational}
\bibfield{author}{\bibinfo{person}{J Nerbonne} {and} \bibinfo{person}{Wilbert
  Heeringa}.} \bibinfo{year}{2002}\natexlab{}.
\newblock \showarticletitle{Computational Comparison and Classification of
  Dialects}.
\newblock \bibinfo{journal}{\emph{Dialectologia et Geolinguistica, Journal of
  the International Society for Dialectology and Geolinguistics}}
  \bibinfo{volume}{9} (\bibinfo{year}{2002}), \bibinfo{pages}{69--84}.
\newblock


\bibitem[Nigatu et~al\mbox{.}(2024)]%
        {nigatu-etal-2024-zenos}
\bibfield{author}{\bibinfo{person}{Hellina~Hailu Nigatu},
  \bibinfo{person}{Atnafu~Lambebo Tonja}, \bibinfo{person}{Benjamin Rosman},
  \bibinfo{person}{Thamar Solorio}, {and} \bibinfo{person}{Monojit Choudhury}.}
  \bibinfo{year}{2024}\natexlab{}.
\newblock \showarticletitle{The Zeno{'}s Paradox of {`}Low-Resource{'}
  Languages}. In \bibinfo{booktitle}{\emph{Proceedings of the 2024 Conference
  on Empirical Methods in Natural Language Processing}},
  \bibfield{editor}{\bibinfo{person}{Yaser Al-Onaizan}, \bibinfo{person}{Mohit
  Bansal}, {and} \bibinfo{person}{Yun-Nung Chen}} (Eds.).
  \bibinfo{publisher}{Association for Computational Linguistics},
  \bibinfo{address}{Miami, Florida, USA}, \bibinfo{pages}{17753--17774}.
\newblock
\urldef\tempurl%
\url{https://doi.org/10.18653/v1/2024.emnlp-main.983}
\showDOI{\tempurl}


\bibitem[Obeid et~al\mbox{.}(2018)]%
        {obeid-etal-2018-madar}
\bibfield{author}{\bibinfo{person}{Ossama Obeid}, \bibinfo{person}{Salam
  Khalifa}, \bibinfo{person}{Nizar Habash}, \bibinfo{person}{Houda Bouamor},
  \bibinfo{person}{Wajdi Zaghouani}, {and} \bibinfo{person}{Kemal Oflazer}.}
  \bibinfo{year}{2018}\natexlab{}.
\newblock \showarticletitle{{MADAR}i: A Web Interface for Joint {A}rabic
  Morphological Annotation and Spelling Correction}. In
  \bibinfo{booktitle}{\emph{LREC}}. \bibinfo{publisher}{European Language
  Resources Association (ELRA)}, \bibinfo{address}{Miyazaki, Japan}.
\newblock
\urldef\tempurl%
\url{https://aclanthology.org/L18-1415}
\showURL{%
\tempurl}


\bibitem[Obeid et~al\mbox{.}(2019)]%
        {obeid2019adida}
\bibfield{author}{\bibinfo{person}{Ossama Obeid}, \bibinfo{person}{Mohammad
  Salameh}, \bibinfo{person}{Houda Bouamor}, {and} \bibinfo{person}{Nizar
  Habash}.} \bibinfo{year}{2019}\natexlab{}.
\newblock \showarticletitle{ADIDA: Automatic dialect identification for
  Arabic}. In \bibinfo{booktitle}{\emph{NAACL (System demonstrations)}}.
  \bibinfo{pages}{6--11}.
\newblock


\bibitem[Obeid et~al\mbox{.}(2020)]%
        {obeid2020camel}
\bibfield{author}{\bibinfo{person}{Ossama Obeid}, \bibinfo{person}{Nasser
  Zalmout}, \bibinfo{person}{Salam Khalifa}, \bibinfo{person}{Dima Taji},
  \bibinfo{person}{Mai Oudah}, \bibinfo{person}{Bashar Alhafni},
  \bibinfo{person}{Go Inoue}, \bibinfo{person}{Fadhl Eryani},
  \bibinfo{person}{Alexander Erdmann}, {and} \bibinfo{person}{Nizar Habash}.}
  \bibinfo{year}{2020}\natexlab{}.
\newblock \showarticletitle{CAMeL tools: An open source python toolkit for
  Arabic natural language processing}. In \bibinfo{booktitle}{\emph{LREC}}.
  \bibinfo{pages}{7022--7032}.
\newblock


\bibitem[Okpala et~al\mbox{.}(2022)]%
        {okpala2022aaebert}
\bibfield{author}{\bibinfo{person}{Ebuka Okpala}, \bibinfo{person}{Long Cheng},
  \bibinfo{person}{Nicodemus Mbwambo}, {and} \bibinfo{person}{Feng Luo}.}
  \bibinfo{year}{2022}\natexlab{}.
\newblock \showarticletitle{AAEBERT: Debiasing BERT-based Hate Speech Detection
  Models via Adversarial Learning}. In \bibinfo{booktitle}{\emph{ICMLA}}. IEEE,
  \bibinfo{pages}{1606--1612}.
\newblock


\bibitem[Olabisi et~al\mbox{.}(2022)]%
        {olabisi2022analyzing}
\bibfield{author}{\bibinfo{person}{Olubusayo Olabisi}, \bibinfo{person}{Aaron
  Hudson}, \bibinfo{person}{Antonie Jetter}, {and} \bibinfo{person}{Ameeta
  Agrawal}.} \bibinfo{year}{2022}\natexlab{}.
\newblock \showarticletitle{Analyzing the Dialect Diversity in Multi-document
  Summaries}. In \bibinfo{booktitle}{\emph{COLING}}.
  \bibinfo{pages}{6208--6221}.
\newblock


\bibitem[Oussous et~al\mbox{.}(2020)]%
        {oussous2020asa}
\bibfield{author}{\bibinfo{person}{Ahmed Oussous},
  \bibinfo{person}{Fatima-Zahra Benjelloun}, \bibinfo{person}{Ayoub~Ait
  Lahcen}, {and} \bibinfo{person}{Samir Belfkih}.}
  \bibinfo{year}{2020}\natexlab{}.
\newblock \showarticletitle{ASA: A framework for Arabic sentiment analysis}.
\newblock \bibinfo{journal}{\emph{Journal of Information Science}}
  \bibinfo{volume}{46}, \bibinfo{number}{4} (\bibinfo{year}{2020}),
  \bibinfo{pages}{544--559}.
\newblock


\bibitem[Paul et~al\mbox{.}(2011)]%
        {paul2011dialect}
\bibfield{author}{\bibinfo{person}{Michael Paul}, \bibinfo{person}{Andrew
  Finch}, \bibinfo{person}{Paul Dixon}, {and} \bibinfo{person}{Eiichiro
  Sumita}.} \bibinfo{year}{2011}\natexlab{}.
\newblock \showarticletitle{Dialect translation: integrating Bayesian
  co-segmentation models with pivot-based SMT}. In
  \bibinfo{booktitle}{\emph{Workshop on Algorithms and Resources for Modelling
  of Dialects and Language Varieties}}.
\newblock


\bibitem[Pl{\"u}ss et~al\mbox{.}(2023)]%
        {pluss-etal-2023-stt4sg}
\bibfield{author}{\bibinfo{person}{Michel Pl{\"u}ss}, \bibinfo{person}{Jan
  Deriu}, \bibinfo{person}{Yanick Schraner}, \bibinfo{person}{Claudio
  Paonessa}, \bibinfo{person}{Julia Hartmann}, \bibinfo{person}{Larissa
  Schmidt}, \bibinfo{person}{Christian Scheller}, \bibinfo{person}{Manuela
  H{\"u}rlimann}, \bibinfo{person}{Tanja Samard{\v{z}}i{\'c}},
  \bibinfo{person}{Manfred Vogel}, {and} \bibinfo{person}{Mark Cieliebak}.}
  \bibinfo{year}{2023}\natexlab{}.
\newblock \showarticletitle{{STT}4{SG}-350: A Speech Corpus for All {S}wiss
  {G}erman Dialect Regions}. In \bibinfo{booktitle}{\emph{ACL (Short Papers)}}.
  \bibinfo{address}{Toronto, Canada}, \bibinfo{pages}{1763--1772}.
\newblock
\urldef\tempurl%
\url{https://aclanthology.org/2023.acl-short.150}
\showURL{%
\tempurl}


\bibitem[Qian et~al\mbox{.}(2024)]%
        {qian-etal-2024-large}
\bibfield{author}{\bibinfo{person}{Shenbin Qian}, \bibinfo{person}{Archchana
  Sindhujan}, \bibinfo{person}{Minnie Kabra}, \bibinfo{person}{Diptesh
  Kanojia}, \bibinfo{person}{Constantin Orasan}, \bibinfo{person}{Tharindu
  Ranasinghe}, {and} \bibinfo{person}{Fred Blain}.}
  \bibinfo{year}{2024}\natexlab{}.
\newblock \showarticletitle{What do Large Language Models Need for Machine
  Translation Evaluation?}. In \bibinfo{booktitle}{\emph{Proceedings of the
  2024 Conference on EMNLP}}. \bibinfo{publisher}{Association for Computational
  Linguistics}, \bibinfo{address}{Miami, Florida, USA},
  \bibinfo{pages}{3660--3674}.
\newblock


\bibitem[Rajai and Ennasser(2022)]%
        {rajai2022dealing}
\bibfield{author}{\bibinfo{person}{Al-Khanji Rajai} {and}
  \bibinfo{person}{Narjes Ennasser}.} \bibinfo{year}{2022}\natexlab{}.
\newblock \showarticletitle{Dealing with dialects in literary translation:
  Problems and strategies}.
\newblock \bibinfo{journal}{\emph{Jordan Journal of Modern Languages and
  Literatures}} \bibinfo{volume}{14}, \bibinfo{number}{1}
  (\bibinfo{year}{2022}), \bibinfo{pages}{145--163}.
\newblock


\bibitem[Ramponi(2024)]%
        {ramponi2024language}
\bibfield{author}{\bibinfo{person}{Alan Ramponi}.}
  \bibinfo{year}{2024}\natexlab{}.
\newblock \showarticletitle{Language Varieties of Italy: Technology Challenges
  and Opportunities}.
\newblock \bibinfo{journal}{\emph{Transactions of the Association for
  Computational Linguistics}}  \bibinfo{volume}{12} (\bibinfo{year}{2024}),
  \bibinfo{pages}{19--38}.
\newblock


\bibitem[Ramponi and Casula(2023a)]%
        {ramponi2023diatopit}
\bibfield{author}{\bibinfo{person}{Alan Ramponi} {and} \bibinfo{person}{Camilla
  Casula}.} \bibinfo{year}{2023}\natexlab{a}.
\newblock \showarticletitle{DIATOPIT: A Corpus of Social Media Posts for the
  Study of Diatopic Language Variation in Italy}. In
  \bibinfo{booktitle}{\emph{VarDial}}. \bibinfo{pages}{187--199}.
\newblock


\bibitem[Ramponi and Casula(2023b)]%
        {ramponi-casula-2023-geolingit}
\bibfield{author}{\bibinfo{person}{Alan Ramponi} {and} \bibinfo{person}{Camilla
  Casula}.} \bibinfo{year}{2023}\natexlab{b}.
\newblock \showarticletitle{{G}eo{L}ing{I}t at {EVALITA} 2023: Overview of the
  Geolocation of Linguistic Variation in {I}taly Task}. In
  \bibinfo{booktitle}{\emph{Evaluation Campaign of Natural Language Processing
  and Speech Tools for Italian}}. \bibinfo{publisher}{CEUR.org},
  \bibinfo{address}{Parma, Italy}.
\newblock


\bibitem[Riabi et~al\mbox{.}(2023)]%
        {riabi-etal-2023-enriching}
\bibfield{author}{\bibinfo{person}{Arij Riabi}, \bibinfo{person}{Menel
  Mahamdi}, {and} \bibinfo{person}{Djam{\'e} Seddah}.}
  \bibinfo{year}{2023}\natexlab{}.
\newblock \showarticletitle{Enriching the {NA}rabizi Treebank: A Multifaceted
  Approach to Supporting an Under-Resourced Language}. In
  \bibinfo{booktitle}{\emph{17th Linguistic Annotation Workshop (LAW-XVII)}}.
  \bibinfo{publisher}{Association for Computational Linguistics},
  \bibinfo{address}{Toronto, Canada}, \bibinfo{pages}{266--278}.
\newblock
\urldef\tempurl%
\url{https://doi.org/10.18653/v1/2023.law-1.26}
\showDOI{\tempurl}


\bibitem[Riley et~al\mbox{.}(2023)]%
        {riley-etal-2023-frmt}
\bibfield{author}{\bibinfo{person}{Parker Riley}, \bibinfo{person}{Timothy
  Dozat}, \bibinfo{person}{Jan~A. Botha}, \bibinfo{person}{Xavier Garcia},
  \bibinfo{person}{Dan Garrette}, \bibinfo{person}{Jason Riesa},
  \bibinfo{person}{Orhan Firat}, {and} \bibinfo{person}{Noah Constant}.}
  \bibinfo{year}{2023}\natexlab{}.
\newblock \showarticletitle{{FRMT}: A Benchmark for Few-Shot Region-Aware
  Machine Translation}.
\newblock \bibinfo{journal}{\emph{Transactions of the Association for
  Computational Linguistics}}  \bibinfo{volume}{11} (\bibinfo{year}{2023}),
  \bibinfo{pages}{671--685}.
\newblock
\urldef\tempurl%
\url{https://doi.org/10.1162/tacl_a_00568}
\showDOI{\tempurl}


\bibitem[Roberts(2021)]%
        {roberts2021linguistic}
\bibfield{author}{\bibinfo{person}{Celia Roberts}.}
  \bibinfo{year}{2021}\natexlab{}.
\newblock \showarticletitle{Linguistic penalties and the job interview}.
\newblock \bibinfo{journal}{\emph{(No Title)}} (\bibinfo{year}{2021}).
\newblock


\bibitem[Roy et~al\mbox{.}(2020)]%
        {roy2020parsing}
\bibfield{author}{\bibinfo{person}{Samapika Roy}, \bibinfo{person}{Sukhada
  Sukhada}, {and} \bibinfo{person}{Anil~Kumar Singh}.}
  \bibinfo{year}{2020}\natexlab{}.
\newblock \showarticletitle{Parsing Indian English News Headlines}. In
  \bibinfo{booktitle}{\emph{ICON}}. \bibinfo{pages}{239--242}.
\newblock


\bibitem[Saadany et~al\mbox{.}(2022)]%
        {saadany-etal-2022-semi}
\bibfield{author}{\bibinfo{person}{Hadeel Saadany}, \bibinfo{person}{Constantin
  Or{\u{a}}san}, \bibinfo{person}{Emad Mohamed}, {and} \bibinfo{person}{Ashraf
  Tantawy}.} \bibinfo{year}{2022}\natexlab{}.
\newblock \showarticletitle{A Semi-supervised Approach for a Better Translation
  of Sentiment in Dialectical {A}rabic {UGT}}. In
  \bibinfo{booktitle}{\emph{Arabic Natural Language Processing Workshop}}.
  \bibinfo{publisher}{Association for Computational Linguistics},
  \bibinfo{address}{Abu Dhabi, United Arab Emirates (Hybrid)}.
\newblock
\urldef\tempurl%
\url{https://doi.org/10.18653/v1/2022.wanlp-1.20}
\showDOI{\tempurl}


\bibitem[Sajjad et~al\mbox{.}(2020)]%
        {sajjad-etal-2020-arabench}
\bibfield{author}{\bibinfo{person}{Hassan Sajjad}, \bibinfo{person}{Ahmed
  Abdelali}, \bibinfo{person}{Nadir Durrani}, {and} \bibinfo{person}{Fahim
  Dalvi}.} \bibinfo{year}{2020}\natexlab{}.
\newblock \showarticletitle{{A}ra{B}ench: Benchmarking Dialectal
  {A}rabic-{E}nglish Machine Translation}. In
  \bibinfo{booktitle}{\emph{COLING}}. \bibinfo{publisher}{International
  Committee on Computational Linguistics}, \bibinfo{address}{Barcelona, Spain
  (Online)}, \bibinfo{pages}{5094--5107}.
\newblock
\urldef\tempurl%
\url{https://doi.org/10.18653/v1/2020.coling-main.447}
\showDOI{\tempurl}


\bibitem[Salameh et~al\mbox{.}(2018)]%
        {salameh2018fine}
\bibfield{author}{\bibinfo{person}{Mohammad Salameh}, \bibinfo{person}{Houda
  Bouamor}, {and} \bibinfo{person}{Nizar Habash}.}
  \bibinfo{year}{2018}\natexlab{}.
\newblock \showarticletitle{Fine-grained Arabic dialect identification}. In
  \bibinfo{booktitle}{\emph{COLING}}.
\newblock


\bibitem[Salloum et~al\mbox{.}(2014)]%
        {salloum-etal-2014-sentence}
\bibfield{author}{\bibinfo{person}{Wael Salloum}, \bibinfo{person}{Heba
  Elfardy}, \bibinfo{person}{Linda Alamir-Salloum}, \bibinfo{person}{Nizar
  Habash}, {and} \bibinfo{person}{Mona Diab}.} \bibinfo{year}{2014}\natexlab{}.
\newblock \showarticletitle{Sentence Level Dialect Identification for Machine
  Translation System Selection}. In \bibinfo{booktitle}{\emph{52nd ACL (Volume
  2: Short Papers)}}. \bibinfo{address}{Baltimore, Maryland},
  \bibinfo{pages}{772--778}.
\newblock
\urldef\tempurl%
\url{https://doi.org/10.3115/v1/P14-2125}
\showDOI{\tempurl}


\bibitem[Salloum and Habash(2022)]%
        {salloum2022unsupervised}
\bibfield{author}{\bibinfo{person}{Wael Salloum} {and} \bibinfo{person}{Nizar
  Habash}.} \bibinfo{year}{2022}\natexlab{}.
\newblock \showarticletitle{Unsupervised Arabic dialect segmentation for
  machine translation}.
\newblock \bibinfo{journal}{\emph{Natural Language Engineering}}
  \bibinfo{volume}{28}, \bibinfo{number}{2} (\bibinfo{year}{2022}),
  \bibinfo{pages}{223--248}.
\newblock


\bibitem[Sandel(2015)]%
        {sandel2015dialects}
\bibfield{author}{\bibinfo{person}{Todd~L Sandel}.}
  \bibinfo{year}{2015}\natexlab{}.
\newblock \showarticletitle{Dialects}.
\newblock \bibinfo{journal}{\emph{The international encyclopedia of language
  and social interaction}} (\bibinfo{year}{2015}), \bibinfo{pages}{1--13}.
\newblock


\bibitem[Sap et~al\mbox{.}(2019)]%
        {sap2019risk}
\bibfield{author}{\bibinfo{person}{Maarten Sap}, \bibinfo{person}{Dallas Card},
  \bibinfo{person}{Saadia Gabriel}, \bibinfo{person}{Yejin Choi}, {and}
  \bibinfo{person}{Noah~A Smith}.} \bibinfo{year}{2019}\natexlab{}.
\newblock \showarticletitle{The risk of racial bias in hate speech detection}.
  In \bibinfo{booktitle}{\emph{ACL}}. \bibinfo{pages}{1668--1678}.
\newblock


\bibitem[Scannell(2020)]%
        {scannell-2020-universal}
\bibfield{author}{\bibinfo{person}{Kevin Scannell}.}
  \bibinfo{year}{2020}\natexlab{}.
\newblock \showarticletitle{{U}niversal {D}ependencies for {M}anx {G}aelic}. In
  \bibinfo{booktitle}{\emph{Workshop on Universal Dependencies (UDW 2020)}}.
  \bibinfo{pages}{152--157}.
\newblock
\urldef\tempurl%
\url{https://aclanthology.org/2020.udw-1.17}
\showURL{%
\tempurl}


\bibitem[Schneider(2012)]%
        {schneider2012appropriate}
\bibfield{author}{\bibinfo{person}{Klaus~P Schneider}.}
  \bibinfo{year}{2012}\natexlab{}.
\newblock \showarticletitle{Appropriate behaviour across varieties of English}.
\newblock \bibinfo{journal}{\emph{Journal of Pragmatics}} \bibinfo{volume}{44},
  \bibinfo{number}{9} (\bibinfo{year}{2012}), \bibinfo{pages}{1022--1037}.
\newblock


\bibitem[Seddah et~al\mbox{.}(2020)]%
        {seddah-etal-2020-building}
\bibfield{author}{\bibinfo{person}{Djam{\'e} Seddah}, \bibinfo{person}{Farah
  Essaidi}, \bibinfo{person}{Amal Fethi}, \bibinfo{person}{Matthieu Futeral},
  \bibinfo{person}{Benjamin Muller}, \bibinfo{person}{Pedro~Javier
  Ortiz~Su{\'a}rez}, \bibinfo{person}{Beno{\^\i}t Sagot}, {and}
  \bibinfo{person}{Abhishek Srivastava}.} \bibinfo{year}{2020}\natexlab{}.
\newblock \showarticletitle{Building a User-Generated Content {N}orth-{A}frican
  {A}rabizi Treebank: Tackling Hell}. In \bibinfo{booktitle}{\emph{ACL}}.
  \bibinfo{publisher}{ACL}, \bibinfo{address}{Online},
  \bibinfo{pages}{1139--1150}.
\newblock
\urldef\tempurl%
\url{https://doi.org/10.18653/v1/2020.acl-main.107}
\showDOI{\tempurl}


\bibitem[Shapiro and Duh(2019)]%
        {shapiro-duh-2019-comparing}
\bibfield{author}{\bibinfo{person}{Pamela Shapiro} {and} \bibinfo{person}{Kevin
  Duh}.} \bibinfo{year}{2019}\natexlab{}.
\newblock \showarticletitle{Comparing Pipelined and Integrated Approaches to
  Dialectal {A}rabic Neural Machine Translation}. In
  \bibinfo{booktitle}{\emph{Sixth Workshop on {NLP} for Similar Languages,
  Varieties and Dialects}}. \bibinfo{publisher}{ACL}, \bibinfo{address}{Ann
  Arbor, Michigan}, \bibinfo{pages}{214--222}.
\newblock
\urldef\tempurl%
\url{https://doi.org/10.18653/v1/W19-1424}
\showDOI{\tempurl}


\bibitem[Shoufan and Alameri(2015)]%
        {shoufan2015natural}
\bibfield{author}{\bibinfo{person}{Abdulhadi Shoufan} {and}
  \bibinfo{person}{Sumaya Alameri}.} \bibinfo{year}{2015}\natexlab{}.
\newblock \showarticletitle{Natural language processing for dialectical Arabic:
  A survey}. In \bibinfo{booktitle}{\emph{Arabic Natural Language Processing
  Workshop}}. \bibinfo{pages}{36--48}.
\newblock


\bibitem[Simaki et~al\mbox{.}(2017)]%
        {simaki-etal-2017-identifying}
\bibfield{author}{\bibinfo{person}{Vasiliki Simaki},
  \bibinfo{person}{Panagiotis Simakis}, \bibinfo{person}{Carita Paradis}, {and}
  \bibinfo{person}{Andreas Kerren}.} \bibinfo{year}{2017}\natexlab{}.
\newblock \showarticletitle{Identifying the Authors{'} National Variety of
  {E}nglish in Social Media Texts}. In \bibinfo{booktitle}{\emph{International
  Conference Recent Advances in Natural Language Processing, {RANLP} 2017}}.
  \bibinfo{publisher}{INCOMA Ltd.}, \bibinfo{address}{Varna, Bulgaria},
  \bibinfo{pages}{671--678}.
\newblock
\urldef\tempurl%
\url{https://doi.org/10.26615/978-954-452-049-6_086}
\showDOI{\tempurl}


\bibitem[Sun et~al\mbox{.}(2023)]%
        {sun-etal-2023-dialect}
\bibfield{author}{\bibinfo{person}{Jiao Sun}, \bibinfo{person}{Thibault
  Sellam}, \bibinfo{person}{Elizabeth Clark}, \bibinfo{person}{Tu Vu},
  \bibinfo{person}{Timothy Dozat}, \bibinfo{person}{Dan Garrette},
  \bibinfo{person}{Aditya Siddhant}, \bibinfo{person}{Jacob Eisenstein}, {and}
  \bibinfo{person}{Sebastian Gehrmann}.} \bibinfo{year}{2023}\natexlab{}.
\newblock \showarticletitle{Dialect-robust Evaluation of Generated Text}. In
  \bibinfo{booktitle}{\emph{ACL}}. \bibinfo{address}{Toronto, Canada},
  \bibinfo{pages}{6010--6028}.
\newblock
\urldef\tempurl%
\url{https://doi.org/10.18653/v1/2023.acl-long.331}
\showDOI{\tempurl}


\bibitem[Tahssin et~al\mbox{.}(2020)]%
        {tahssin-etal-2020-identifying}
\bibfield{author}{\bibinfo{person}{Rawan Tahssin}, \bibinfo{person}{Youssef
  Kishk}, {and} \bibinfo{person}{Marwan Torki}.}
  \bibinfo{year}{2020}\natexlab{}.
\newblock \showarticletitle{Identifying Nuanced Dialect for {A}rabic Tweets
  with Deep Learning and Reverse Translation Corpus Extension System}. In
  \bibinfo{booktitle}{\emph{Fifth Arabic Natural Language Processing
  Workshop}}. \bibinfo{publisher}{ACL}, \bibinfo{address}{Barcelona, Spain
  (Online)}, \bibinfo{pages}{288--294}.
\newblock
\urldef\tempurl%
\url{https://aclanthology.org/2020.wanlp-1.30}
\showURL{%
\tempurl}


\bibitem[Talafha et~al\mbox{.}(2024)]%
        {talafha-etal-2024-casablanca}
\bibfield{author}{\bibinfo{person}{Bashar Talafha}, \bibinfo{person}{Karima
  Kadaoui}, \bibinfo{person}{Samar~Mohamed Magdy}, \bibinfo{person}{Mariem
  Habiboullah}, \bibinfo{person}{Chafei~Mohamed Chafei},
  \bibinfo{person}{Ahmed~Oumar El-Shangiti}, \bibinfo{person}{Hiba Zayed},
  \bibinfo{person}{Mohamedou~Cheikh Tourad}, \bibinfo{person}{Rahaf Alhamouri},
  \bibinfo{person}{Rwaa Assi}, \bibinfo{person}{Aisha Alraeesi},
  \bibinfo{person}{Hour Mohamed}, \bibinfo{person}{Fakhraddin Alwajih},
  \bibinfo{person}{Abdelrahman Mohamed}, \bibinfo{person}{Abdellah El~Mekki},
  \bibinfo{person}{El~Moatez~Billah Nagoudi}, \bibinfo{person}{Benelhadj
  Djelloul~Mama Saadia}, \bibinfo{person}{Hamzah~A. Alsayadi},
  \bibinfo{person}{Walid Al-Dhabyani}, \bibinfo{person}{Sara Shatnawi},
  \bibinfo{person}{Yasir Ech-chammakhy}, \bibinfo{person}{Amal Makouar},
  \bibinfo{person}{Yousra Berrachedi}, \bibinfo{person}{Mustafa Jarrar},
  \bibinfo{person}{Shady Shehata}, \bibinfo{person}{Ismail Berrada}, {and}
  \bibinfo{person}{Muhammad Abdul-Mageed}.} \bibinfo{year}{2024}\natexlab{}.
\newblock \showarticletitle{{C}asablanca: Data and Models for Multidialectal
  {A}rabic Speech Recognition}. In \bibinfo{booktitle}{\emph{EMNLP}}.
  \bibinfo{publisher}{Association for Computational Linguistics},
  \bibinfo{address}{Miami, Florida, USA}, \bibinfo{pages}{21745--21758}.
\newblock
\urldef\tempurl%
\url{https://aclanthology.org/2024.emnlp-main.1211}
\showURL{%
\tempurl}


\bibitem[Tan et~al\mbox{.}(2020)]%
        {tan2020mind}
\bibfield{author}{\bibinfo{person}{Samson Tan}, \bibinfo{person}{Shafiq Joty},
  \bibinfo{person}{Lav Varshney}, {and} \bibinfo{person}{Min-Yen Kan}.}
  \bibinfo{year}{2020}\natexlab{}.
\newblock \showarticletitle{Mind Your Inflections! Improving NLP for
  Non-Standard Englishes with Base-Inflection Encoding}. In
  \bibinfo{booktitle}{\emph{EMNLP}}. \bibinfo{pages}{5647--5663}.
\newblock


\bibitem[Vaillant(2008)]%
        {vaillant-2008-layered}
\bibfield{author}{\bibinfo{person}{Pascal Vaillant}.}
  \bibinfo{year}{2008}\natexlab{}.
\newblock \showarticletitle{A Layered Grammar Model: Using {T}ree-{A}djoining
  {G}rammars to Build a Common Syntactic Kernel for Related Dialects}. In
  \bibinfo{booktitle}{\emph{Ninth International Workshop on Tree Adjoining
  Grammar and Related Frameworks ({TAG}+9)}}. \bibinfo{address}{T{\"u}bingen,
  Germany}, \bibinfo{pages}{157--164}.
\newblock
\urldef\tempurl%
\url{https://aclanthology.org/W08-2321}
\showURL{%
\tempurl}


\bibitem[Van Der~Goot et~al\mbox{.}(2021)]%
        {van2021masked}
\bibfield{author}{\bibinfo{person}{Rob Van Der~Goot}, \bibinfo{person}{Ibrahim
  Sharaf}, \bibinfo{person}{Aizhan Imankulova}, \bibinfo{person}{Ahmet
  {\"U}st{\"u}n}, \bibinfo{person}{Marija Stepanovi{\'c}},
  \bibinfo{person}{Alan Ramponi}, \bibinfo{person}{Siti~Oryza Khairunnisa},
  \bibinfo{person}{Mamoru Komachi}, {and} \bibinfo{person}{Barbara Plank}.}
  \bibinfo{year}{2021}\natexlab{}.
\newblock \showarticletitle{From Masked Language Modeling to Translation:
  Non-English Auxiliary Tasks Improve Zero-shot Spoken Language Understanding}.
  In \bibinfo{booktitle}{\emph{NAACL}}. \bibinfo{pages}{2479--2497}.
\newblock


\bibitem[Vidal-Gor{\`e}ne et~al\mbox{.}(2024)]%
        {vidal-gorene-etal-2024-cross}
\bibfield{author}{\bibinfo{person}{Chahan Vidal-Gor{\`e}ne},
  \bibinfo{person}{Nadi Tomeh}, {and} \bibinfo{person}{Victoria Khurshudyan}.}
  \bibinfo{year}{2024}\natexlab{}.
\newblock \showarticletitle{Cross-Dialectal Transfer and Zero-Shot Learning for
  {A}rmenian Varieties: A Comparative Analysis of {RNN}s, Transformers and
  {LLM}s}. In \bibinfo{booktitle}{\emph{4th International Conference on Natural
  Language Processing for Digital Humanities}}. \bibinfo{publisher}{Association
  for Computational Linguistics}, \bibinfo{address}{Miami, USA},
  \bibinfo{pages}{438--449}.
\newblock


\bibitem[Wang et~al\mbox{.}(2017)]%
        {wang-etal-2017-universal}
\bibfield{author}{\bibinfo{person}{Hongmin Wang}, \bibinfo{person}{Yue Zhang},
  \bibinfo{person}{GuangYong~Leonard Chan}, \bibinfo{person}{Jie Yang}, {and}
  \bibinfo{person}{Hai~Leong Chieu}.} \bibinfo{year}{2017}\natexlab{}.
\newblock \showarticletitle{{U}niversal {D}ependencies Parsing for Colloquial
  Singaporean {E}nglish}. In \bibinfo{booktitle}{\emph{ACL}}.
  \bibinfo{pages}{1732--1744}.
\newblock
\urldef\tempurl%
\url{https://doi.org/10.18653/v1/P17-1159}
\showDOI{\tempurl}


\bibitem[Wang et~al\mbox{.}(2022)]%
        {wang-etal-2022-perceptual}
\bibfield{author}{\bibinfo{person}{Yizhou Wang}, \bibinfo{person}{Rikke~L.
  Bundgaard-Nielsen}, \bibinfo{person}{Brett~J. Baker}, {and}
  \bibinfo{person}{Olga Maxwell}.} \bibinfo{year}{2022}\natexlab{}.
\newblock \showarticletitle{Perceptual Overlap in Classification of {L}2
  Vowels: {A}ustralian {E}nglish Vowels Perceived by Experienced {M}andarin
  Listeners}. In \bibinfo{booktitle}{\emph{PACLIC}}. \bibinfo{pages}{317--324}.
\newblock


\bibitem[Xiao et~al\mbox{.}(2023)]%
        {xiao2023task}
\bibfield{author}{\bibinfo{person}{Zedian Xiao}, \bibinfo{person}{William
  Held}, \bibinfo{person}{Yanchen Liu}, {and} \bibinfo{person}{Diyi Yang}.}
  \bibinfo{year}{2023}\natexlab{}.
\newblock \showarticletitle{Task-Agnostic Low-Rank Adapters for Unseen English
  Dialects}. In \bibinfo{booktitle}{\emph{Findings of ACL}}.
\newblock


\bibitem[Xie et~al\mbox{.}(2024)]%
        {xie-etal-2024-extracting}
\bibfield{author}{\bibinfo{person}{Roy Xie}, \bibinfo{person}{Orevaoghene
  Ahia}, \bibinfo{person}{Yulia Tsvetkov}, {and} \bibinfo{person}{Antonios
  Anastasopoulos}.} \bibinfo{year}{2024}\natexlab{}.
\newblock \showarticletitle{Extracting Lexical Features from Dialects via
  Interpretable Dialect Classifiers}. In \bibinfo{booktitle}{\emph{NAACL}}.
  \bibinfo{pages}{54--69}.
\newblock
\urldef\tempurl%
\url{https://doi.org/10.18653/v1/2024.naacl-short.5}
\showDOI{\tempurl}


\bibitem[Xu et~al\mbox{.}(2015)]%
        {xu-etal-2015-building}
\bibfield{author}{\bibinfo{person}{Fan Xu}, \bibinfo{person}{Xiongfei Xu},
  \bibinfo{person}{Mingwen Wang}, {and} \bibinfo{person}{Maoxi Li}.}
  \bibinfo{year}{2015}\natexlab{}.
\newblock \showarticletitle{Building Monolingual Word Alignment Corpus for the
  Greater {C}hina Region}. In \bibinfo{booktitle}{\emph{Workshop on Language
  Technology for Closely Related Languages, Varieties and Dialects}}.
  \bibinfo{publisher}{Association for Computational Linguistics},
  \bibinfo{address}{Hissar, Bulgaria}, \bibinfo{pages}{85--94}.
\newblock
\urldef\tempurl%
\url{https://aclanthology.org/W15-5414}
\showURL{%
\tempurl}


\bibitem[Yadav(1974)]%
        {yadav1974interactions}
\bibfield{author}{\bibinfo{person}{RS Yadav}.} \bibinfo{year}{1974}\natexlab{}.
\newblock \showarticletitle{Interactions of Written and Spoken Hindi}.
\newblock \bibinfo{journal}{\emph{Indian Literature}} \bibinfo{volume}{17},
  \bibinfo{number}{3} (\bibinfo{year}{1974}), \bibinfo{pages}{61--66}.
\newblock


\bibitem[Younes et~al\mbox{.}(2020)]%
        {younes2020language}
\bibfield{author}{\bibinfo{person}{Jihene Younes}, \bibinfo{person}{Emna
  Souissi}, \bibinfo{person}{Hadhemi Achour}, {and} \bibinfo{person}{Ahmed
  Ferchichi}.} \bibinfo{year}{2020}\natexlab{}.
\newblock \showarticletitle{Language resources for Maghrebi Arabic dialects’
  NLP: a survey}.
\newblock \bibinfo{journal}{\emph{Language Resources and Evaluation}}
  \bibinfo{volume}{54} (\bibinfo{year}{2020}), \bibinfo{pages}{1079--1142}.
\newblock


\bibitem[Zampieri et~al\mbox{.}(2019)]%
        {zampieri-etal-2019-report}
\bibfield{author}{\bibinfo{person}{Marcos Zampieri}, \bibinfo{person}{Shervin
  Malmasi}, \bibinfo{person}{Yves Scherrer}, \bibinfo{person}{Tanja
  Samard{\v{z}}i{\'c}}, \bibinfo{person}{Francis Tyers},
  \bibinfo{person}{Miikka Silfverberg}, \bibinfo{person}{Natalia Klyueva},
  \bibinfo{person}{Tung-Le Pan}, \bibinfo{person}{Chu-Ren Huang},
  \bibinfo{person}{Radu~Tudor Ionescu}, \bibinfo{person}{Andrei~M. Butnaru},
  {and} \bibinfo{person}{Tommi Jauhiainen}.} \bibinfo{year}{2019}\natexlab{}.
\newblock \showarticletitle{A Report on the Third {V}ar{D}ial Evaluation
  Campaign}. In \bibinfo{booktitle}{\emph{VarDial}}. \bibinfo{pages}{1--16}.
\newblock
\urldef\tempurl%
\url{https://doi.org/10.18653/v1/W19-1401}
\showDOI{\tempurl}


\bibitem[Zampieri and Nakov(2021)]%
        {zampieri2021similar}
\bibfield{author}{\bibinfo{person}{Marcos Zampieri} {and}
  \bibinfo{person}{Preslav Nakov}.} \bibinfo{year}{2021}\natexlab{}.
\newblock \bibinfo{booktitle}{\emph{Similar languages, varieties, and dialects:
  a computational perspective}}.
\newblock \bibinfo{publisher}{Cambridge University Press}.
\newblock


\bibitem[Zampieri et~al\mbox{.}(2020)]%
        {zampieri2020natural}
\bibfield{author}{\bibinfo{person}{Marcos Zampieri}, \bibinfo{person}{Preslav
  Nakov}, {and} \bibinfo{person}{Yves Scherrer}.}
  \bibinfo{year}{2020}\natexlab{}.
\newblock \showarticletitle{Natural language processing for similar languages,
  varieties, and dialects: A survey}.
\newblock \bibinfo{journal}{\emph{Natural Language Engineering}}
  \bibinfo{volume}{26}, \bibinfo{number}{6} (\bibinfo{year}{2020}),
  \bibinfo{pages}{595--612}.
\newblock


\bibitem[Zampieri et~al\mbox{.}(2014)]%
        {zampieri2014report}
\bibfield{author}{\bibinfo{person}{Marcos Zampieri}, \bibinfo{person}{Liling
  Tan}, \bibinfo{person}{Nikola Ljube{\v{s}}i{\'c}}, {and}
  \bibinfo{person}{J{\"o}rg Tiedemann}.} \bibinfo{year}{2014}\natexlab{}.
\newblock \showarticletitle{A report on the DSL shared task 2014}. In
  \bibinfo{booktitle}{\emph{first workshop on applying NLP tools to similar
  languages, varieties and dialects}}. \bibinfo{pages}{58--67}.
\newblock


\bibitem[Zampieri et~al\mbox{.}(2015)]%
        {zampieri2015overview}
\bibfield{author}{\bibinfo{person}{Marcos Zampieri}, \bibinfo{person}{Liling
  Tan}, \bibinfo{person}{Nikola Ljube{\v{s}}i{\'c}}, \bibinfo{person}{J{\"o}rg
  Tiedemann}, {and} \bibinfo{person}{Preslav Nakov}.}
  \bibinfo{year}{2015}\natexlab{}.
\newblock \showarticletitle{Overview of the DSL shared task 2015}. In
  \bibinfo{booktitle}{\emph{Workshop on Language Technology for Closely Related
  Languages, Varieties and Dialects}}. \bibinfo{pages}{1--9}.
\newblock


\bibitem[Zbib et~al\mbox{.}(2012)]%
        {zbib-etal-2012-machine}
\bibfield{author}{\bibinfo{person}{Rabih Zbib}, \bibinfo{person}{Erika
  Malchiodi}, \bibinfo{person}{Jacob Devlin}, \bibinfo{person}{David Stallard},
  \bibinfo{person}{Spyros Matsoukas}, \bibinfo{person}{Richard Schwartz},
  \bibinfo{person}{John Makhoul}, \bibinfo{person}{Omar~F. Zaidan}, {and}
  \bibinfo{person}{Chris Callison-Burch}.} \bibinfo{year}{2012}\natexlab{}.
\newblock \showarticletitle{Machine Translation of {A}rabic Dialects}. In
  \bibinfo{booktitle}{\emph{NAACL}}. \bibinfo{address}{Montr{\'e}al, Canada},
  \bibinfo{pages}{49--59}.
\newblock
\urldef\tempurl%
\url{https://aclanthology.org/N12-1006}
\showURL{%
\tempurl}


\bibitem[Zerva et~al\mbox{.}(2024)]%
        {zerva-etal-2024-findings}
\bibfield{author}{\bibinfo{person}{Chrysoula Zerva}, \bibinfo{person}{Frederic
  Blain}, \bibinfo{person}{Jos{\'e}~G. C.~De~Souza}, \bibinfo{person}{Diptesh
  Kanojia}, \bibinfo{person}{Sourabh Deoghare}, \bibinfo{person}{Nuno~M.
  Guerreiro}, \bibinfo{person}{Giuseppe Attanasio}, \bibinfo{person}{Ricardo
  Rei}, \bibinfo{person}{Constantin Orasan}, \bibinfo{person}{Matteo Negri},
  \bibinfo{person}{Marco Turchi}, \bibinfo{person}{Rajen Chatterjee},
  \bibinfo{person}{Pushpak Bhattacharyya}, \bibinfo{person}{Markus Freitag},
  {and} \bibinfo{person}{Andr{\'e} Martins}.} \bibinfo{year}{2024}\natexlab{}.
\newblock \showarticletitle{Findings of the Quality Estimation Shared Task at
  {WMT} 2024: Are {LLM}s Closing the Gap in {QE}?}. In
  \bibinfo{booktitle}{\emph{Proceedings of the Ninth WMT}}.
  \bibinfo{publisher}{ACL}, \bibinfo{address}{Miami, Florida, USA},
  \bibinfo{pages}{82--109}.
\newblock


\bibitem[Zhan et~al\mbox{.}(2024)]%
        {zhan2024renovi}
\bibfield{author}{\bibinfo{person}{Haolan Zhan}, \bibinfo{person}{Zhuang Li},
  \bibinfo{person}{Xiaoxi Kang}, \bibinfo{person}{Tao Feng},
  \bibinfo{person}{Yuncheng Hua}, \bibinfo{person}{Lizhen Qu},
  \bibinfo{person}{Yi Ying}, \bibinfo{person}{Mei~Rianto Chandra},
  \bibinfo{person}{Kelly Rosalin}, \bibinfo{person}{Jureynolds Jureynolds},
  {et~al\mbox{.}}} \bibinfo{year}{2024}\natexlab{}.
\newblock \showarticletitle{RENOVI: A Benchmark Towards Remediating Norm
  Violations in Socio-Cultural Conversations}.
\newblock  (\bibinfo{year}{2024}).
\newblock


\bibitem[Zhan et~al\mbox{.}(2023)]%
        {zhan2023socialdial}
\bibfield{author}{\bibinfo{person}{Haolan Zhan}, \bibinfo{person}{Zhuang Li},
  \bibinfo{person}{Yufei Wang}, \bibinfo{person}{Linhao Luo},
  \bibinfo{person}{Tao Feng}, \bibinfo{person}{Xiaoxi Kang},
  \bibinfo{person}{Yuncheng Hua}, \bibinfo{person}{Lizhen Qu},
  \bibinfo{person}{Lay-Ki Soon}, \bibinfo{person}{Suraj Sharma},
  {et~al\mbox{.}}} \bibinfo{year}{2023}\natexlab{}.
\newblock \showarticletitle{Socialdial: A benchmark for socially-aware dialogue
  systems}. In \bibinfo{booktitle}{\emph{ACM SIGIR}}.
  \bibinfo{pages}{2712--2722}.
\newblock


\bibitem[Zhang et~al\mbox{.}(2021)]%
        {zhang2021disentangling}
\bibfield{author}{\bibinfo{person}{Xiongyi Zhang}, \bibinfo{person}{Jan-Willem
  van~de Meent}, {and} \bibinfo{person}{Byron~C Wallace}.}
  \bibinfo{year}{2021}\natexlab{}.
\newblock \showarticletitle{Disentangling Representations of Text by Masking
  Transformers}. In \bibinfo{booktitle}{\emph{EMNLP}}.
  \bibinfo{pages}{778--791}.
\newblock


\bibitem[Zhao et~al\mbox{.}(2023)]%
        {zhao2023survey}
\bibfield{author}{\bibinfo{person}{Wayne~Xin Zhao}, \bibinfo{person}{Kun Zhou},
  \bibinfo{person}{Junyi Li}, \bibinfo{person}{Tianyi Tang},
  \bibinfo{person}{Xiaolei Wang}, \bibinfo{person}{Yupeng Hou},
  \bibinfo{person}{Yingqian Min}, \bibinfo{person}{Beichen Zhang},
  \bibinfo{person}{Junjie Zhang}, \bibinfo{person}{Zican Dong},
  {et~al\mbox{.}}} \bibinfo{year}{2023}\natexlab{}.
\newblock \showarticletitle{A survey of large language models}.
\newblock \bibinfo{journal}{\emph{arXiv preprint arXiv:2303.18223}}
  (\bibinfo{year}{2023}).
\newblock


\bibitem[Zhao et~al\mbox{.}(2020)]%
        {zhao-etal-2020-semantic}
\bibfield{author}{\bibinfo{person}{Yuanyuan Zhao}, \bibinfo{person}{Weiwei
  Sun}, \bibinfo{person}{Junjie Cao}, {and} \bibinfo{person}{Xiaojun Wan}.}
  \bibinfo{year}{2020}\natexlab{}.
\newblock \showarticletitle{Semantic Parsing for {E}nglish as a Second
  Language}. In \bibinfo{booktitle}{\emph{ACL}}.
  \bibinfo{publisher}{Association for Computational Linguistics},
  \bibinfo{address}{Online}, \bibinfo{pages}{6783--6794}.
\newblock
\urldef\tempurl%
\url{https://doi.org/10.18653/v1/2020.acl-main.606}
\showDOI{\tempurl}


\bibitem[Ziems et~al\mbox{.}(2022)]%
        {ziems-etal-2022-value}
\bibfield{author}{\bibinfo{person}{Caleb Ziems}, \bibinfo{person}{Jiaao Chen},
  \bibinfo{person}{Camille Harris}, \bibinfo{person}{Jessica Anderson}, {and}
  \bibinfo{person}{Diyi Yang}.} \bibinfo{year}{2022}\natexlab{}.
\newblock \showarticletitle{{VALUE}: {U}nderstanding Dialect Disparity in
  {NLU}}. In \bibinfo{booktitle}{\emph{ACL}}. \bibinfo{address}{Dublin,
  Ireland}, \bibinfo{pages}{3701--3720}.
\newblock
\urldef\tempurl%
\url{https://doi.org/10.18653/v1/2022.acl-long.258}
\showDOI{\tempurl}


\bibitem[Ziems et~al\mbox{.}(2023)]%
        {ziems-etal-2023-multi}
\bibfield{author}{\bibinfo{person}{Caleb Ziems}, \bibinfo{person}{William
  Held}, \bibinfo{person}{Jingfeng Yang}, \bibinfo{person}{Jwala Dhamala},
  \bibinfo{person}{Rahul Gupta}, {and} \bibinfo{person}{Diyi Yang}.}
  \bibinfo{year}{2023}\natexlab{}.
\newblock \showarticletitle{Multi-{VALUE}: A Framework for Cross-Dialectal
  {E}nglish {NLP}}. In \bibinfo{booktitle}{\emph{ACL}}.
  \bibinfo{pages}{744--768}.
\newblock
\urldef\tempurl%
\url{https://doi.org/10.18653/v1/2023.acl-long.44}
\showDOI{\tempurl}


\end{thebibliography}
\bibliographystyle{ACM-Reference-Format}
\end{document}